\documentclass{article}

\usepackage[dblblindworkshop,final,nonatbib]{neurips_2026}

\usepackage[utf8]{inputenc}
\usepackage[T1]{fontenc}
\usepackage{url}
\usepackage{booktabs}
\usepackage{tabularx}
\usepackage{amsfonts}
\usepackage{amsmath}
\usepackage{nicefrac}
\usepackage{microtype}
\usepackage{xcolor}
\usepackage{float}
\usepackage{graphicx}
\usepackage[percent]{overpic}
\usepackage[
  backend=biber,
  style=ieee,
  citestyle=numeric,
  sorting=none,
  sortcites=true,
  natbib=true
]{biblatex}
\usepackage[hidelinks]{hyperref}
\usepackage{orcidlink}

\title{Nowcasting Future Performance for Population Pruning in Survival-of-the-Fittest Training}

\workshoptitle{Neural Network Artifacts as a New Data Modality}

\author{%
  Rico St\"adeli$^{1}$\,\orcidlink{0009-0007-9741-6633}
  \quad Léo Meynent$^{1}$\,\orcidlink{0009-0009-8897-1971}
  \quad Boris Knyazev$^{2,3}$\,\orcidlink{0000-0002-9484-1534}
  \quad Damian Borth$^{1,3}$\,\orcidlink{0000-0002-4660-2627} \\[2mm]
  $^{1}$University of St.Gallen, $^{2}$University of Montreal \\
  $^{3}$Weight Space Labs \\[2mm]
  \texttt{\{rico.staedeli, leo.meynent, damian.borth\}@unisg.ch}
}

\begin{document}

\maketitle

\begin{abstract}
Training high-performing neural networks often involves evaluating populations of candidate models with varying hyperparameters, 
architectures, or random seeds. Training every candidate to completion is computationally expensive, yet stopping seemingly weaker 
runs early risks discarding models that would ultimately perform better, as learning curves can cross. To address this challenge, 
we investigate whether intermediate training checkpoints can predict future model accuracy and enable more informed pruning decisions.
Our nowcasters predict test accuracy five epochs ahead or at the end of training.
They are trained with accuracy and loss histories, weight statistics, and SANE representations as inputs.
SANE is a transformer autoencoder that encodes a checkpoint as a sequence of latent tokens.
We then use the nowcasters in \textbf{Survival-of-the-Fittest (SOTF)} training.
SOTF repeatedly ranks runs during training by predicted accuracy and stops the lowest-ranked ones.
Our results show that weight statistics work best for five-epoch forecasts, and SANE representations work best for final accuracy prediction.
For evaluation, we compare SOTF with pruning the population by current accuracy.
In homogeneous model zoos such as the CNN zoos, pruning by current accuracy already retains a final model only slightly worse than the best.
Model zoos with a larger spread in performance, such as the ResNet-18 and ViT zoos, are more difficult to prune by current accuracy alone.
In these zoos, SOTF with nowcasted final accuracy almost closes the gap to the best model. \\
Code is available at: \url{https://github.com/RicoStaedeli/sotf-training}.
\end{abstract}

\section{Introduction}
\label{sec:introduction}

Training a neural network is rarely an isolated procedure. Practical machine-learning
workflows typically train populations of candidate models that vary in their random
seeds, hyperparameters, widths, or architectures before selecting a model for deployment
or further study. Fully training every candidate is expensive, particularly when only a
small fraction of the population will ultimately be retained. Adaptive resource-allocation
methods such as Successive Halving, Hyperband, BOHB, and Population Based Training reduce
this cost by pruning candidates during training and concentrating computation on the most
promising runs
\citep{jamiesonNonstochasticBestArm,liHyperbandNovelBanditBased2018,
falknerBOHBRobustEfficient2018,jaderbergPopulationBasedTraining2017}. Their effectiveness,
however, depends on whether the information available at an intermediate checkpoint
reliably identifies the models that will perform well later.
Current performance is not always such a signal. At any epoch, ordering the candidates by
their observed accuracy defines a \emph{ranking} of the population but this ranking is only
useful for pruning the population if it agrees with the ranking that would be observed once training
finishes. In practice, the accuracy curves of two candidates can \emph{cross}: a model that
currently trails may later overtake a model that currently leads, so their relative order in
the ranking swaps. When such crossings still lie ahead, the intermediate ranking is a
misleading guide to final quality, even though it accurately reflects current performance.
In an irreversible pruning procedure, an eliminated candidate cannot be reconsidered, so a
crossing that occurs after a pruning decision permanently removes what would have been the
better-performing model. Reliable early selection therefore requires more than an accurate
description of what a model currently achieves. It requires also evidence about how its training
trajectory, and hence its position in the ranking, is likely to continue to change.
Weight-space learning offers a natural perspective on this problem by treating neural
network parameters as a data modality rather than only as the output of optimization \citep{han2026wslSurvey}.
Following the taxonomy of Ettling et al.~\citep{ettling2026weight}, existing downstream
tasks can be organized along two complementary axes. The first concerns
the prediction target. \emph{Discriminative} methods map weights to model properties such as
accuracy, generalization, or training hyperparameters
\citep{unterthinerPredictingNeuralNetwork2021,
schurholtSelfSupervisedRepresentationLearning}, whereas \emph{generative} methods synthesize,
edit, or update parameters \citep{hannaGeoSANELearningGeospatial, anonymous2026lfnt}. The second axis concerns the observations supplied to the method,
namely whether a model is represented by an isolated checkpoint or by a sequence of
checkpoints sampled along its optimization trajectory. Combining these axes
distinguishes static property prediction from the less explored setting in which a
partial weight trajectory is used to anticipate future learning behavior.
Neuron Interaction and Nowcasting Networks (NiNo) occupy the generative,
trajectory-conditioned part of this design space. Given recent parameters, NiNo predicts
the parameters of a future checkpoint and uses the generated state to skip optimizer steps
along an individual training run \citep{knyazevAcceleratingTrainingNeuron2025a}. We study the
neighboring discriminative, trajectory-conditioned problem. Given intermediate checkpoint
representations and the metrics observed so far, we predict future scalar performance and
use the resulting nowcasts to rank a population of training runs. In analogy to NiNo, we
refer to these predictions as \emph{performance nowcasts} and to the resulting selection
procedure as Survival-of-the-Fittest training (SOTF training). We propose SOTF training as a new
population-level resource-allocation method for concentrating computation on models with the greatest
predicted potential. By stopping less promising trajectories early, it aims to reduce the
total compute and associated energy required to train and select from a model population.

Our central question is whether intermediate checkpoint weights contain information about
future model quality that is complementary to the observed learning curve, and whether this
information leads to better pruning decisions. The distinction between these two questions
is important. A predictor may reduce numerical nowcasting error without changing which models
are retained, while an imperfect nowcast may still preserve the strongest candidates. We
therefore study the problem in three stages. First, we quantify consequential rank
instability and pairwise cross-overs across nine checkpointed model
zoos. Second, we nowcast horizon-specific and final test accuracy from performance histories,
checkpoint-derived weight statistics, and representations obtained with SANE
\citep{schurholtScalableVersatileWeight2024}. SANE represents a checkpoint as a sequence of
latent tokens, each encoding a local subset of its parameters. We compare a compact embedding
obtained by averaging these tokens with predictors that preserve their window- or token-level
structure. The latter can use differences and relationships between parts of the checkpoint
that are lost when all tokens are collapsed into a single vector. Third, we conduct retrospective
SOTF training evaluations on held-out recorded trajectories. Starting with a population of
partially trained models, each selection round retains the candidates with the strongest
predicted future performance. We then measure whether the models that ultimately perform best
remain in the population and how much training compute the pruning decisions can save, using
this quantity as a proxy for energy savings. Throughout these evaluations, predictors are
trained once on completed historical runs and then applied without refitting to a held-out population.

\section{Related work}
Our work lies at the intersection of three lines of research: the analysis of learning
dynamics, learning from neural network weights, and learning-curve
prediction for adaptive resource allocation.

\label{sec:related-work}
\subsection{Learning dynamics and weight-space trajectories}
Intermediate checkpoints sample a trajectory through parameter space, exposing structure
absent from the final model. Weight changes have been used to monitor training progress
and distinguish optimization phases \citep{schurholtInvestigationWeightSpace2021}. In
representation space, learned features evolve in structured ways during training
\citep{raghuSVCCASingularVector2017}, and directional analyses of parameter trajectories
reveal recurring patterns of exploration and redundancy
\citep{singhHallmarksOptimizationTrajectories2024}. Phase-transition model zoos extend
this view from individual runs to deliberately heterogeneous populations with
qualitatively different learning behavior \citep{schurholtModelZooPhase2025}.
These studies show that optimization trajectories are structured, but not when this
structure helps allocate compute among partially trained models. We therefore ask not
only whether weights or rankings change, but whether these changes are large enough to
affect population quality, and connect trajectory analysis to model selection by
measuring local rank cross-overs throughout training.

\subsection{Learning from neural network weights}

A model zoo is a systematically generated population of neural networks with recorded
training configurations, weights, and performance metrics
\citep{schurholtModelZoosDataset2022, falkModelZooVision2025}. When intermediate checkpoints are retained, each
model also provides a sampled trajectory through training time and weight space. Model zoos
therefore provide the population-level data needed to learn functions over neural network
parameters. On the discriminative side, Unterthiner
et al.\ show that simple layer-wise weight statistics can predict the accuracy of trained
networks without evaluating them on input data
\citep{unterthinerPredictingNeuralNetwork2021}. Learned hyper-representations instead
encode checkpoints into latent spaces that support the prediction of properties such as
accuracy, epoch, and hyperparameters
\citep{schurholtSelfSupervisedRepresentationLearning}. SANE, the Sequential Autoencoder
for Neural Embeddings, is a transformer-based autoencoder for neural-network checkpoints
\citep{schurholtScalableVersatileWeight2024}. It splits parameter tensors into
position-encoded weight tokens and encodes bounded token windows into contextualized latent
tokens, which scales to larger networks; a reconstruction decoder ensures that the latents
retain local parameter structure. Existing
property-prediction tasks mainly ask about the current state of a single model or checkpoint.
We extend this discriminative setting to account for training dynamics by predicting the
future or final accuracy of a model that is still training from its current checkpoint and,
potentially, its preceding trajectory. Generative weight-space methods use similar artifacts for a different output. They generate
new model parameters, reconstruct checkpoints, or modify existing networks. For example,
hyper-representations have been used to sample previously unseen neural network weights
\citep{schurholtHyperRepresentationsGenerativeModels2022,asefawWeightCLIPAligningDatasets2026}. Most closely related to our
focus on training dynamics are weight nowcasting methods. Weight Nowcaster Networks predict
a future parameter state from recent checkpoints and intermittently replace conventional
optimizer steps \citep{jangLearningBoostTraining2023}. NiNo improves this approach by representing
neuron connectivity explicitly and modeling interactions between parameters
\citep{knyazevAcceleratingTrainingNeuron2025a}. Both nowcasting methods generate the future
weights of one trajectory. In contrast, we predict the future scalar performance of many trajectories 
and use these predictions to select among them at the population level: NiNo advances individual
trajectories, whereas our method allocates resources across trajectories.

\subsection{Learning-curve prediction and adaptive resource allocation}

Learning-curve prediction estimates future performance from a partially observed sequence
of metrics. Parametric extrapolation fits a predefined curve family, whereas methods such
as LCNet learn distributions over learning curves across configurations
\citep{kleinLEARNINGCURVEPREDICTION2017}. These approaches provide important baselines for
our task because they use the same metric history available to a pruning policy, but not
the corresponding checkpoint state. We test whether weight statistics and learned SANE
representations supply information beyond this observed performance history.
Adaptive resource-allocation methods turn intermediate observations into training
decisions. Successive Halving repeatedly removes weak configurations, Hyperband evaluates
different trade-offs between initial population size and per-model budget, and BOHB combines
this multi-fidelity allocation with model-based configuration search
\citep{jamiesonNonstochasticBestArm,liHyperbandNovelBanditBased2018,
falknerBOHBRobustEfficient2018}. Population Based Training additionally updates the
configurations of retained models during training
\citep{jaderbergPopulationBasedTraining2017}. These methods primarily rank candidates using
observed performance or configuration-level models. Our SOTF training evaluations
hold the candidate population and its recorded trajectories fixed and isolate a different
question: whether checkpoint-derived nowcasts produce better elimination decisions than
current accuracy or learning curves alone.

\section{Method}
\label{sec:method}

We study a checkpointed model zoo as a population of independently trained models.
For model $i$, the available checkpoints form a trajectory
$\{(\boldsymbol{\theta}_{i,e},y_{i,e}) : e\in\mathcal{E}_i\}$, where
$\boldsymbol{\theta}_{i,e}$ denotes the parameters and $y_{i,e}$ the recorded test
accuracy at epoch $e$. Our corpus contains nine image-classification zoos spanning
CIFAR-10 \cite{krizhevskyLearningMultipleLayers2009}, STL-10 \cite{coatesAnalysisSingleLayerNetworks2011}, 
and Tiny ImageNet \cite{leTinyImageNetVisual2015}. Six zoos contain CNNs and vary random seeds or
training hyperparameters; the remaining zoos contain width-varying ResNet-18s or
Vision Transformers constructed to cover different phase-transition regimes
\citep{schurholtModelZoosDataset2022,schurholtModelZooPhase2025}. The training
horizons range from 50 to 300 epochs. For all predictor experiments, we partition
each zoo at the model level, so all checkpoints from one trajectory belong to exactly
one of the training, validation, or test splits. The training split is used to fit representations,
feature transformations, and predictors; the validation split is used for
hyperparameter tuning and predictor selection; and the test split is reserved for
the retrospective SOTF training evaluation.

\begin{figure}[H]
    \centering
    \includegraphics[width=\linewidth]{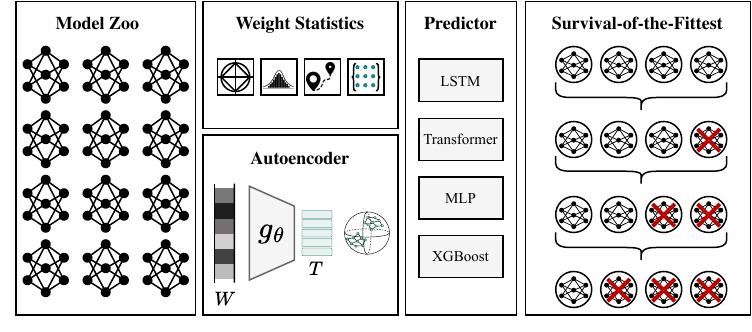}
    \caption[Overview of the methodology]{\textbf{Overview of the methodology.}
    Input to the pipeline is a raw model zoo containing checkpoints with weights
    and recorded performance metrics. The pipeline produces aligned checkpoint-level
    datasets with weight-space representations and performance metrics. These datasets
    are used for descriptive learning-dynamics analysis and for training nowcasting predictors.
    The trained predictors are then held fixed during the
    Survival-of-the-Fittest training (SOTF training) on a held-out population, where weak models are removed before the end of
    their recorded trajectories.}
    \label{fig:overview}
\end{figure}

\subsection{Learning dynamics}
\label{subsec:method-learning-dynamics}

The learning-dynamics analysis establishes when intermediate accuracy is an
unreliable selection signal. Accuracy distributions
describe population-level heterogeneity over time, whereas cross-overs measure
changes in the relative ordering of individual models.

\paragraph{Accuracy distributions over training.}
For every zoo and saved epoch $e$, we form the empirical distribution
$\mathcal{Y}_e=\{y_{i,e}:e\in\mathcal{E}_i\}$. We visualize these epoch-wise
distributions together with a small sample of individual learning curves and
summarize their center and spread using the mean, standard deviation, interquartile
range, and 10th--90th percentile interval. Because the zoos have different training
horizons, comparisons across zoos use normalized progress $p_{i,e}=e/e_i^{\max}$;
within-zoo analyses retain the original epoch scale. The resulting summaries reveal
whether a population converges to a narrow performance band or retains distinct
groups of weak and strong models throughout training.

\paragraph{Pairwise cross-overs.}
For two models $i$ and $j$ observed at consecutive shared checkpoints $e<e'$, a raw
cross-over occurs when their accuracy difference changes sign. The number of such
events at a transition is
\begin{equation}
  C_{e\rightarrow e'}=
  \sum_{i<j}\mathbb{I}\!\left[
    (y_{i,e}-y_{j,e})(y_{i,e'}-y_{j,e'})<0
  \right].
  \label{eq:pairwise-crossovers}
\end{equation}
Raw counts can be dominated by reversals between nearly tied models. We therefore
also report \emph{$\epsilon$-significant cross-overs}, for which both pairwise
margins must exceed $\epsilon$:
\begin{equation}
  C^{(\epsilon)}_{e\rightarrow e'}=
  \sum_{i<j}\mathbb{I}\!\left[
    (y_{i,e}-y_{j,e})(y_{i,e'}-y_{j,e'})<0,
    \ |y_{i,e}-y_{j,e}|\geq\epsilon,
    \ |y_{i,e'}-y_{j,e'}|\geq\epsilon
  \right],
  \label{eq:significant-crossovers}
\end{equation}
with $\epsilon=0.02$, i.e., two accuracy percentage points on the $[0,1]$ scale.
We accompany these local transition counts with Kendall's $\tau$ between the
ranking at epoch $e$ and the final ranking. This distinguishes transient local
reordering from changes that affect which models eventually finish near the top.

\subsection{Performance predictors and SOTF training}
\label{subsec:method-predictors-sotf}

\paragraph{Predictor training.}
For each model zoo and prediction target, we train a predictor once before the SOTF training.
The completed trajectories in the training split provide the checkpoint features and future
target values used for fitting, and the configuration is selected on the validation split. The resulting predictor is then transferred to the test
split, which represents a new population from the same zoo, and remains fixed throughout
population pruning. Learned representation encoders and feature transformations remain fixed as well.
At a decision epoch, the predictor may inspect only the current
and preceding checkpoints of an active trajectory. Later checkpoints and the target value of
that trajectory remain hidden from the policy and are revealed only when computing outcome metrics.

\paragraph{Nowcasting task.}
We consider two regression targets: (1) the final value
$z^{\mathrm{final}}_{i,e}=y_{i,e_i^{\max}}$ and (2) the fixed-horizon value
$z^{\mathrm{hor}}_{i,e}=y_{i,e+h}$, using $h=5$ epochs in the main horizon
experiments. Samples without the required future checkpoint are omitted. For a
fixed-length checkpoint representation $\mathbf{x}_{i,e}$, the design vector
contains the current representation, the three preceding available representations,
their first-order differences, epoch and normalized-progress variables, and,
depending on the ablation, the observed accuracy and loss history. We compare three sources of information: performance history alone, a checkpoint
representation alone, and their combination. Performance features comprise current
and lagged accuracy and loss, recent changes, epoch, and training progress.
Engineered weight features are distribution statistics of the parameters (mean, variance,
percentiles, and range), computed for the full model and, depending on the configuration,
per layer (Appendix~\ref{app:weight-statistics}). Learned representations
are produced by SANE \citep{schurholtScalableVersatileWeight2024}. Its
128-dimensional latent tokens are used either as a mean-pooled checkpoint embedding,
as a sequence of pooled latent windows, or as the complete token sequence. The last
two variants preserve within-checkpoint structure that is removed by global mean
pooling. Fixed-length features are modeled with ridge regression, multilayer perceptrons, and
XGBoost. A unidirectional LSTM consumes the latent-window sequence, while a
hierarchical Transformer first encodes local groups of SANE tokens and then relates
the resulting group representations globally. We compare these models with the
current-value rule, deterministic parametric learning-curve extrapolation, and an
LCNet-style learning-curve predictor \citep{kleinLEARNINGCURVEPREDICTION2017}.
Predictor configurations are compared on validation trajectories with RMSE, MAE,
and $R^2$. Feature ablations compare representation-only, performance-only, and
hybrid inputs under the same model-level splits. The selected predictors are then
held fixed and evaluated as ranking policies in SOTF training on the test trajectories.

\begin{figure}[t]
  \centering
  \small
  \includegraphics[width=\linewidth]{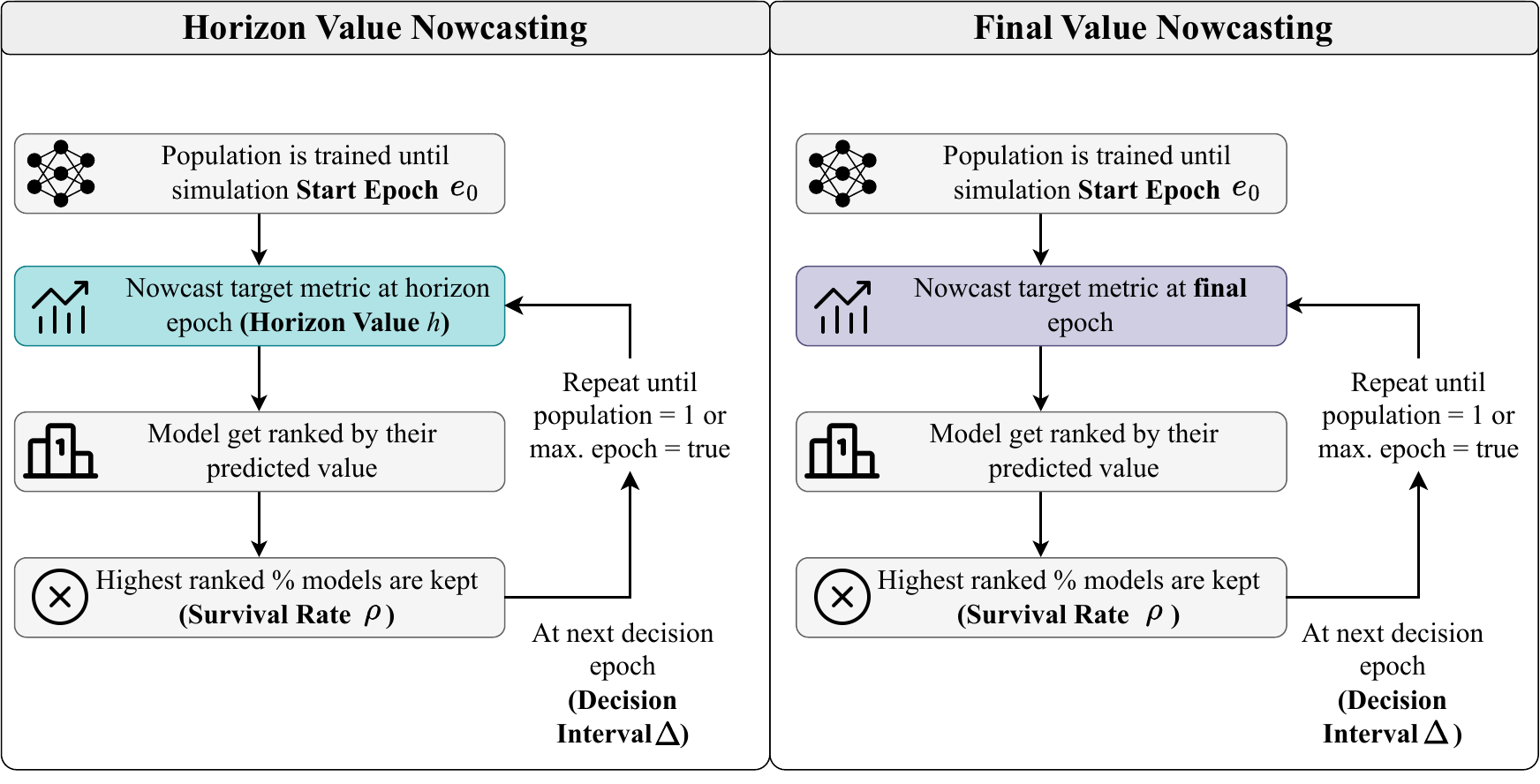}
  \caption{\textbf{Survival-of-the-Fittest training.} Starting at epoch $e_0$,
  the same fixed predictor, trained beforehand, estimates either performance five epochs ahead
  (left) or final performance (right). At intervals of $\Delta$ epochs, active models
  are ranked by their nowcasted test accuracy and the top fraction $\rho$ is retained. The procedure
  stops when one model remains or the recorded training horizon is reached.}
  \label{fig:sotf-algorithm}
\end{figure}

\paragraph{Nowcasting-guided population pruning.}
Prediction error does not fully determine the value of a nowcasting method because two
predictors with similar RMSE can induce different rankings near a pruning boundary.
We therefore propose \emph{Survival-of-the-Fittest training} (SOTF training), a
population-level resource-allocation method designed to reduce the compute and associated energy cost of
training a population of models. It uses nowcasted future performance to allocate the remaining
training budget, continuing promising runs while stopping weaker candidates early. SOTF training begins
with all candidate models active. At each decision round, it assigns every active model a
performance nowcast, ranks the population by these estimates, and retains the top fraction
specified by the retention rate $\rho$. The number of retained models is rounded up, and at
least one candidate always remains. Eliminated models cannot return. The main experiment uses a
five-epoch decision interval, retention rates $\rho\in\{0.10,0.25,0.50,0.75,0.90\}$, and three
zoo-specific start epochs between $7\%$ and $50\%$ of the training horizon
(Appendix Table~\ref{tab:app-sotf-setup-sweeps}). We vary
the nowcasting target, representation, and predictor family. At every decision round,
SOTF training applies the same fixed predictor described above. Because the complete test trajectories
are recorded, we can measure what eliminated models would ultimately have achieved. Their future
checkpoints are used only to compute outcomes and are never exposed to the predictor. Our primary measure
is the best-retained-model gap
\begin{equation}
  G=\max_{i\in\mathcal{A}_0}y_{i,e_i^{\max}}
  -\max_{i\in\mathcal{A}_{\mathrm{final}}}y_{i,e_i^{\max}},
  \label{eq:best-retained-gap}
\end{equation}
where $\mathcal{A}_0$ is the initial population and $\mathcal{A}_{\mathrm{final}}$ is the
set of models that survive pruning. Thus, $G$ is the final test-accuracy loss caused by
pruning: it subtracts the accuracy of the best survivor from that of the best model in the
original population, using the fully recorded trajectories to evaluate both; $G=0$ means
that the population's best final model survives. To assess whether SOTF training preserves the
best models more broadly, we also report oracle top-$K$ recall, where
$K=|\mathcal{A}_{\mathrm{final}}|$: the fraction of the true top $K$ models by recorded final
accuracy that appear among SOTF training's $K$ survivors. We additionally report final population size
and the fraction of training epochs saved
relative to fully training the initial population. Saved epochs serve as an architecture-
and hardware-independent proxy for compute and energy reduction; we do not directly measure
wall-clock energy consumption. Reporting selection quality and efficiency together prevents
the degenerate no-pruning policy from appearing optimal and tests whether lower nowcasting
error translates into a better allocation of the population's training budget.

\section{Results}
\label{sec:results}

We first characterize how model rankings evolve during training before evaluating
nowcasting accuracy and selection performance.

\subsection{Accuracy rankings remain unstable during training}

\begin{figure}[tb]
  \centering
  \small
  \includegraphics[width=0.75\linewidth]{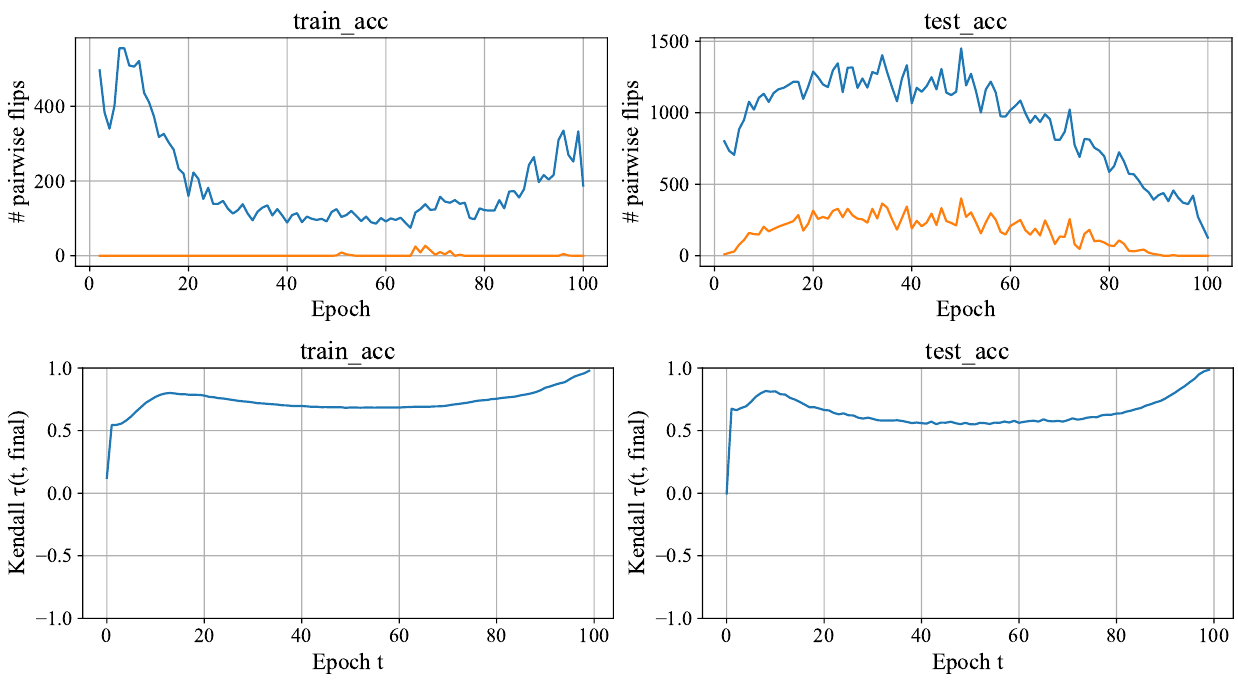}
  \caption{\textbf{Tiny ImageNet ResNet-18.} The upper row shows pairwise rank changes:
  blue denotes the raw number of cross-overs and orange the number of $\epsilon$-significant
  cross-overs. The lower row shows Kendall's $\tau$ between the ranking at epoch $e$ and the
  final ranking.}
  \label{fig:rn18_tiniyimagenet_co_kendall}
\end{figure}

The checkpoint distributions reveal two qualitatively different population regimes
(Appendix Figure~\ref{fig:accuracy-distributions-all}). The uniform CIFAR-10 and
STL-10 CNN zoos converge to narrow final distributions, with standard deviations below
one percentage point. Their hyperparameter variants, the width-varying ResNet-18 zoos,
and the ViT zoo retain much wider spreads. For example, final test accuracy has standard
deviation $16.1$ percentage points in the Tiny ImageNet ResNet-18 zoo and $22.8$
percentage points in the CIFAR-10 ViT zoo. These spreads provide the scale for the
ranking analysis. Rank changes in a narrow zoo often correspond to small absolute
accuracy differences, whereas the same changes in a heterogeneous zoo can span much
larger differences. Intermediate rankings nevertheless remain unstable in both regimes.
Figure~\ref{fig:rn18_tiniyimagenet_co_kendall}
visualizes the Tiny ImageNet ResNet-18 zoo, whose 192 models induce $18{,}336$
unordered pairs at every transition. Moreover, agreement with the final ranking is
non-monotonic. Kendall's $\tau$ falls from $0.814$
at epoch 10 to $0.551$ at epoch 50 before approaching one near the end of training.
The distinction between raw and significant reversals is important. In the uniform
STL-10 CNN zoo (Appendix Figure~\ref{fig:app-flips-ds8}), the raw count remains high because many models are nearly tied, but
the significant count falls from $23{,}817$ at the first transition to 74 at the
second, 1 at the third, and zero thereafter. The Tiny ImageNet zoo follows a different pattern and retains
many significant local cross-overs throughout training. Raw cross-over counts, significant
cross-over counts, and rank correlation therefore describe complementary properties of the
observed learning dynamics.

\begin{figure}[H]
  \centering
  \small
  \includegraphics[width=0.9\linewidth]{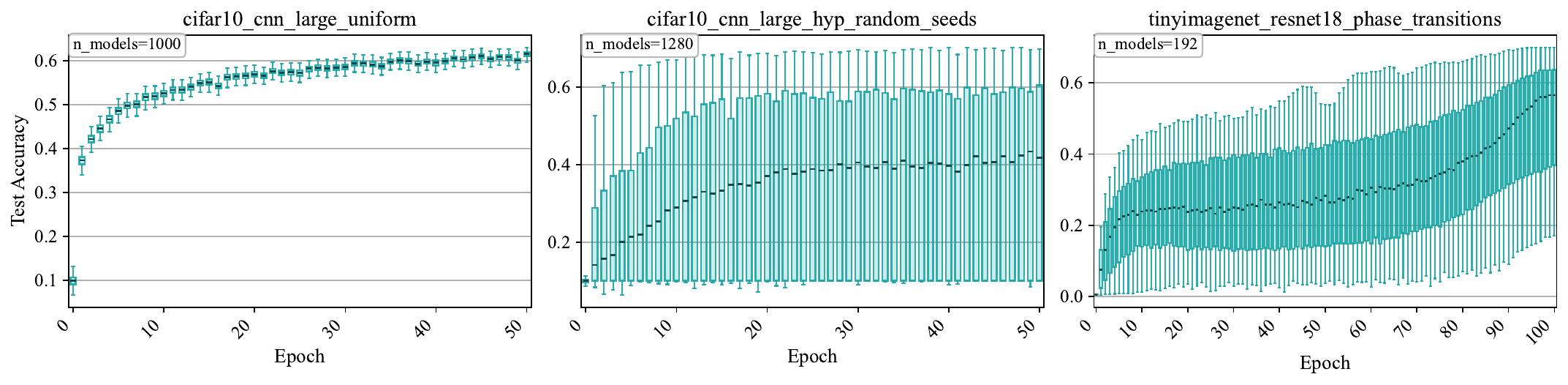}
  \caption{\textbf{Distribution of test accuracy.} Distribution of test accuracy over training for three model zoos.}
  \label{fig:accuracy-dist-3-zoos}
\end{figure}

\subsection{Checkpoint predictors improve metric nowcasting}

\begin{figure}[H]
  \centering
  \includegraphics[width=0.9\linewidth]{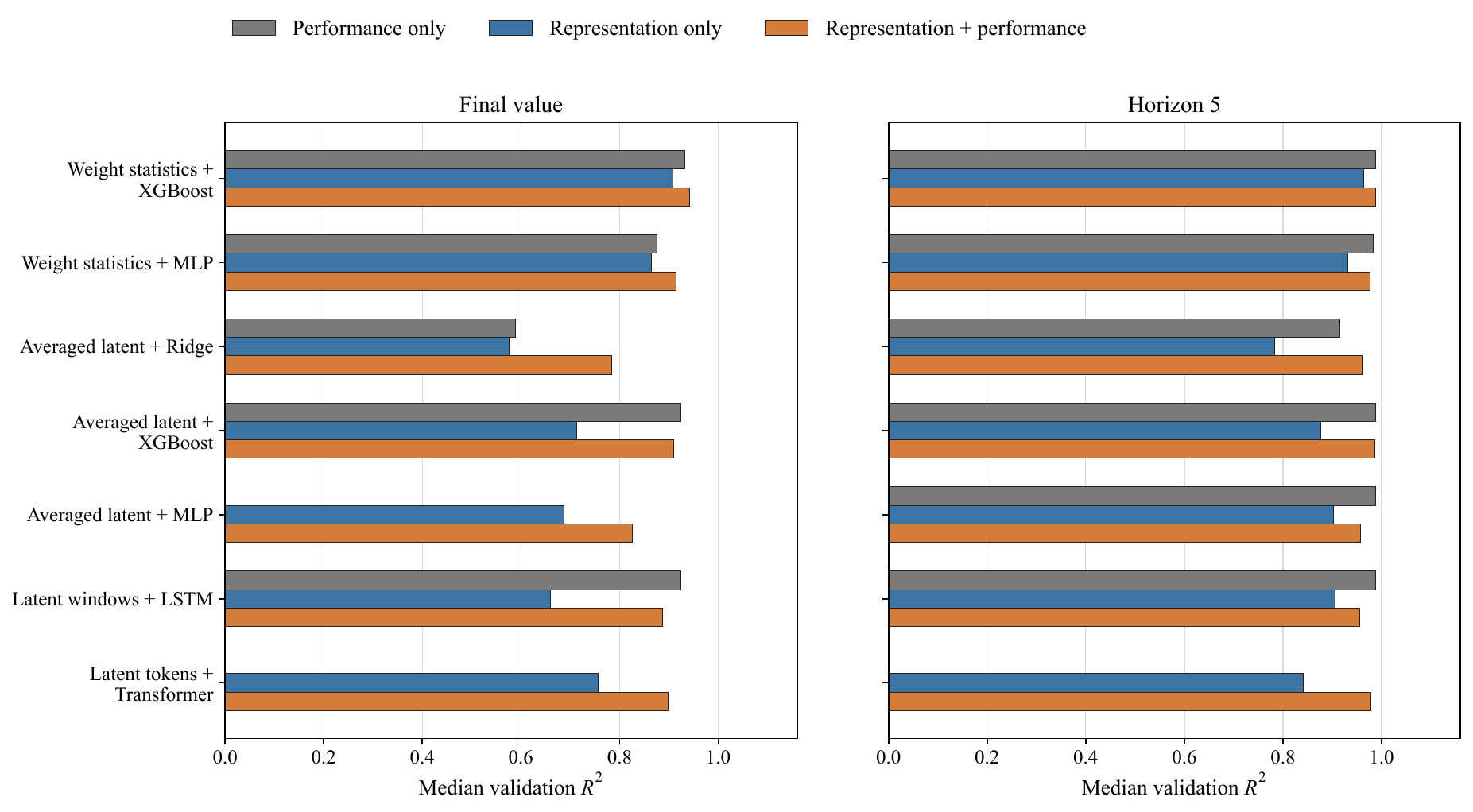}
  \caption{\textbf{Nowcasting feature ablation.} Bars show median validation $R^2$ for
  performance-only, representation-only, and hybrid configurations. Values below zero
  are clipped to zero. A performance-only token-Transformer variant is undefined because
  its architecture requires a latent-token sequence.}
  \label{fig:predictor-feature-ablation}
\end{figure}

\begin{figure}[tbp]
  \centering
  \includegraphics[width=\linewidth]{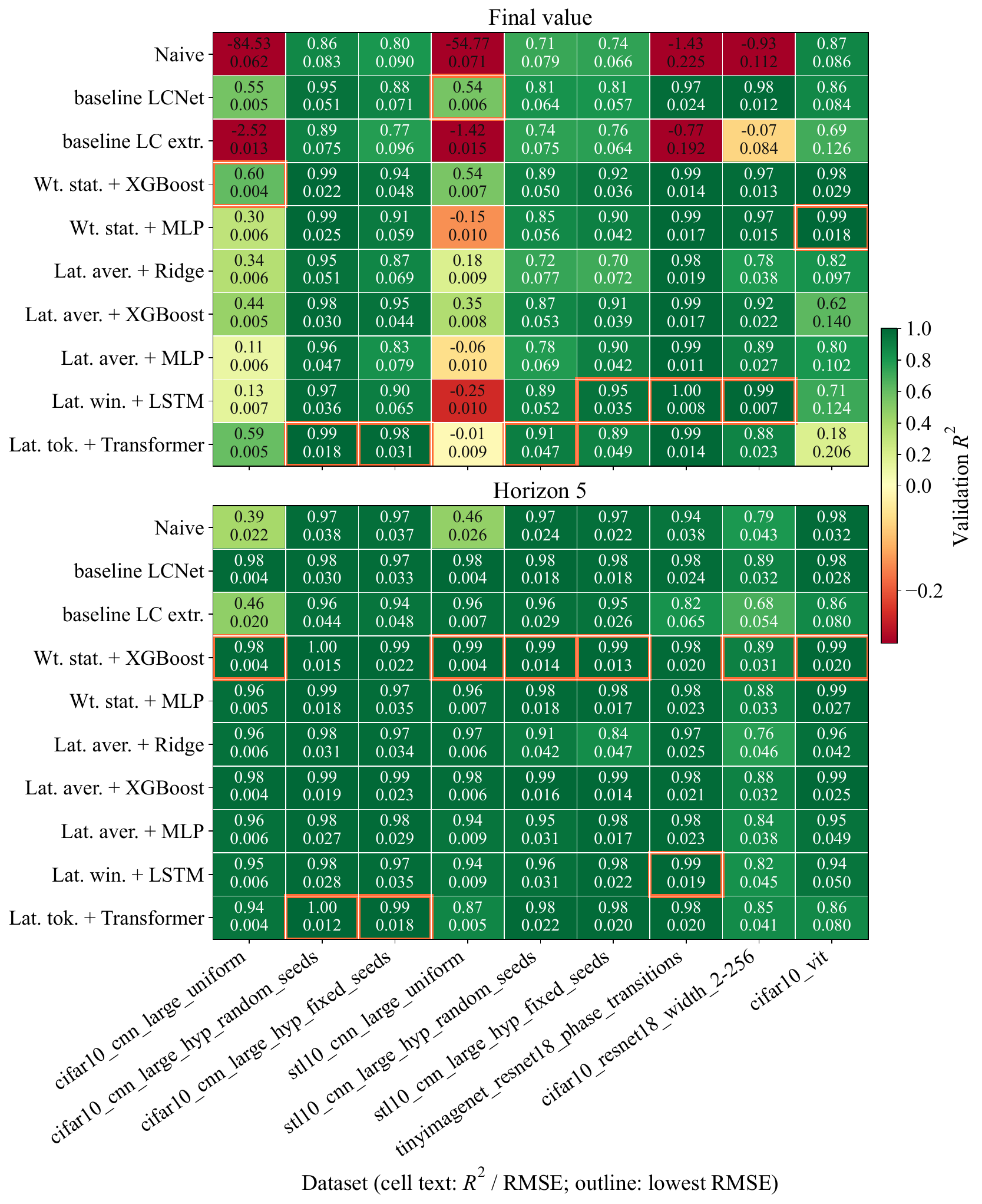}
  \caption{\textbf{Predictor-family comparison.} Each cell reports validation $R^2$ (top)
  and RMSE (bottom) for the selected configuration; outlines mark the lowest-RMSE
  learned predictor for each zoo and target.}
  \label{fig:predictor-family-comparison}
\end{figure}

Figure~\ref{fig:predictor-family-comparison} compares the best validated configuration
within each predictor family under the training procedure described above. Horizon-5 nowcasting is close to saturated on most zoos,
yet checkpoint-based predictors still improve on the learning-curve baselines. The best
weight-statistic predictor outperforms LCNet on all nine zoos, wins six of nine comparisons
against the best SANE predictor, and reduces RMSE by a median of $21.5\%$ relative to
LCNet. XGBoost is the most robust model for these short-range nowcasts, indicating that
recent metric changes and compact checkpoint statistics provide an effective local
continuation signal. Final-value prediction produces a clearer separation between representation families.
The best SANE-based predictor has lower RMSE than the best weight-statistic predictor
on six of nine zoos and achieves a median reduction of $37.9\%$ relative to LCNet,
compared with $32.1\%$ for the best weight-statistic predictor. Several wins come from
the latent-window LSTM and latent-token Transformer, suggesting that relationships among
parts of a checkpoint matter more when extrapolating to the end of training. No family
dominates every zoo, however; representation utility depends on both population structure
and nowcasting horizon. The feature ablation in Figure~\ref{fig:predictor-feature-ablation} clarifies that the
weights are complementary to, rather than a substitute for, observed performance. Pure
checkpoint representations retain substantial signal, particularly for final-value
prediction, while performance-only features are strong over the short horizon. Combining
both sources is usually best or close to best and substantially stabilizes the SANE
sequence predictors. The metric history describes how well a model currently learns;
the checkpoint representation adds information about the parameter state from which that
trajectory will continue.

\subsection{SOTF training improves retained-model quality in the difficult regimes}

Nowcasting can improve pruning decisions, but the amount of available headroom varies
sharply by zoo. Table~\ref{tab:sotf-regime} therefore reports selection results separately
for the six CNN zoos and the three ResNet-18 and ViT zoos. On the CNN zoos, pruning by current
accuracy already retains a model within half a percentage point of the population's best
(mean gap $0.0051$), and no learned predictor improves on this by more than $0.13$ percentage
points. These populations are nearly interchangeable at the top, so little remains to be gained.
The ResNet-18 and ViT zoos expose the relevant regime, in which current performance and
remaining potential diverge: pruning by current accuracy loses $6.4$ percentage points on
average and retains only $11\%$ of the oracle top-$K$. For final-value selection, LCNet, which
sees only the accuracy history, reduces the gap to $0.0247$, and an XGBoost nowcaster trained
on performance features alone to $0.0051$. Adding
weight statistics reduces it further to $0.0017$ and raises oracle top-$K$ recall from $0.20$
to $0.72$; a paired bootstrap over the 44 matched configurations places this additional gap
reduction at $0.0034$ (95\% CI $[0.0018, 0.0051]$). Weight statistics without performance
features perform almost as well and are never worse than current-accuracy pruning in any of
these configurations. Horizon-5 selection gains are smaller and not robust: the best policy
reduces the gap from $0.0629$ to $0.0479$, but the 95\% confidence interval of this improvement
includes zero. SANE-based predictors help on individual zoos -- on Tiny ImageNet, averaged-latent
XGBoost with performance features retains the best model in nearly every configuration
(Appendix Table~\ref{tab:app-sotf-per-zoo}) -- but are unstable on the ViT zoo. Against LCNet,
only averaged-latent XGBoost among them improves the regime-level gap ($0.0185$ vs.\ $0.0247$).
This means a model that produces more accurate nowcasts at every
checkpoint need not preserve the strongest candidate at a pruning boundary.

\begin{table}[tb]
  \centering
  \small
  \caption{\textbf{SOTF training results by population regime.} Each cell reports the mean
  best-retained-model gap / mean oracle top-$K$ recall over all start epochs and retention rates,
  separately for the six CNN zoos and the three ResNet-18 and ViT zoos. Lower gap and higher recall
  are better; the best value per column is in bold. ``perf.'' denotes accuracy and loss history
  features, ``weights'' engineered weight statistics, and ``SANE'' SANE latent representations. All
  predictors are trained once and held fixed. Results for all policies are in
  Appendix~\ref{app:sotf-full-results}.}
  \label{tab:sotf-regime}
  \setlength{\tabcolsep}{4pt}
  \footnotesize
  \begin{tabular}{@{}lcccc@{}}
    \toprule
     & \multicolumn{2}{c}{\textbf{Final value}} & \multicolumn{2}{c}{\textbf{Horizon 5}} \\
    \cmidrule(lr){2-3}\cmidrule(lr){4-5}
    \textbf{Ranking policy} & CNN & ResNet-18 / ViT & CNN & ResNet-18 / ViT \\
    \midrule
    Current accuracy & 0.0051 / \textbf{0.48} & 0.0643 / 0.11 & 0.0051 / 0.48 & 0.0629 / 0.11 \\
    LC extrapolation & 0.0062 / 0.47 & 0.0689 / 0.05 & 0.0050 / \textbf{0.51} & 0.0490 / 0.05 \\
    LCNet & 0.0056 / \textbf{0.48} & 0.0247 / 0.10 & 0.0104 / 0.43 & 0.0656 / 0.11 \\
    \midrule
    XGB (perf.) & \textbf{0.0040} / 0.45 & 0.0051 / 0.20 & \textbf{0.0039} / 0.46 & 0.0495 / 0.09 \\
    XGB (weights) & 0.0060 / 0.34 & 0.0018 / 0.67 & 0.0056 / 0.37 & 0.0608 / 0.09 \\
    XGB (perf.\ + weights) & 0.0054 / 0.42 & \textbf{0.0017} / \textbf{0.72} & 0.0043 / 0.45 & \textbf{0.0479} / \textbf{0.16} \\
    \midrule
    Avg.\ latent XGB (perf.\ + SANE) & 0.0055 / 0.38 & 0.0185 / 0.44 & 0.0058 / 0.44 & 0.0622 / 0.09 \\
    Latent-window LSTM (perf.\ + SANE) & 0.0107 / 0.33 & 0.0241 / 0.27 & 0.0110 / 0.33 & 0.0580 / 0.05 \\
    Latent-token Transf.\ (perf.\ + SANE) & 0.0060 / 0.36 & 0.0991 / 0.19 & 0.0061 / 0.43 & 0.0504 / 0.06 \\
    \bottomrule
  \end{tabular}
\end{table}

\begin{figure}[tb]
  \centering
  \includegraphics[width=\linewidth]{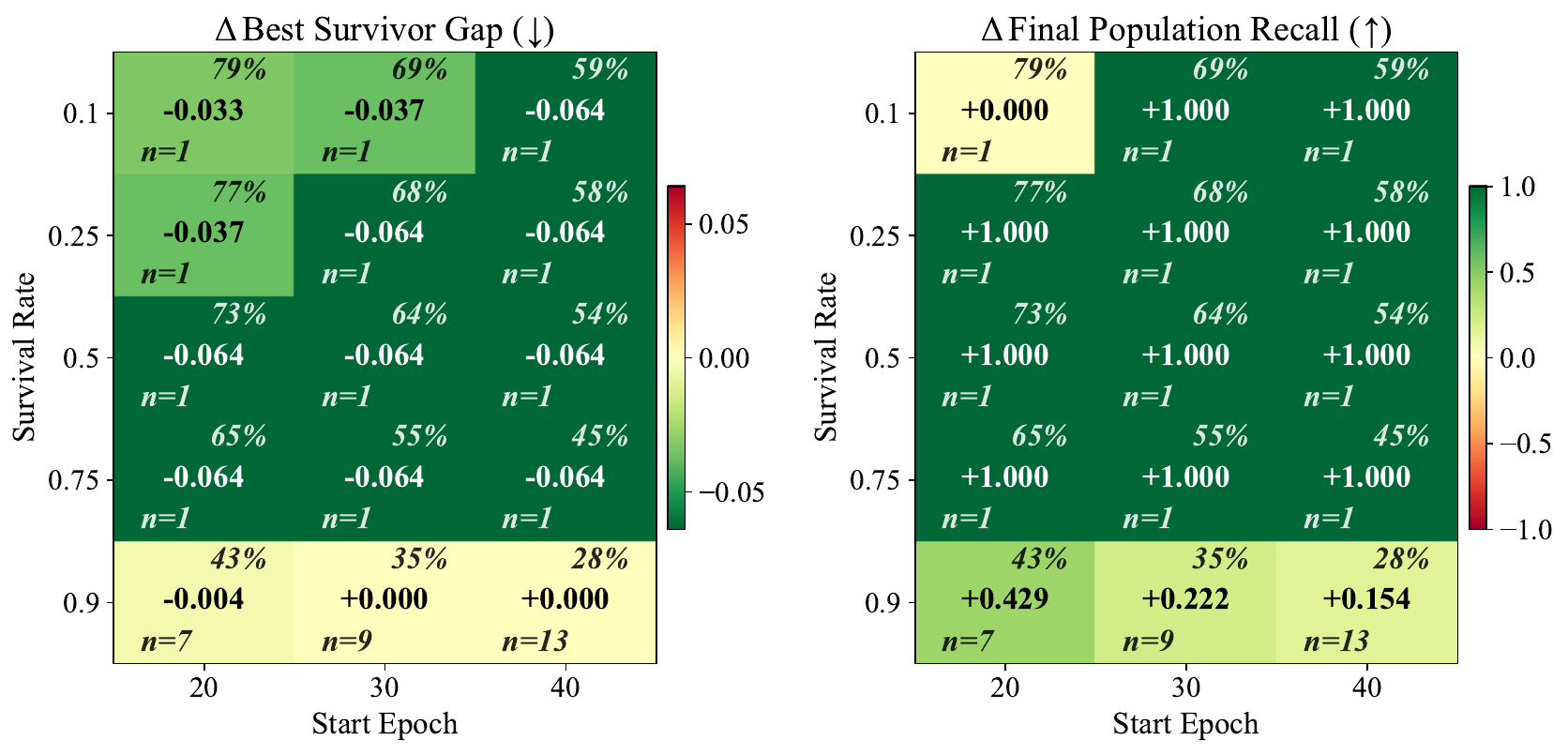}
  \caption{\textbf{SOTF training sensitivity on the Tiny ImageNet ResNet-18 zoo.} The latent-token
  Transformer, trained beforehand and held fixed throughout SOTF training, is compared with learning-curve extrapolation for final-value
  nowcasting. Negative gap differences and positive recall differences favor SOTF training. Each
  cell additionally reports compute savings (top right) and final population size $n$
  (bottom left). ``Best survivor gap'' denotes the best-retained-model gap defined in
  Equation~\ref{eq:best-retained-gap}.}
  \label{fig:sotf-tinyimagenet}
\end{figure}

Observed performance remains important inside SOTF training. Across matched predictor pairs,
adding performance features reduces the gap in all but one case
(Appendix Table~\ref{tab:app-sotf-regime-full}) and is especially valuable for SANE predictors
in horizon mode. Figure~\ref{fig:sotf-tinyimagenet} shows that the aggregate improvement is not obtained
by sacrificing efficiency. On the Tiny ImageNet ResNet-18 zoo, the latent-token Transformer reduces the gap relative to learning-curve
extrapolation by $0.033$--$0.064$ in all 12 configurations with retention rates between $0.10$ and $0.75$,
while saving $45\%$--$79\%$ of training epochs. Oracle recall improves in 11 of the 12 configurations.
At retention rate $0.90$, the advantage largely disappears as the policy keeps more models and compute savings fall.
Against LCNet, however, the Transformer is not better on this zoo (mean gap $0.0065$ vs.\ $0.0050$;
Appendix Table~\ref{tab:app-sotf-per-zoo}), unlike averaged-latent XGBoost and the latent-window LSTM.

The useful schedule remains population-specific. On the CIFAR-10 width-varying
ResNet-18 zoo (Figure~\ref{fig:app-sens-15}), the latent-window LSTM helps mainly when pruning starts at epoch 40 or
60, where it reduces the gap relative to learning-curve extrapolation by $0.037$ in five of ten
configurations while saving $58\%$--$72\%$ of training epochs. Starting at epoch 20 yields no
benefit at retention rates up to $0.50$ and smaller gains of $0.014$ at higher ones. Averaged over
all configurations, its gap ($0.0033$) matches LCNet ($0.0031$). The CIFAR-10 ViT zoo exhibits the opposite extreme.
At start epoch 20 and retention rates $0.10$, $0.25$, and $0.50$,
weight-statistic XGBoost reduces the final-value gap relative to learning-curve
extrapolation by $0.702$, $0.610$, and $0.651$, respectively, while saving
$89\%$--$92\%$ of training epochs. Its horizon results are substantially less stable.
SOTF training can therefore yield large quality--efficiency gains, but only when its nowcasting
target and pruning schedule match the learning dynamics of the candidate population.


\section{Discussion and limitations}
\label{sec:discussion}

Our results support performance nowcasting from neural network checkpoints, but also
delimit when it is useful. Checkpoint representations complement rather than replace the
learning curve, which the feature ablations reflect: hybrid predictors are usually
strongest or close to it. The two prediction targets differ similarly. Over a five-epoch
horizon, recent metrics and compact weight statistics already give a robust local
continuation signal, whereas predicting the end of training requires longer-range
extrapolation, where preserving the window- or token-level structure of SANE
representations helps. No single representation dominates, so the prediction horizon and
the source of population heterogeneity should guide the choice of checkpoint encoding.
The learning-dynamics and pruning results further show that rank instability alone does
not justify a more complex policy: in homogeneous CNN zoos, most rank reversals occur
between nearly interchangeable models, and current accuracy already retains one close to
the population optimum. SOTF training instead pays off when intermediate and final
rankings diverge meaningfully and checkpoint features can recover that difference, as in
the heterogeneous ResNet-18 and ViT populations -- though even there the useful start
epoch, retention rate, and prediction target remain population-specific.
This also explains why regression and selection results need not agree. RMSE weights
every prediction error, whereas pruning depends mainly on ordering near the retention
boundary and penalizes eliminating a future winner heavily, so decision-aware measures such as the
best-retained-model gap and oracle top-$K$ recall remain necessary complements to
standard regression metrics. Uncertainty-aware variants of SOTF training could better
reflect this asymmetric cost by postponing decisions for candidates whose predicted
ranks remain ambiguous.

Several limitations constrain these findings. The nine zoos cover only image
classification with a limited set of CNN, ResNet-18, and Vision Transformer populations,
which limits statistical power for high-capacity sequence predictors and does not
establish generalization to larger architectures, other modalities, optimizer families,
or longer training horizons. SOTF training is also evaluated retrospectively on
completed trajectories: this enables controlled counterfactual comparison but does not
capture scheduling, checkpoint I/O, variable per-epoch costs, or wall-clock contention in
a live system, so saved epochs are only a proxy for compute and energy use, excluding the
cost of constructing weight statistics and of training and applying SANE. Since SANE
backbones are trained separately per zoo, the experiments demonstrate the value of
zoo-specific representations rather than a reusable, transferable encoder, and these
costs could outweigh the savings for small populations even if amortized in larger or
recurring workloads. In zoos whose members differ mainly in width or training
hyperparameters, weight statistics may also partly encode the configuration itself: on the
width-varying ResNet-18 zoo, weight-statistic XGBoost recovers the oracle top-$K$ almost
perfectly (recall $0.99$). A baseline that receives the configuration directly is needed to
separate these effects. Finally, we do not repeat experiments across multiple random seeds,
so we cannot quantify variability from initialization, data ordering, or predictor
training; the
main study also fixes a five-epoch horizon over a finite grid of start epochs and
retention rates. Future work should repeat the evaluation with several seeds, and the
strong sensitivity to horizon and schedule choices motivates adaptive schedules that jointly account for prediction uncertainty, expected
accuracy loss, and the actual marginal cost of continuing each candidate.

\section{Conclusion}
\label{sec:conclusion}

We studied whether intermediate neural-network checkpoints contain predictive
signal about future model quality beyond the observed learning curve, and
whether that signal improves pruning decisions in a population of training
runs. Across nine checkpointed model zoos, the answer is yes, but conditionally.
Weight statistics provide the most reliable short-horizon continuation signal,
whereas structured SANE representations are more useful for longer-range
final-value extrapolation. The strongest predictors usually combine weight-space embeddings
with observed performance rather than relying on either source alone. In the SOTF training
experiments, the benefit depends on the population. In the CNN zoos, current accuracy
already retains a near-optimal model. In the heterogeneous ResNet-18 and ViT zoos, where
current performance can be misleading, ranking by nowcasted final accuracy reduces the mean
best-retained-model gap from $0.064$ to $0.0017$. The learning curve accounts for most of
this reduction, while weight statistics further reduce the gap and raise oracle top-$K$
recall from $0.20$ to $0.72$. The potential positive societal impact of this work lies primarily in
improving the efficiency of model development. If training runs can be ranked
more reliably before completion, it may avoid unnecessary compute,
energy use, and associated emissions, while making large-scale experimentation
more accessible to smaller research groups with limited resources. 

\newpage
\printbibliography

\newpage
\appendix
\raggedbottom

\section{Model zoo corpus}
\label{app:model-zoo-corpus}

Table~\ref{tab:app-model-zoo-corpus} lists the nine model zoos used throughout the paper,
together with the dataset name used to refer to each zoo, the architecture family, how the
population was generated, the number of models, and the training-epoch horizon. \emph{Uniform}
denotes a zoo with fixed hyperparameters and varying random seeds. \emph{Hyp. random} follows
the Hyp-10-rand setting of Sch\"urholt et al.~\citep{schurholtModelZoosDataset2022}: a grid of
hyperparameter configurations is repeated with ten randomly drawn seeds per grid point.
\emph{Hyp. fixed} follows their Hyp-10-fix setting, using the same hyperparameter grid but
repeating each grid point with the fixed seeds $1,\ldots,10$. \emph{Phase trans.} denotes
phase-transition zoos in which model width is varied to span under- and over-parameterized
regimes~\citep{schurholtModelZooPhase2025}. Splits are defined at the model level, so all
checkpoints belonging to one trajectory fall into exactly one of the training, validation, or
test partitions described in Section~\ref{sec:method}; the pipeline does not fix a single
split ratio shared across zoos, so exact per-zoo split sizes are not reported here.

\begin{table}[H]
    \centering
    \footnotesize
    \caption[Model zoo corpus]{\textbf{Model zoo corpus.} The dataset name is used to refer to
    the model zoo throughout the paper. Architecture abbreviations: CNN (l) for large
    convolutional networks, RN-18 width for ResNet-18 populations with varying width, and ViT
    for Vision Transformers.}
    \label{tab:app-model-zoo-corpus}
    \begingroup
    \setlength{\tabcolsep}{3pt}
    \resizebox{\textwidth}{!}{%
    \begin{tabular}{@{}l l l l r r@{}}
        \toprule
        \textbf{Dataset name} & \textbf{Image data} & \textbf{Architecture} & \textbf{Population} & \textbf{Models} & \textbf{Epochs} \\
        \midrule
        cifar10\_cnn\_large\_uniform & CIFAR-10 & CNN (l) & Uniform & 1000 & 50 \\
        cifar10\_cnn\_large\_hyp\_random\_seeds & CIFAR-10 & CNN (l) & Hyp. random & 1280 & 50 \\
        cifar10\_cnn\_large\_hyp\_fixed\_seeds & CIFAR-10 & CNN (l) & Hyp. fixed & 1280 & 50 \\
        stl10\_cnn\_large\_uniform & STL-10 & CNN (l) & Uniform & 1000 & 50 \\
        stl10\_cnn\_large\_hyp\_random\_seeds & STL-10 & CNN (l) & Hyp. random & 2560 & 50 \\
        stl10\_cnn\_large\_hyp\_fixed\_seeds & STL-10 & CNN (l) & Hyp. fixed & 2560 & 50 \\
        tinyimagenet\_resnet18\_phase\_transitions & Tiny ImageNet & RN-18 width & Phase trans. & 192 & 100 \\
        cifar10\_resnet18\_width\_2-256 & CIFAR-10 & RN-18 width & Phase trans. & 192 & 150 \\
        cifar10\_vit & CIFAR-10 & ViT & Phase trans. & 147 & 300 \\
        \bottomrule
    \end{tabular}%
    }
    \endgroup
\end{table}

\section{Learning dynamics}
\label{app:learning-dynamics}

\begin{figure}[H]
      \centering
    \includegraphics[width=\linewidth]{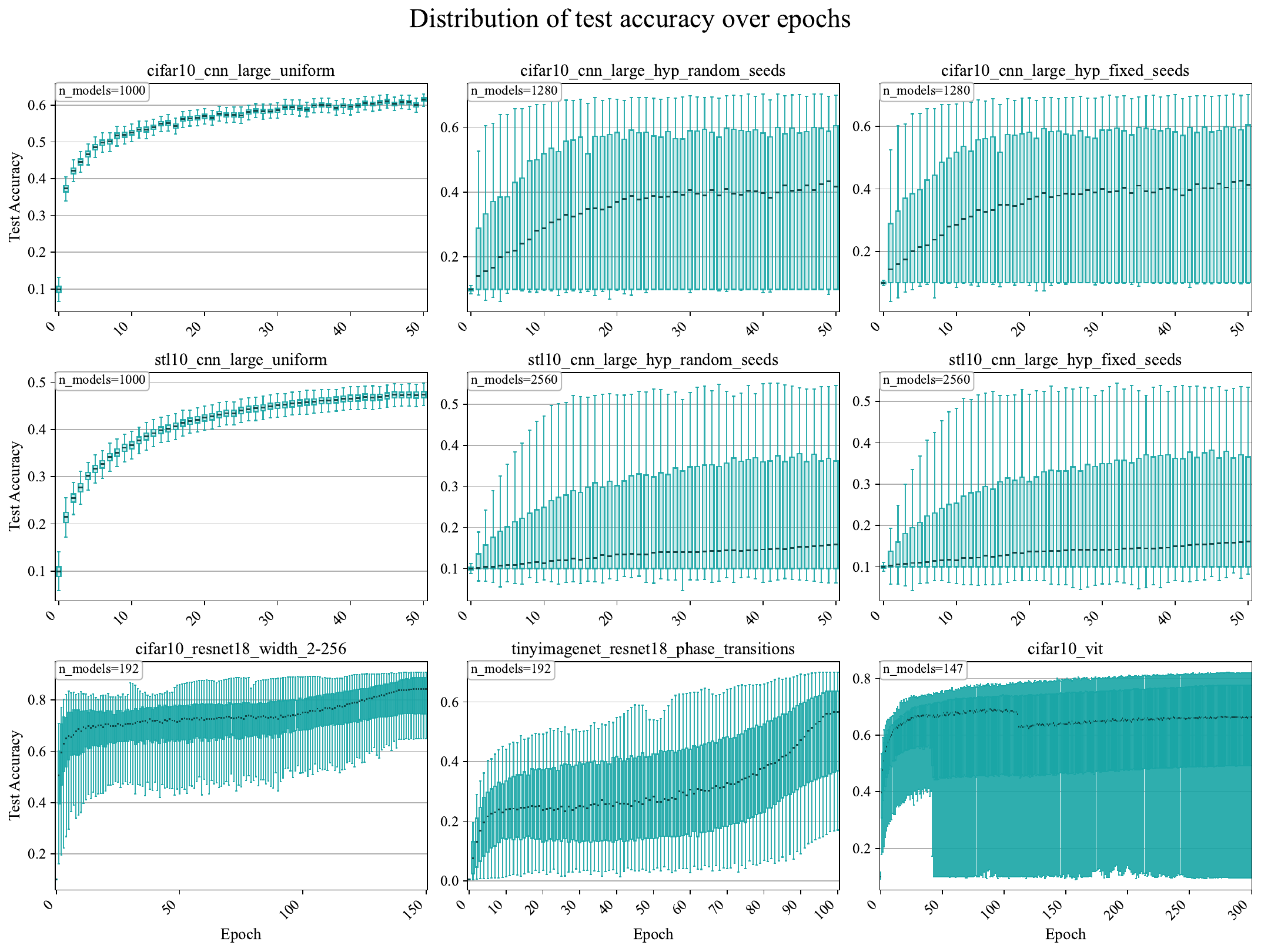}
    \caption[Accuracy distributions across all model zoos]{\textbf{Accuracy distributions across all model zoos.}
    Test-accuracy distributions at each checkpoint for all nine model zoos. Boxes summarize
    the population-level spread and reveal the contrast between narrow uniform zoos and
    heterogeneous architecture or width variants.}
    \label{fig:accuracy-distributions-all}
\end{figure}

\begin{figure}[H]
    \centering
    \includegraphics[width=0.75\linewidth]{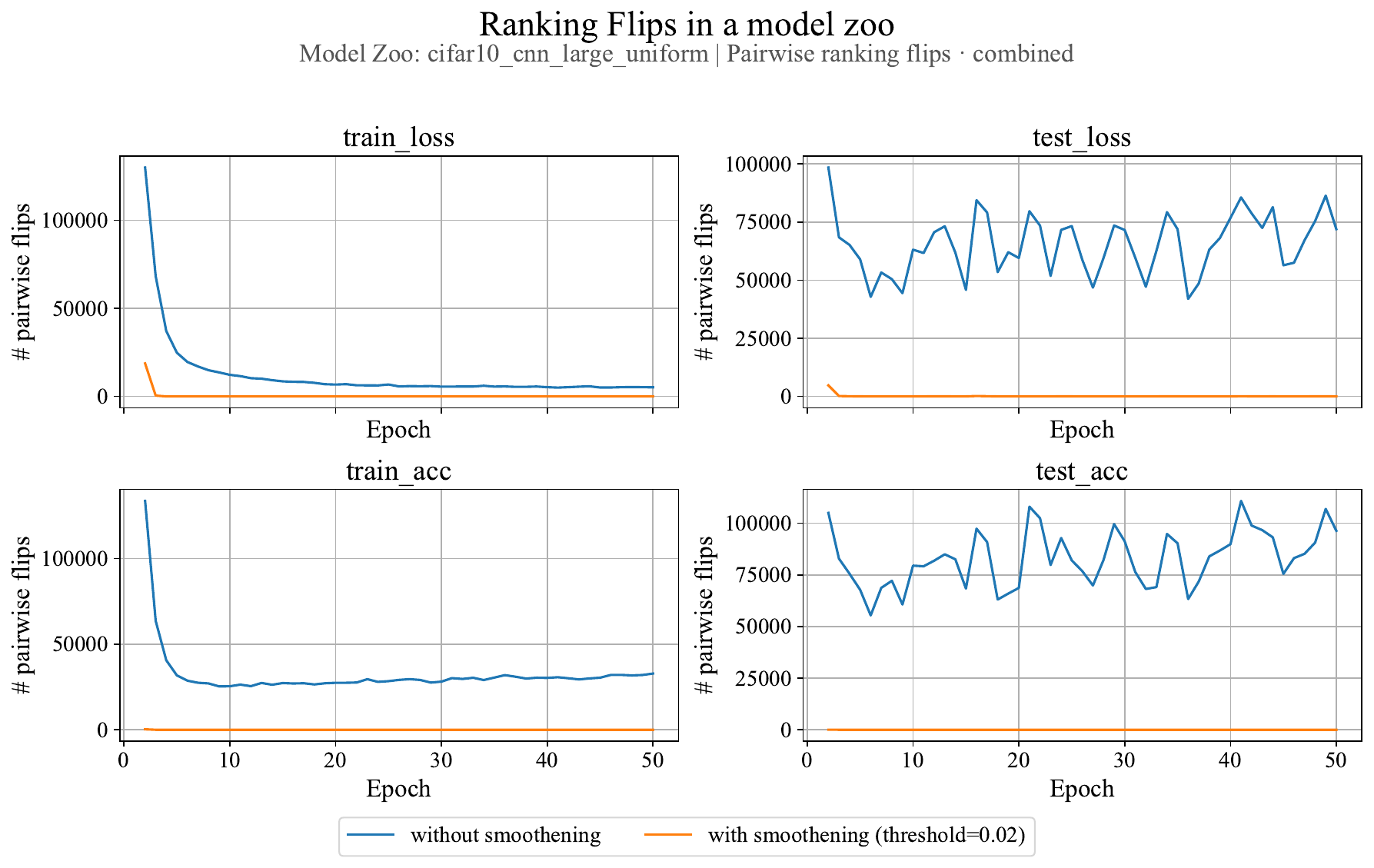}
    \caption{\textbf{Pairwise rank flips --- CIFAR-10 uniform CNN zoo.} Raw and smoothed pairwise rank flips between consecutive epochs.}
    \label{fig:app-flips-ds1}
\end{figure}

\begin{figure}[H]
    \centering
    \includegraphics[width=0.75\linewidth]{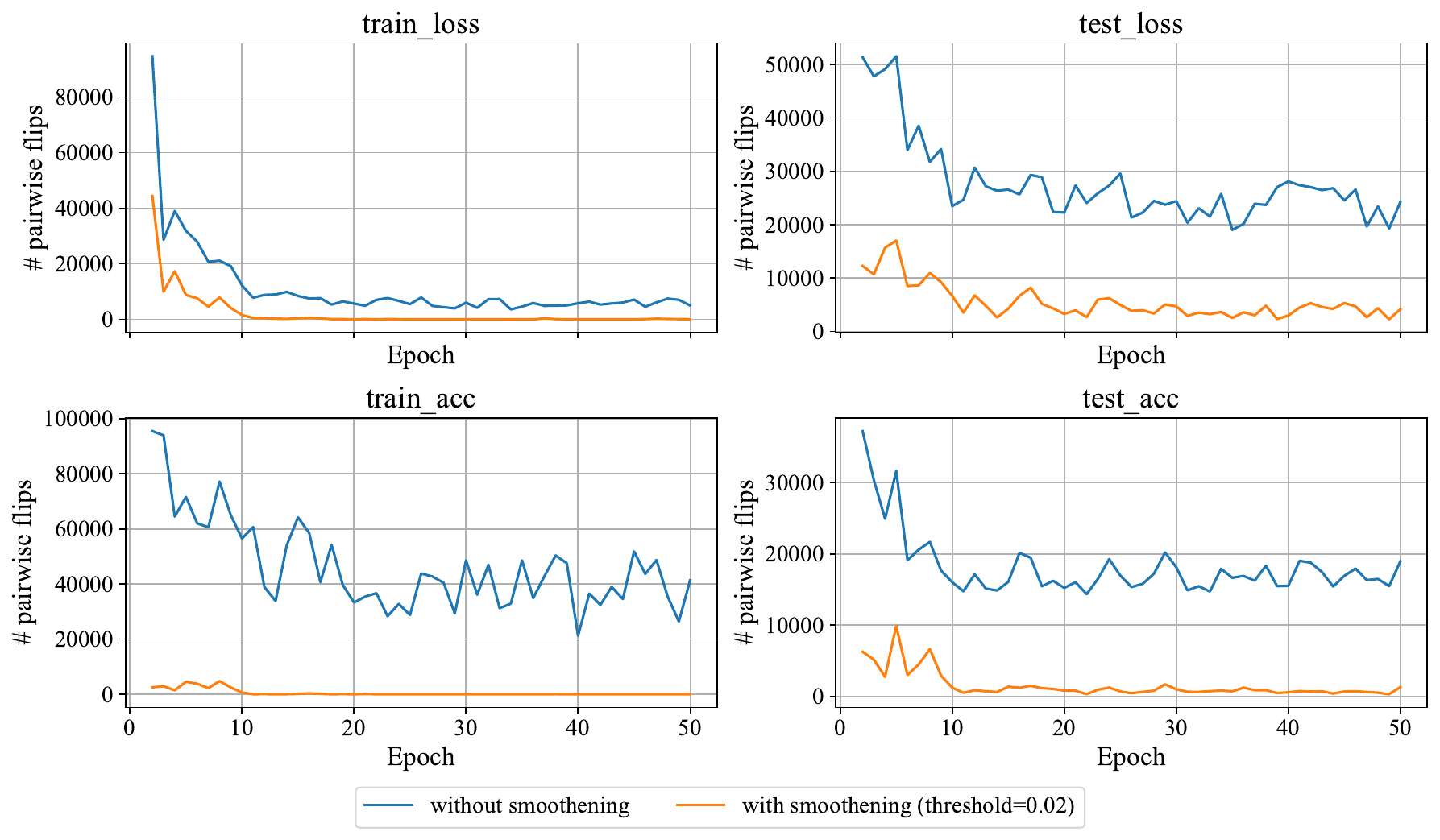}
    \caption{\textbf{Pairwise rank flips --- CIFAR-10 CNN Hyp-10-rand.} Raw and smoothed pairwise rank flips between consecutive epochs.}
    \label{fig:app-flips-ds2}
\end{figure}

\begin{figure}[H]
    \centering
    \includegraphics[width=0.75\linewidth]{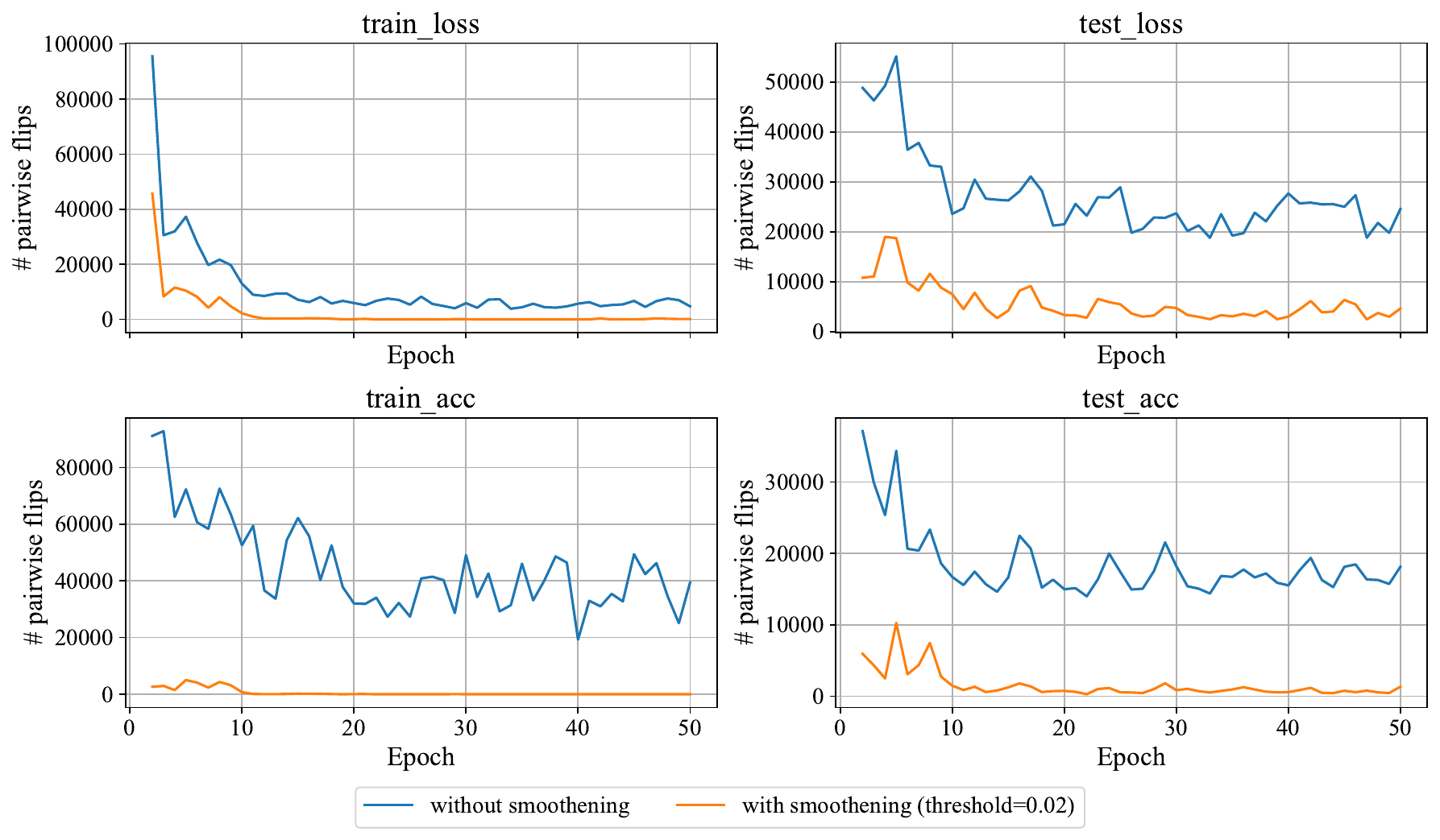}
    \caption{\textbf{Pairwise rank flips --- CIFAR-10 CNN Hyp-10-fix.} Raw and smoothed pairwise rank flips between consecutive epochs.}
    \label{fig:app-flips-ds3}
\end{figure}

\begin{figure}[H]
    \centering
    \includegraphics[width=0.75\linewidth]{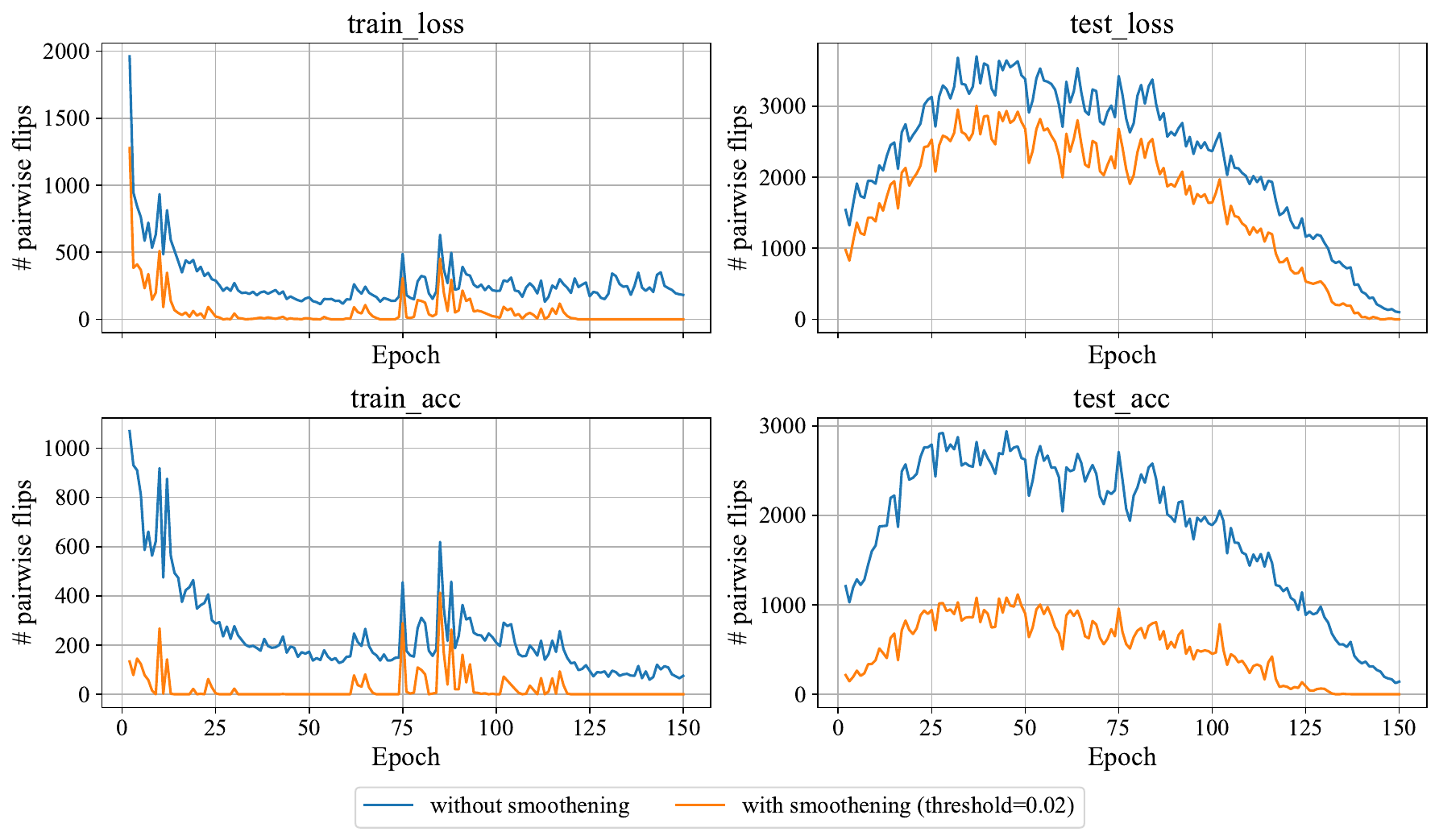}
    \caption{\textbf{Pairwise rank flips --- CIFAR-10 ResNet-18 width zoo.} Raw and smoothed pairwise rank flips between consecutive epochs.}
    \label{fig:app-flips-ds5}
\end{figure}

\begin{figure}[H]
    \centering
    \includegraphics[width=0.75\linewidth]{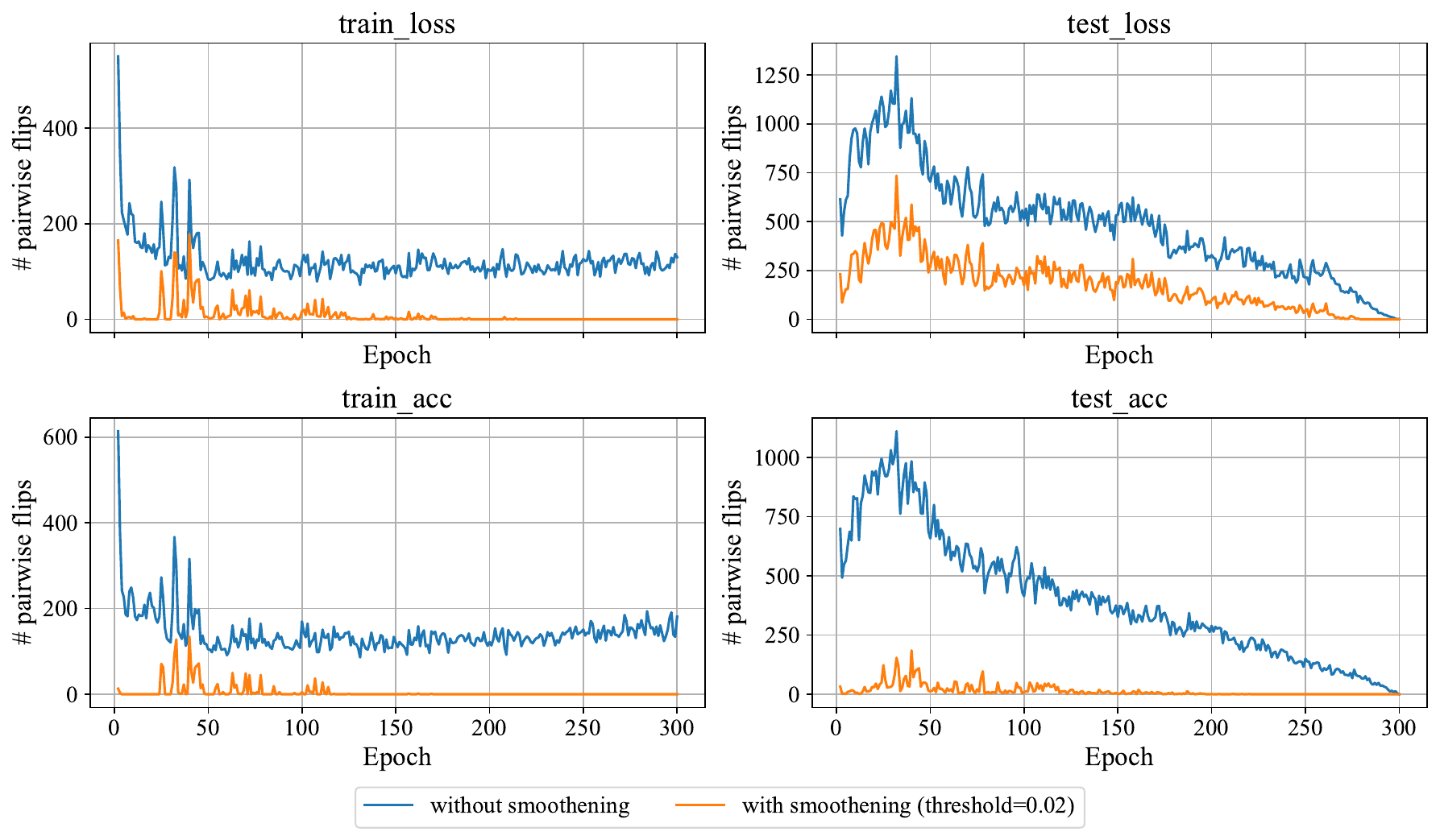}
    \caption{\textbf{Pairwise rank flips --- CIFAR-10 ViT zoo.} Raw and smoothed pairwise rank flips between consecutive epochs.}
    \label{fig:app-flips-ds7}
\end{figure}

\begin{figure}[H]
    \centering
    \includegraphics[width=0.75\linewidth]{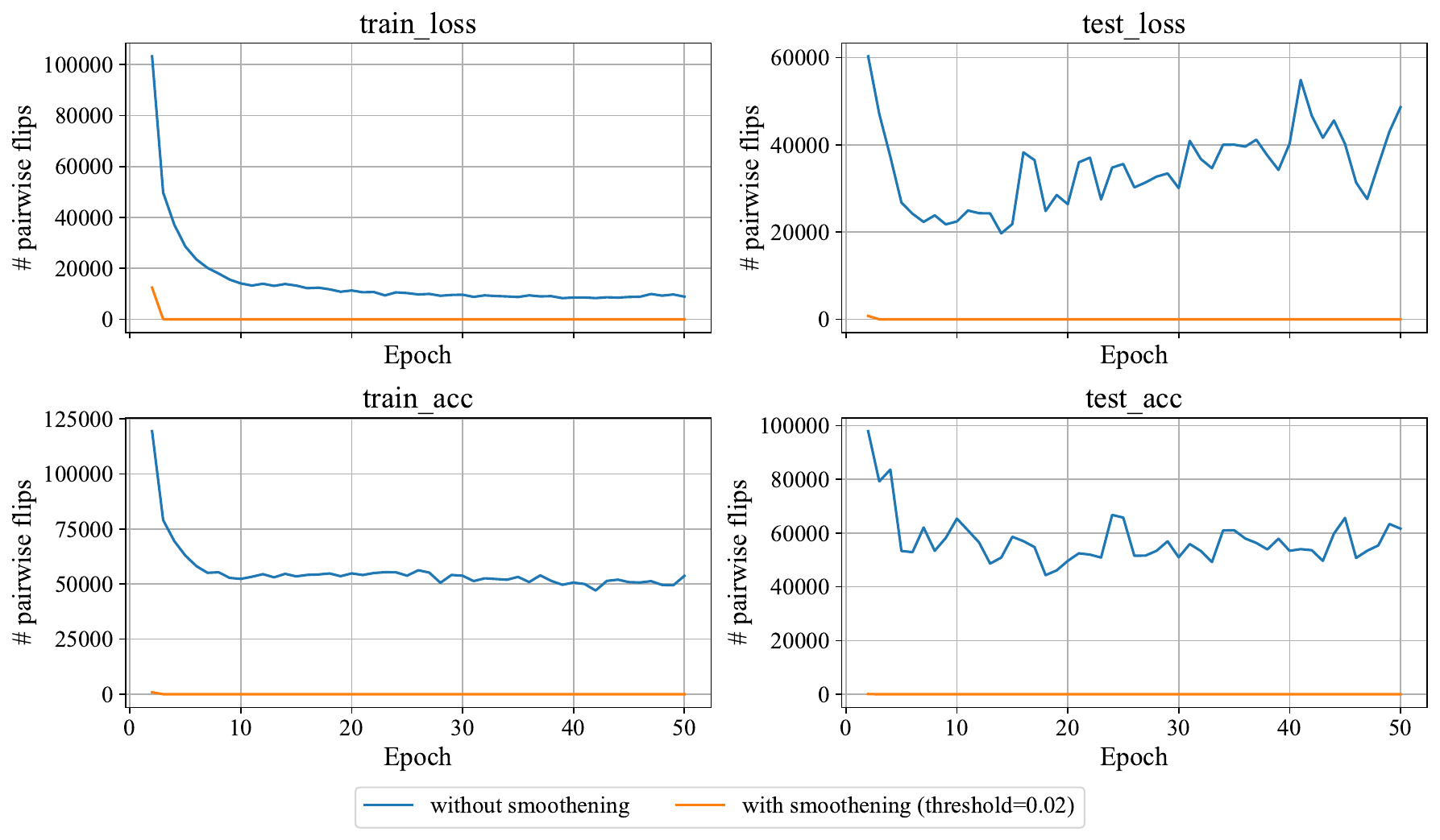}
    \caption{\textbf{Pairwise rank flips --- STL-10 uniform CNN zoo.} Raw and smoothed pairwise rank flips between consecutive epochs.}
    \label{fig:app-flips-ds8}
\end{figure}

\begin{figure}[H]
    \centering
    \includegraphics[width=0.75\linewidth]{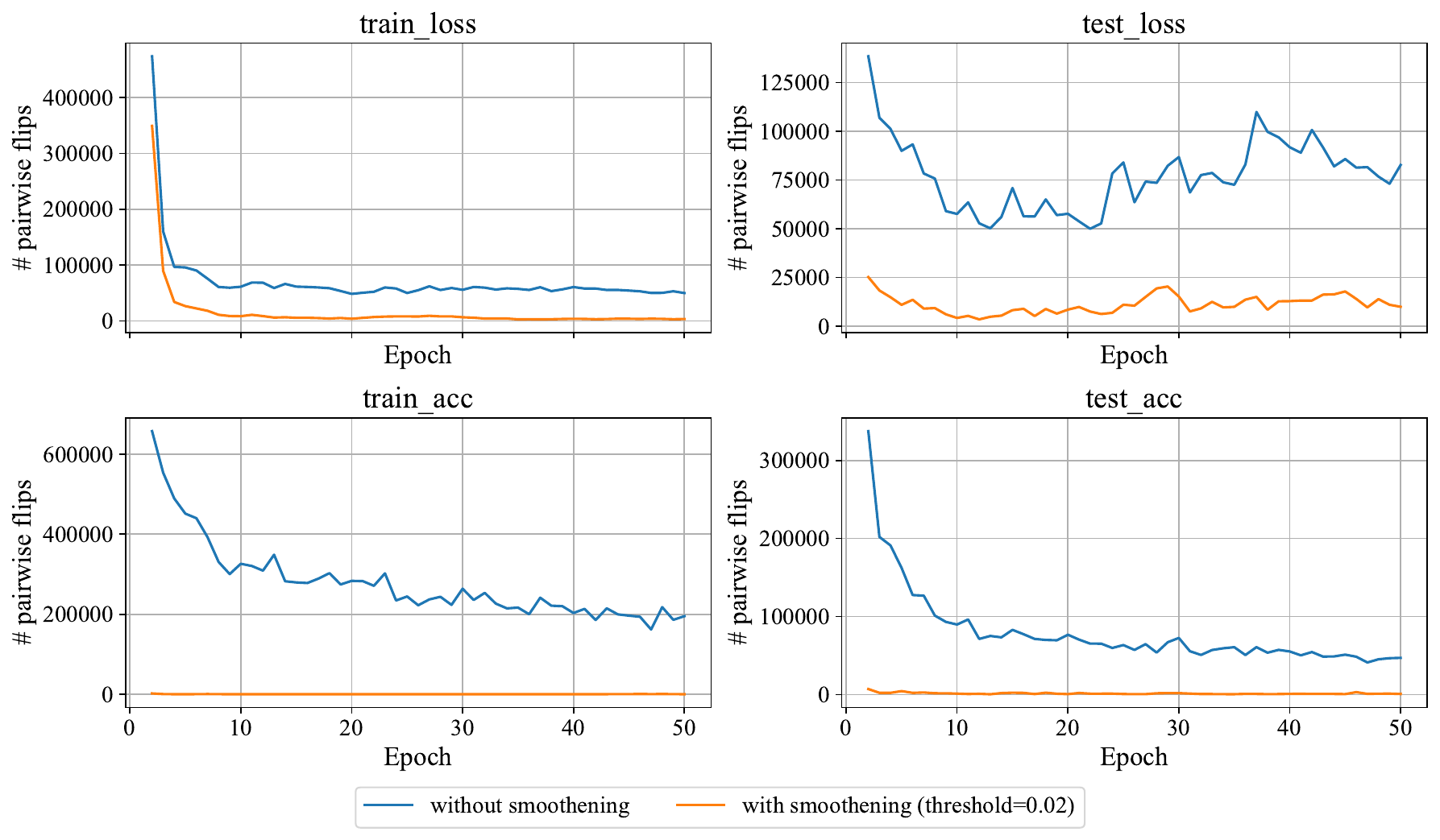}
    \caption{\textbf{Pairwise rank flips --- STL-10 CNN Hyp-10-rand.} Raw and smoothed pairwise rank flips between consecutive epochs.}
    \label{fig:app-flips-ds9}
\end{figure}

\begin{figure}[H]
    \centering
    \includegraphics[width=0.75\linewidth]{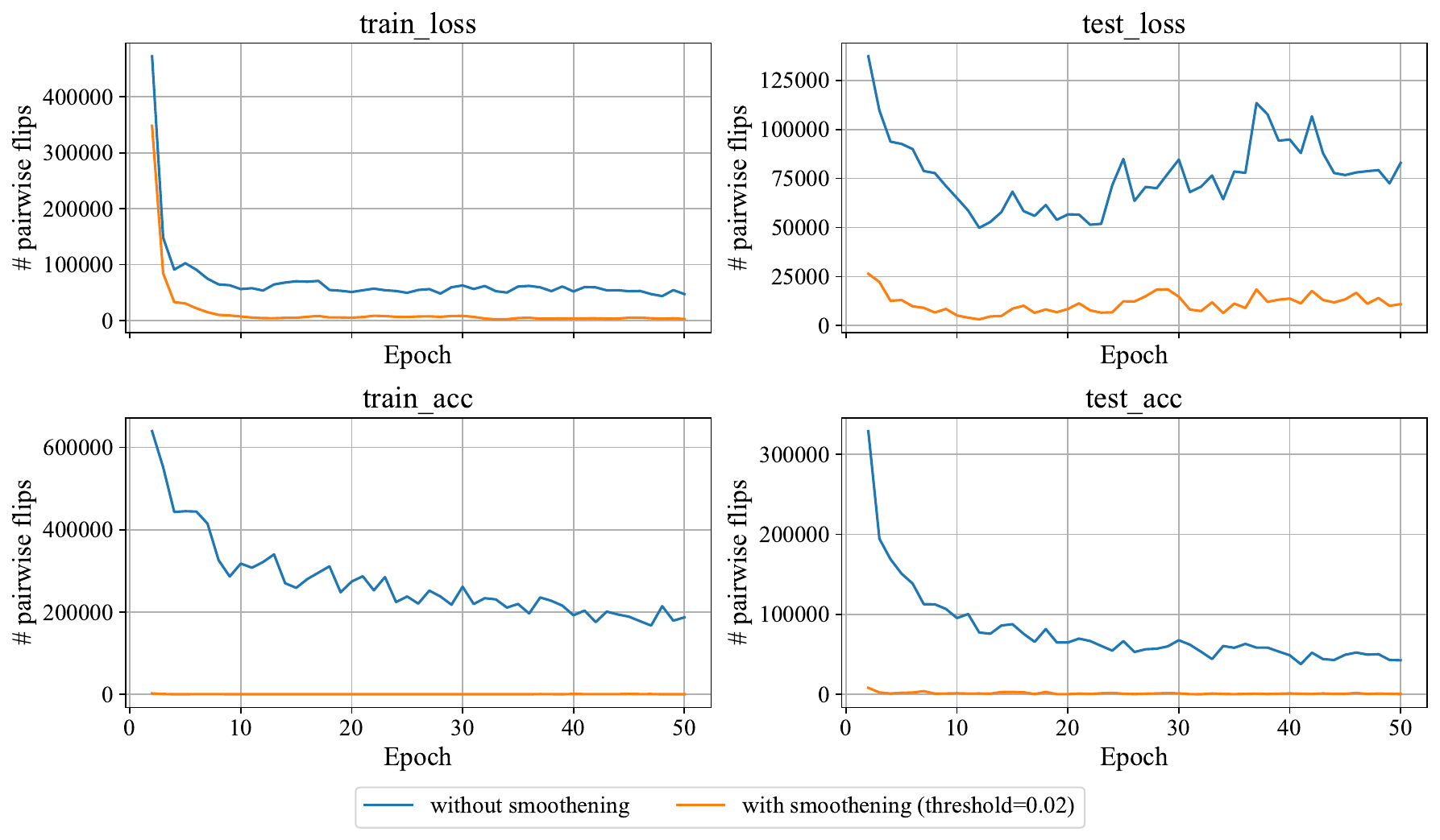}
    \caption{\textbf{Pairwise rank flips --- STL-10 CNN Hyp-10-fix.} Raw and smoothed pairwise rank flips between consecutive epochs.}
    \label{fig:app-flips-ds10}
\end{figure}

Section~\ref{sec:results} discusses ranking stability in detail for the Tiny ImageNet
ResNet-18 phase-transition zoo (Figure~\ref{fig:rn18_tiniyimagenet_co_kendall}) and the
STL-10 uniform CNN zoo (Figure~\ref{fig:app-flips-ds8}). Figures~\ref{fig:app-flips-ds1}--\ref{fig:app-flips-ds10}
report the same raw and $\epsilon$-smoothed pairwise rank-flip diagnostic for all eight zoos
other than Tiny ImageNet, so that the rank-instability evidence underlying
Section~\ref{sec:results} can be inspected for every zoo in the corpus.

\section{Weight-statistic features}
\label{app:weight-statistics}

Table~\ref{tab:app-weight-statistic-families} lists the four families of checkpoint-derived
weight statistics summarized in Section~\ref{subsec:method-predictors-sotf}. Statistics are
computed at both the full-model and the per-layer level; parameters are grouped by layer name,
pairing a weight tensor with its bias tensor where available, and depending on the statistic
are computed on weights only, biases only, the combined weight-and-bias group, or the complete
flattened checkpoint. Layer-level norms are aggregated to model level (e.g., the model $\ell_2$
norm is the square root of the sum of squared layer $\ell_2$ norms); matrix statistics are
aggregated using their mean and maximum across layers.

\begin{table}[H]
    \centering
    \small
    \caption{Checkpoint-derived weight-statistic families.}
    \label{tab:app-weight-statistic-families}
    \begin{tabularx}{\textwidth}{@{}l X X@{}}
        \toprule
        \textbf{Family} & \textbf{Included statistics} & \textbf{Captured property} \\
        \midrule
        Norm & $\ell_1$, $\ell_2$, and $\ell_\infty$ norms & Overall parameter scale and the presence of large individual values. \\
        Distribution & Mean, variance, percentiles, and range & Center, spread, and tails of the flattened parameter distribution. \\
        Matrix & Frobenius norm, mean and maximum absolute value, root mean square, dimensions, aspect ratio, and row- and column-wise norms & Layer shape, scale, and variation across input and output directions. \\
        Distance & Distance and cosine similarity to initialization and to the previous checkpoint, cumulative path length, and path straightness & Movement of the full model through parameter space during training. \\
        \bottomrule
    \end{tabularx}
\end{table}

The distance family is defined relative to the initial checkpoint $\theta_{i,e_0}$ and the
previous saved checkpoint $\theta_{i,e_q^-}$:
\begin{equation}
  d_{\mathrm{init}}(i,e_q) = \lVert\theta_{i,e_q}-\theta_{i,e_0}\rVert_2, \qquad
  d_{\mathrm{prev}}(i,e_q) = \lVert\theta_{i,e_q}-\theta_{i,e_q^-}\rVert_2,
  \label{eq:app-distance-features}
\end{equation}
\begin{equation}
  L(i,e_t) = \sum_{q=1}^{t}\lVert\theta_{i,e_q}-\theta_{i,e_q^-}\rVert_2, \qquad
  r_{\mathrm{straight}}(i,e_t) = \frac{d_{\mathrm{init}}(i,e_t)}{L(i,e_t)+\varepsilon},
  \label{eq:app-path-straightness}
\end{equation}
where $L$ is the cumulative path length through weight space and $r_{\mathrm{straight}}$ is
close to one for a comparatively direct trajectory and smaller for a less direct one. The
regression experiments in Section~\ref{sec:method} use the distribution family as the
compact, architecture-agnostic weight-statistic representation; the norm, matrix, and
distance families are computed by the same pipeline and used in the descriptive
learning-dynamics analysis.

\section{Predictor configurations and baseline specifications}
\label{app:predictor-configurations}

This appendix reports the fixed configuration used for each predictor family in the main
experiments, and the exact specification of the two learning-curve-only baselines.

\begin{table}[H]
    \centering
    \small
    \caption{Ridge and XGBoost configurations.}
    \label{tab:app-ridge-xgboost-configuration}
    \begin{tabularx}{\textwidth}{@{}l X X@{}}
        \toprule
        \textbf{Parameter} & \textbf{Ridge} & \textbf{XGBoost} \\
        \midrule
        Objective / penalty & Squared error $+\ \ell_2$ & Squared error \\
        Regularization strength & $\eta_{\mathrm{ridge}}=1.0$ & $\ell_2$ leaf reg.\ $=1.0$, $\ell_1$ leaf reg.\ $=0$ \\
        Boosting rounds & --- & 400 \\
        Maximum tree depth & --- & 6 \\
        Learning rate & --- & 0.05 \\
        Row / feature subsampling & --- & 0.8 / 0.8 \\
        Tree construction & --- & Histogram \\
        Solver / other & Automatic solver selection, fitted intercept & Fixed random seed \\
        \bottomrule
    \end{tabularx}
\end{table}

\begin{table}[H]
    \centering
    \small
    \caption{Feed-forward and sequence predictor configurations for the SANE representation and weight-statistic MLPs.}
    \label{tab:app-nn-predictor-configuration}
    \begin{tabularx}{\textwidth}{@{}l X X X X@{}}
        \toprule
        \textbf{Parameter} & \textbf{Weight-stat.\ MLP} & \textbf{Avg.-latent MLP} & \textbf{Latent-window LSTM} & \textbf{Latent-token Transf.} \\
        \midrule
        Input dimension & Weight-statistic vector & 128 (SANE latent) & 128 (window embedding) & 128 (token embedding) \\
        Architecture & Hidden widths 256, 128, 64; LayerNorm & Hidden widths 256, 128, 64; 128, 64; or 64, 32 & LSTM hidden size 64, 1 layer & Internal dim.\ 64, 4 heads, 1 local + 2 global layers \\
        Local chunk size $w$ & --- & --- & --- & 8, 16, 32, 128, 256, or 512 \\
        Regression head & Linear readout & Linear readout & Width 32, GELU & Width 32, GELU \\
        Activation & GELU & GELU & --- & --- \\
        Dropout & 0.1 & 0 or 0.1 & 0 & 0 \\
        Positional encoding & --- & --- & --- & Sinusoidal; learned {[}CLS{]} token before global stage \\
        Optimizer & AdamW & AdamW & AdamW & AdamW \\
        Batch size & 128 & 16 & 16 & 8 or 16 \\
        Learning rate & $10^{-3}$ & $10^{-3}$ & $10^{-3}$ & $10^{-3}$ \\
        Weight decay & $10^{-4}$ & $10^{-3}$ & $10^{-3}$ & $10^{-3}$ \\
        Maximum epochs & 500 & 500 & 200 & 200 \\
        Loss & Smooth $\ell_1$ & Smooth $\ell_1$ & Smooth $\ell_1$ & Smooth $\ell_1$ \\
        \bottomrule
    \end{tabularx}
\end{table}

Larger local chunks together with a batch size of 8 are used for the longest token
sequences, while shorter sequences use smaller chunks and a batch size of 16; the local
transformer stage is skipped when a sequence is shorter than the local chunk size. Model
hyperparameters are fixed for ridge regression, XGBoost, the weight-statistic MLP, and the
latent-window LSTM; for these families, only the input feature set varies across runs (with or
without lagged checkpoints, first-order differences, layer-level statistics, and performance
features). For the averaged-latent MLP and the latent-token Transformer, the predefined
architecture variants listed above are additionally trained per zoo. In all cases, the variant
with the lowest validation RMSE is reported (Appendix~\ref{app:metric-prediction-detailed}); no
automated hyperparameter search is used. The codebase separately
exposes a Ray Tune search space for XGBoost and LightGBM (num.\ samples $=6$, 4 concurrent
trials, CPU-only) that widens $n_{\mathrm{estimators}}$, tree depth, learning rate, subsampling,
and $\ell_1$/$\ell_2$ regularization around the values above; this search capability exists in
the pipeline for exploratory use but was not the source of the fixed configurations reported
here.

\paragraph{Learning-curve extrapolation baseline.}
The baseline fits four parametric curve families to the observed accuracy prefix ---
linear, logarithmic, exponential saturation, and power-law saturation --- and predicts the
median of all successful extrapolations, clipped to the target range observed in the
corresponding training and validation split. It receives only the observed scalar learning
curve up to the current checkpoint: no weight statistics, SANE latents, architecture metadata,
or hyperparameters. This is a deterministic, point-prediction simplification of the
eleven-curve-family Bayesian extrapolation method of Domhan et al., omitting the MCMC-based
predictive uncertainty and termination rule.

\paragraph{LCNet-style baseline.}
Following Klein et al.~\citep{kleinLEARNINGCURVEPREDICTION2017}, the baseline replaces the
original Bayesian-neural-network posterior sampler and parametric basis-function layer with a
two-hidden-layer MLP (width 128, ReLU activations, dropout $p=0.1$) trained with AdamW
(learning rate $10^{-3}$, weight decay $10^{-4}$, Smooth-$\ell_1$ loss, early stopping). Its
input feature map contains: the current and target epoch; the progress fractions
$e/e_{\max}$ and $(e^{\star}-e)/e_{\max}$; the current test accuracy; the first observed test
accuracy in the prefix; the prefix mean and standard deviation; the total gain from the first
observed checkpoint to the current one; the recent slope and recent first-order accuracy
deltas; the last eight observed test-accuracy values and their epochs; and binary masks
indicating which of those eight lag positions are actually observed. At inference, dropout is
kept active for $S=30$ Monte Carlo forward passes; the mean prediction is used for RMSE and
$R^2$, and the resulting variance is retained but not scored. No weight statistics, SANE
latents, architecture metadata, or hyperparameters are provided as input.

\section{Detailed metric-prediction results by predictor family}
\label{app:metric-prediction-detailed}

This appendix reports the per-family metric-prediction results underlying the aggregate
comparison in Figure~\ref{fig:predictor-family-comparison}. For each dataset and target
mode, the configuration with the lowest validation RMSE is shown for the given family.

\begin{figure}[H]
	\centering
	\includegraphics[width=0.75\linewidth]{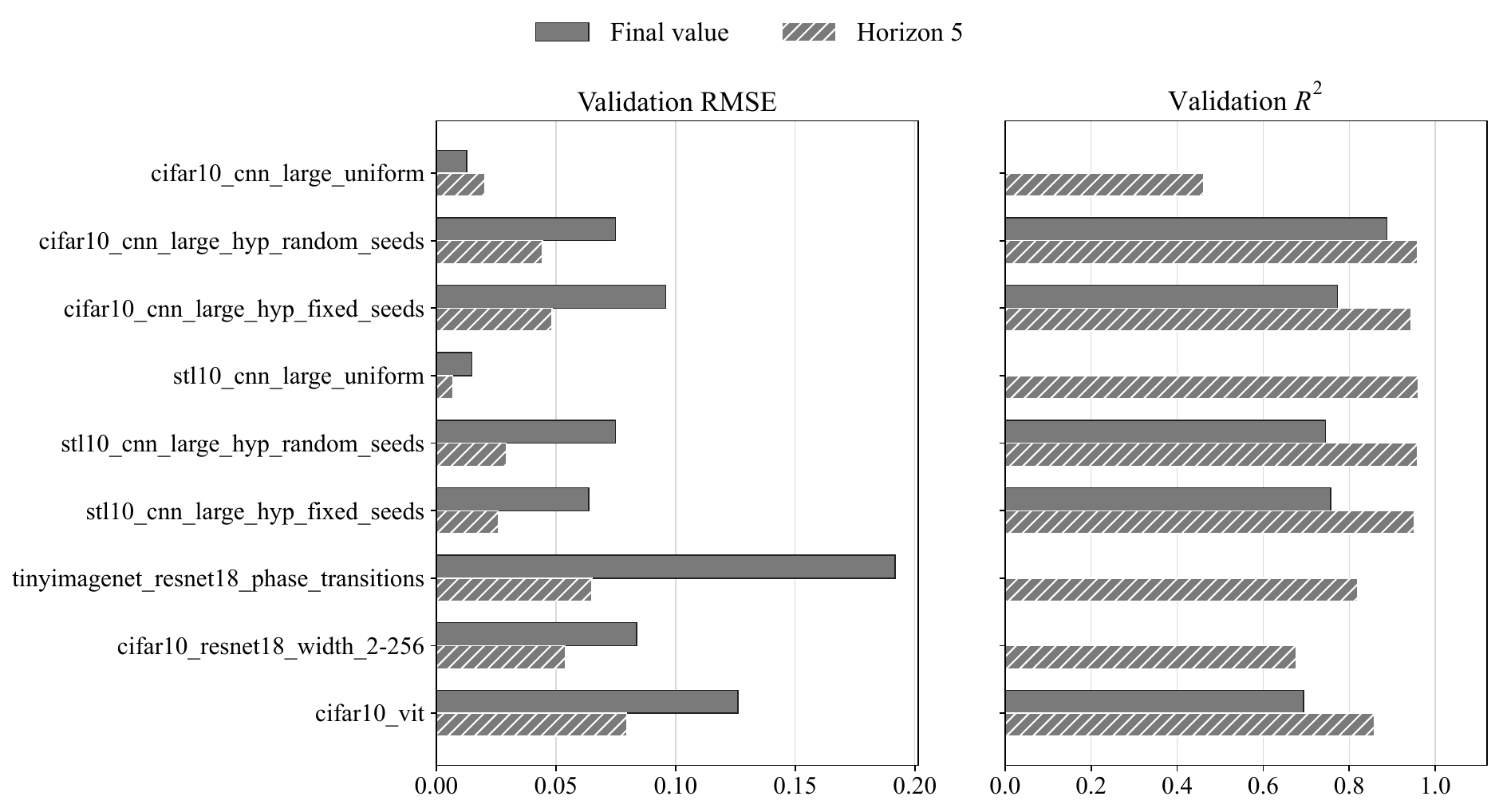}
	\caption{\textbf{Learning-curve extrapolation baseline performance.} Validation RMSE and
	$R^2$ for the parametric learning-curve extrapolation baseline on final-value and
	horizon-5 nowcasting. The $R^2$ axis is clipped at 0 for readability.}
	\label{fig:app-lc-extrapolation-performance}
\end{figure}

The baseline is strong over short horizons, reaching a median validation $R^2$ of 0.94 and a
median RMSE of 0.044 for horizon-5 nowcasting, but is less reliable for final-value
nowcasting (median $R^2=0.69$, negative mean $R^2$ because several zoos extrapolate poorly
from their observed prefixes).

\begin{figure}[H]
	\centering
	\includegraphics[width=0.75\linewidth]{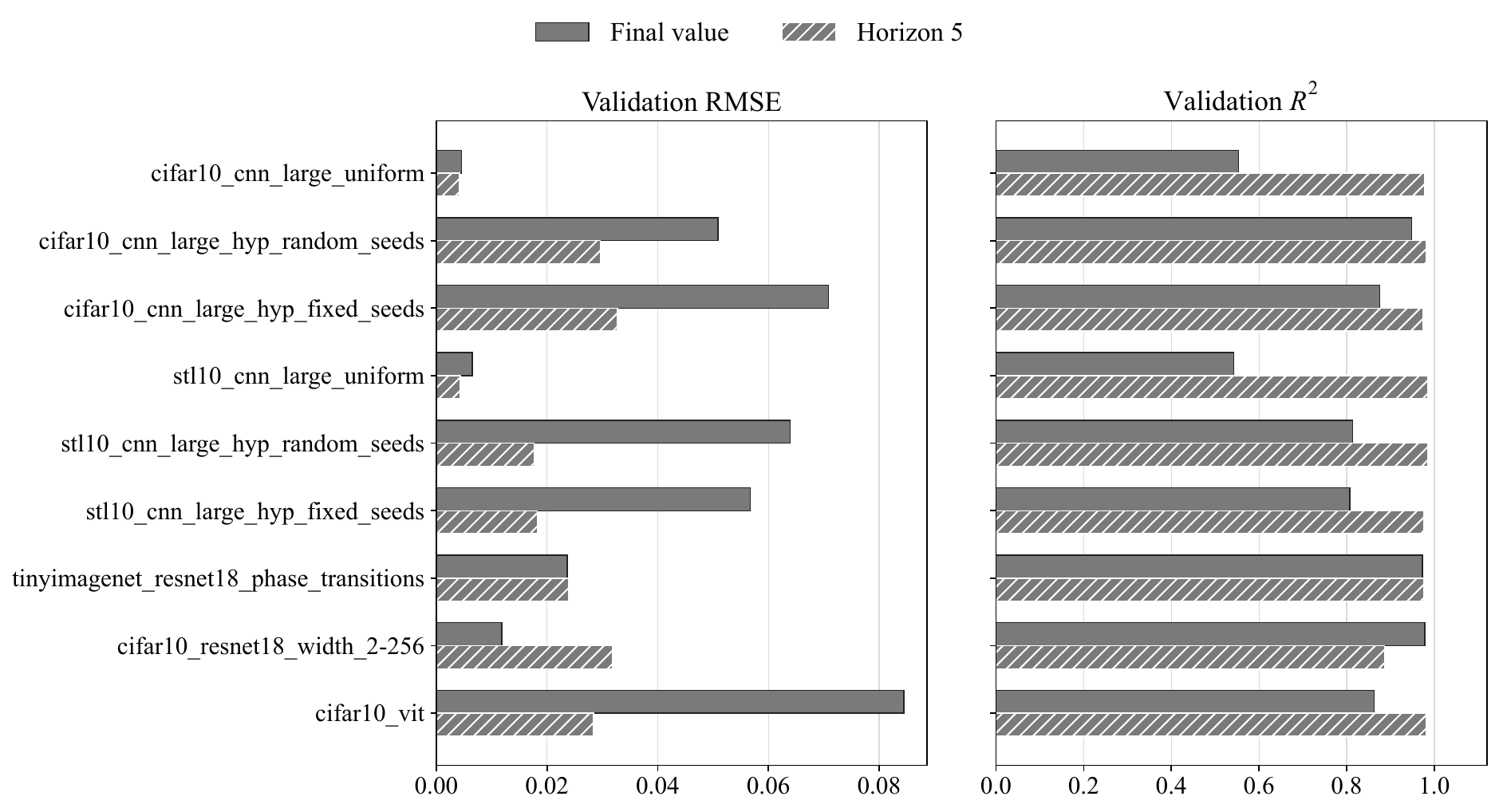}
	\caption{\textbf{LCNet learning-curve baseline performance.} Validation RMSE and $R^2$ for
	the LCNet-style baseline on final-value and horizon-5 nowcasting.}
	\label{fig:app-lcnet-performance}
\end{figure}

LCNet improves over parametric extrapolation on every dataset and target mode: median
final-value $R^2$ rises from 0.69 to 0.86 (median RMSE $0.075\rightarrow0.051$), and median
horizon-5 $R^2$ rises from 0.94 to 0.98 (median RMSE $0.044\rightarrow0.024$). All final-value
$R^2$ values are positive, making LCNet the relevant learning-curve-only reference for the
checkpoint-based predictors below.

\begin{figure}[H]
	\centering
	\includegraphics[width=0.75\linewidth]{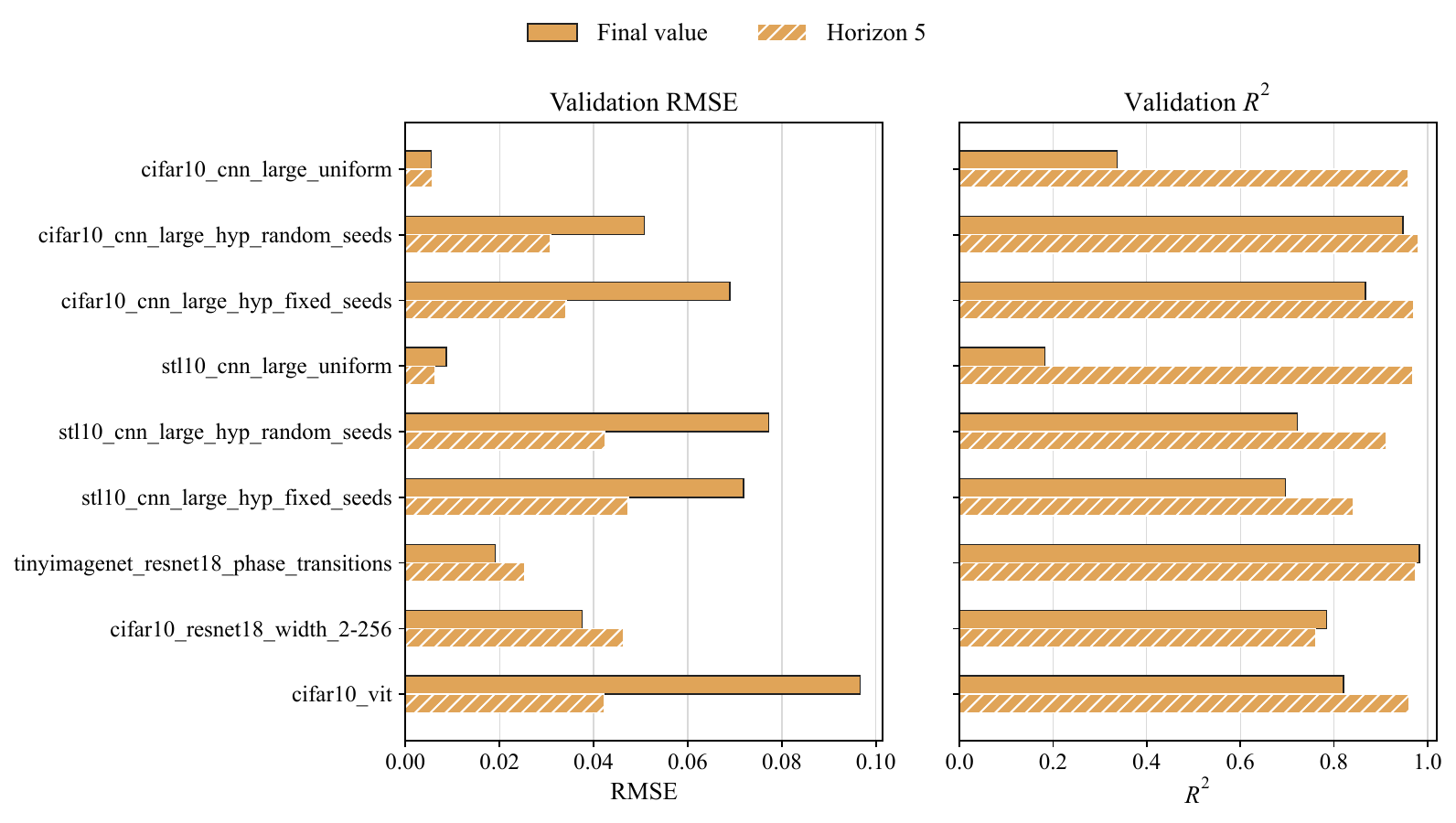}
	\caption{\textbf{Averaged-latent ridge performance by dataset.} Validation RMSE and $R^2$
	for final-value and horizon-5 nowcasting from the mean-pooled SANE embedding.}
	\label{fig:app-ridge-performance}
\end{figure}

\begin{figure}[H]
	\centering
	\includegraphics[width=0.75\linewidth]{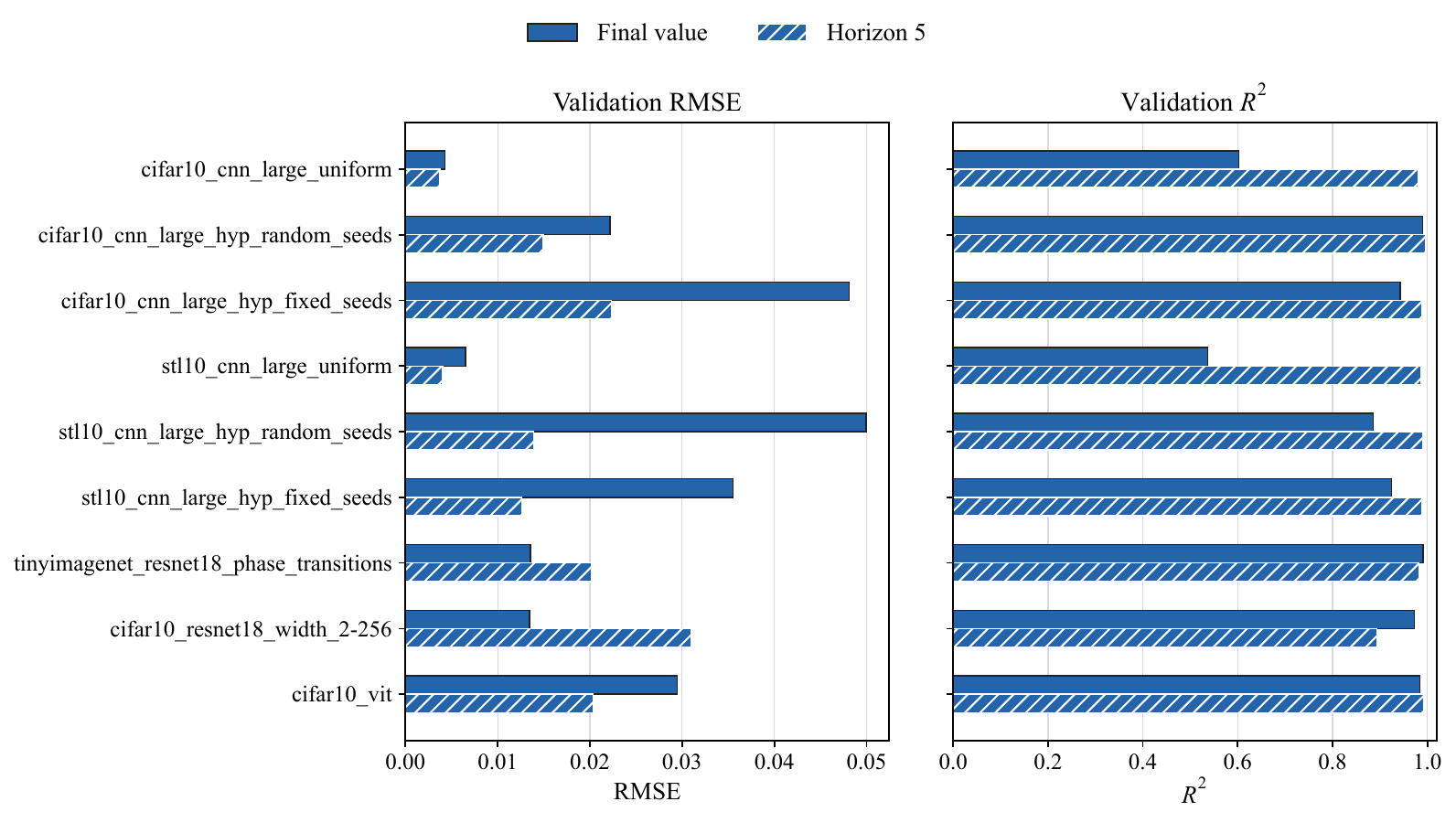}
	\caption{\textbf{Weight-statistic XGBoost performance by dataset.}}
	\label{fig:app-ws-xgboost-performance}
\end{figure}

\begin{figure}[H]
	\centering
	\includegraphics[width=0.75\linewidth]{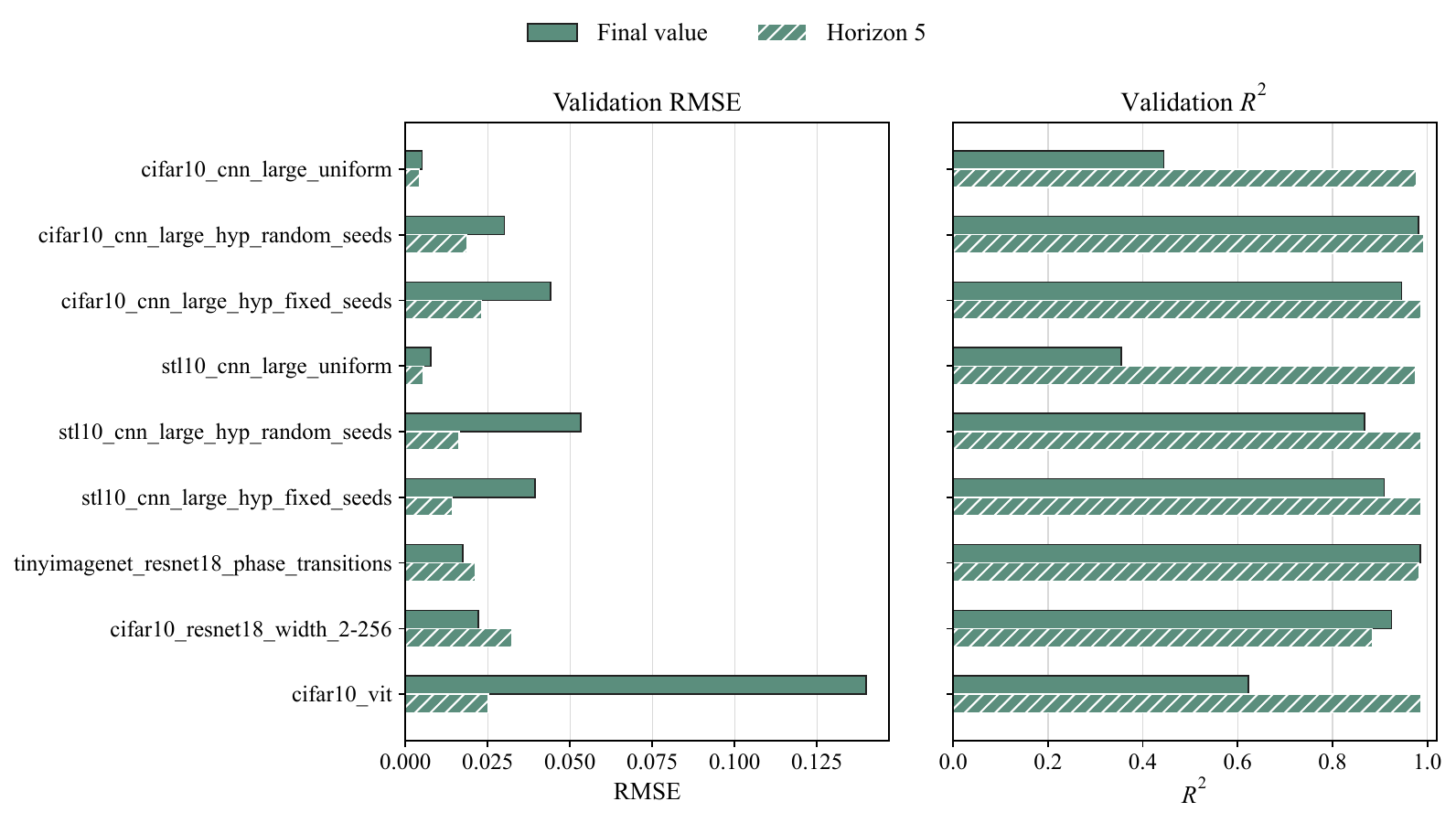}
	\caption{\textbf{Averaged-latent XGBoost performance by dataset.} The ViT zoo is a notable
	failure case: even with performance features appended, the mean-pooled checkpoint vector
	provides little reliable information about final accuracy beyond the learning-curve
	position.}
	\label{fig:app-avglat-xgboost-performance}
\end{figure}

\begin{figure}[H]
	\centering
	\includegraphics[width=0.75\linewidth]{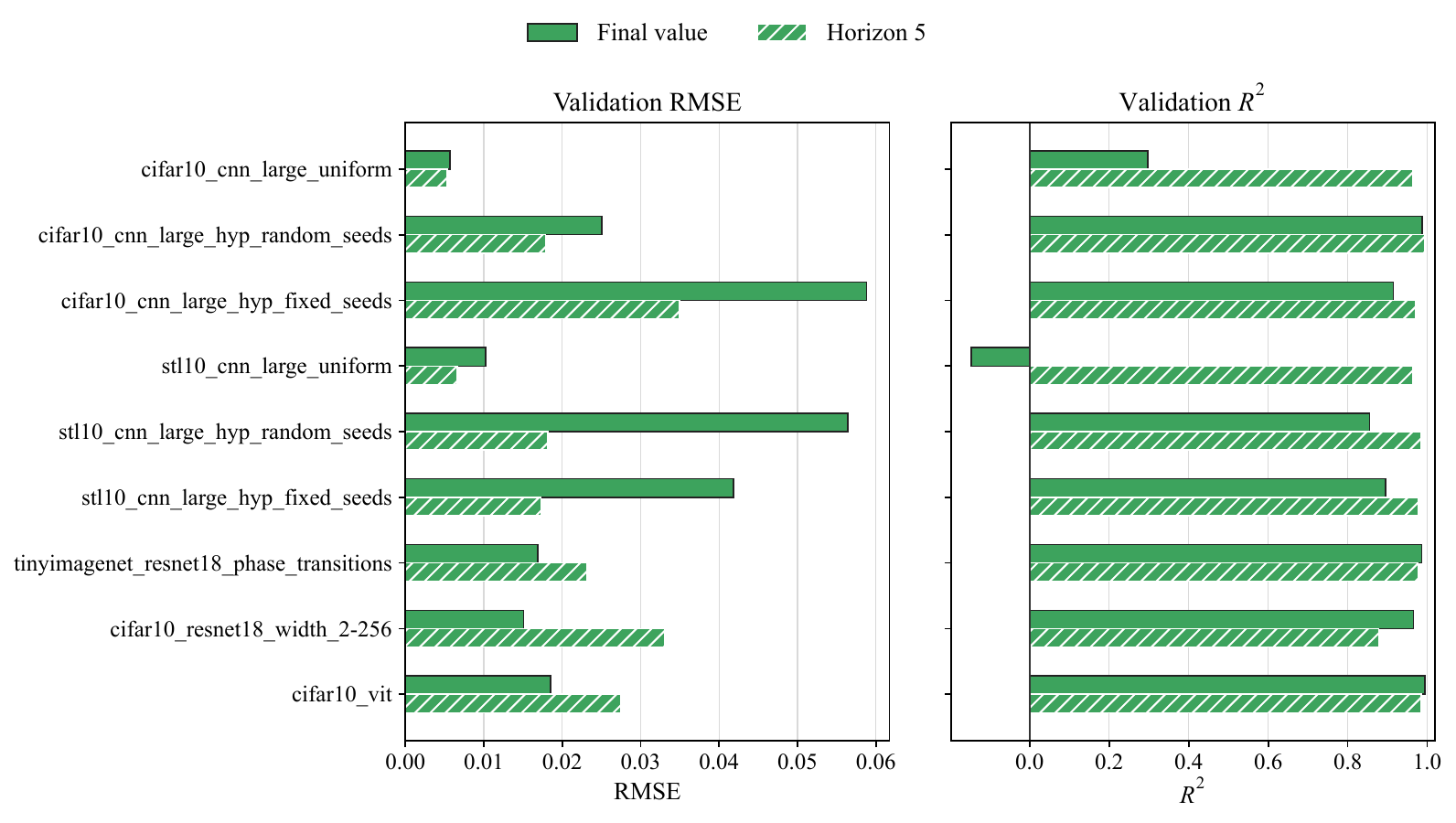}
	\caption{\textbf{Weight-statistic MLP performance by dataset.} Competitive for final-value
	nowcasting but selected less consistently than XGBoost.}
	\label{fig:app-ws-mlp-performance}
\end{figure}

\begin{figure}[H]
	\centering
	\includegraphics[width=0.75\linewidth]{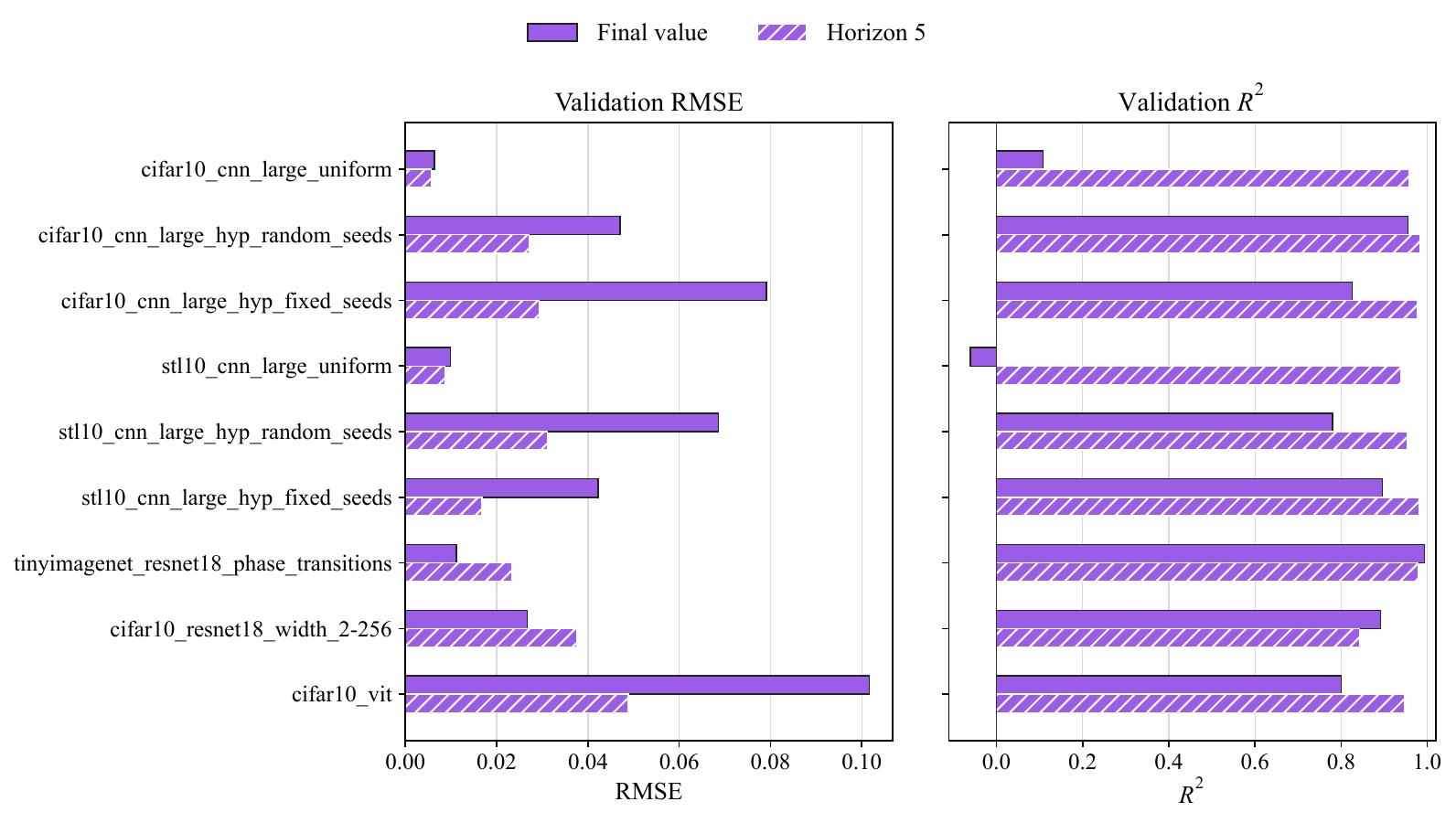}
	\caption{\textbf{Averaged-latent MLP performance by dataset.}}
	\label{fig:app-avglat-mlp-performance}
\end{figure}

\begin{figure}[H]
	\centering
	\includegraphics[width=0.75\linewidth]{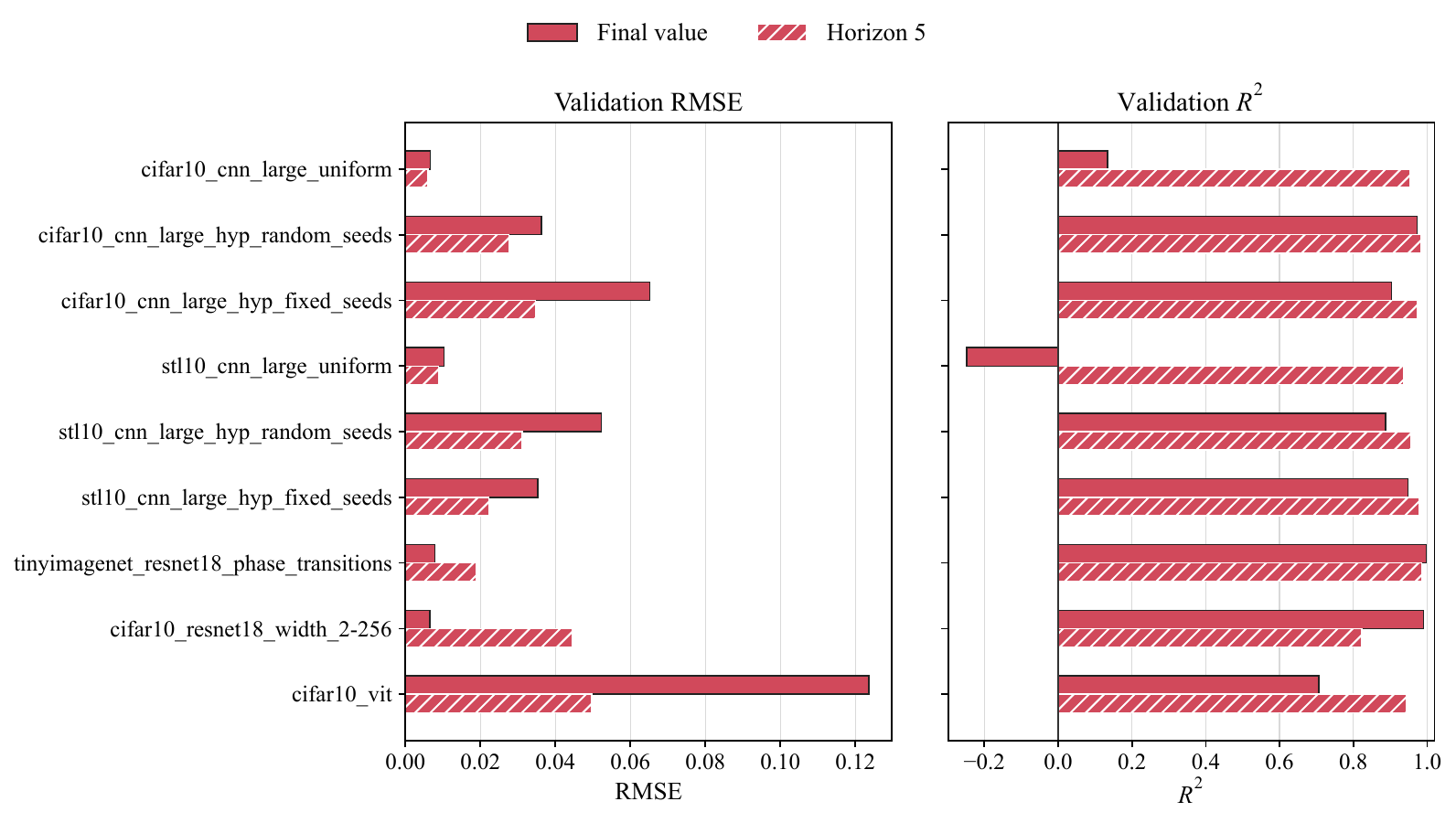}
	\caption{\textbf{Latent-window LSTM performance by dataset.} Strongest on final-value
	prediction for the trajectory-sensitive Tiny ImageNet phase-transition and CIFAR-10
	ResNet-18 width zoos.}
	\label{fig:app-lstm-performance}
\end{figure}

\begin{figure}[H]
	\centering
	\includegraphics[width=0.75\linewidth]{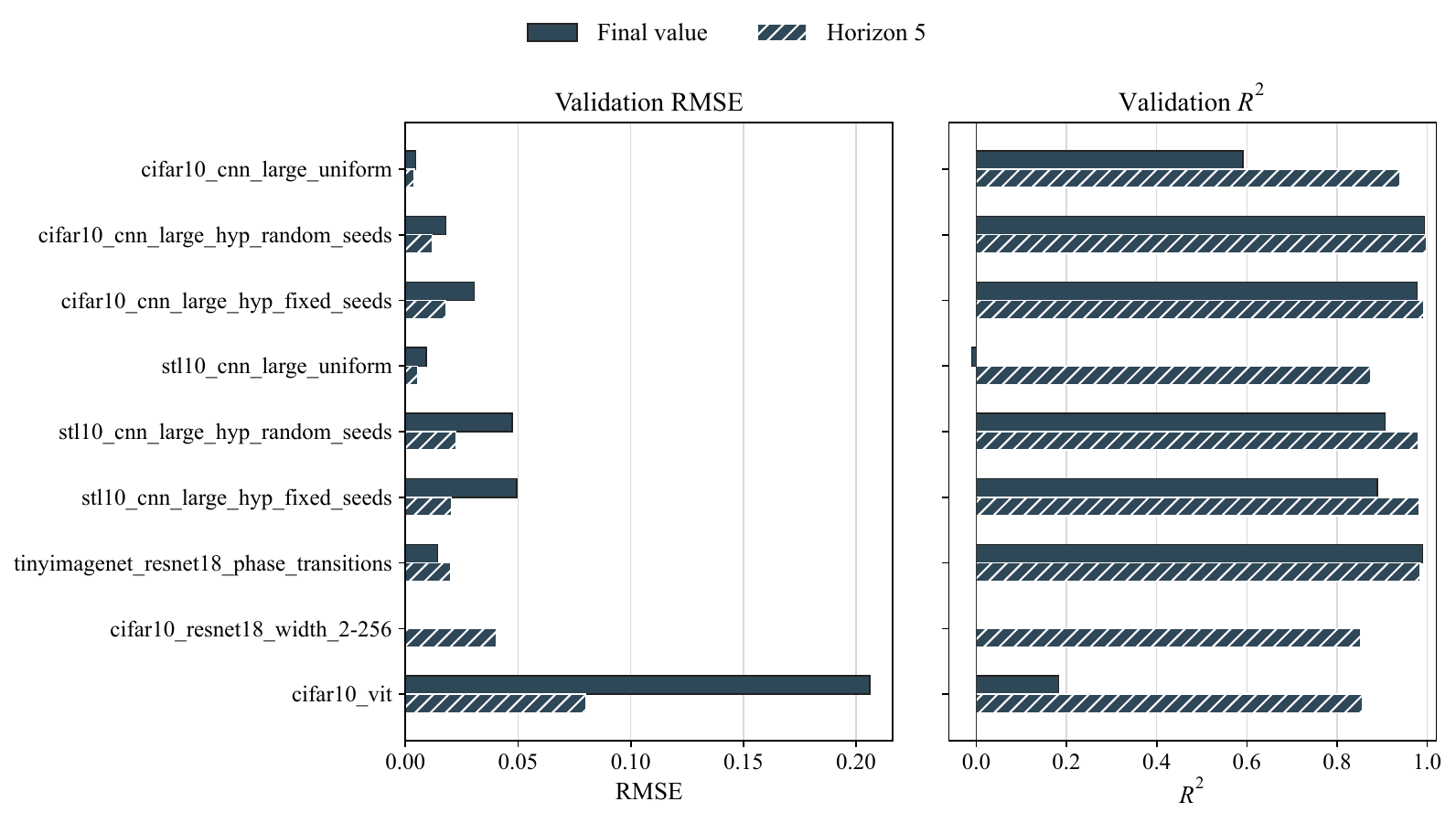}
	\caption{\textbf{Latent-token Transformer performance by dataset.}}
	\label{fig:app-transformer-performance}
\end{figure}

\begin{figure}[H]
	\centering
	\includegraphics[width=0.75\linewidth]{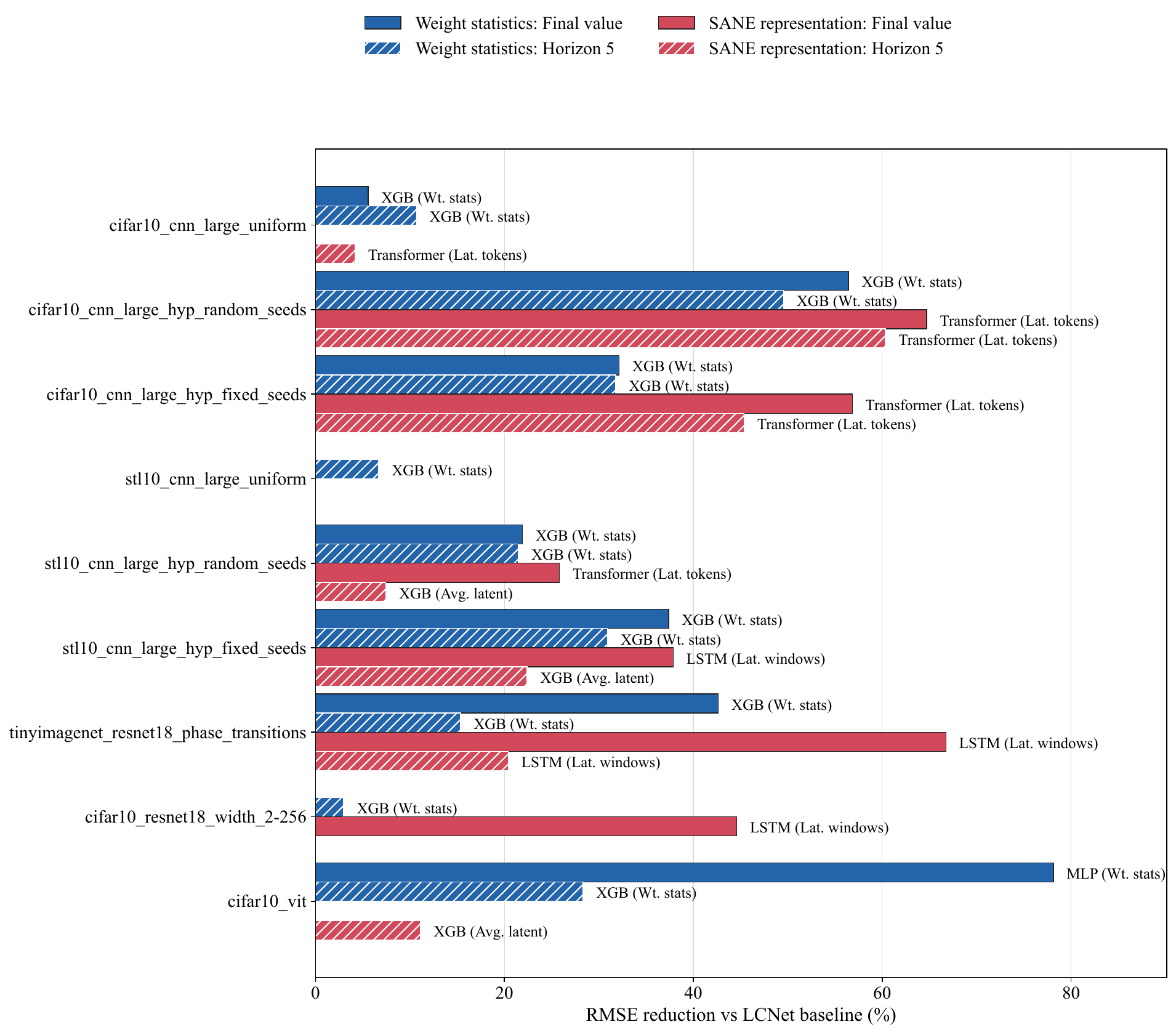}
	\caption{\textbf{Predictor improvement over the LCNet baseline.} Validation RMSE reduction
	relative to LCNet for the best weight-statistic and best SANE-representation predictor per
	dataset and target mode. Bars at zero indicate no improvement; labels report the selected
	predictor family.}
	\label{fig:app-improvement-vs-lcnet}
\end{figure}

Table~\ref{tab:app-predictor-improvement-vs-lcnet} expands Figure~\ref{fig:app-improvement-vs-lcnet}
to every predictor family, in percent RMSE reduction relative to LCNet (positive is better;
best family per row in bold).

\begin{table}[H]
\centering
\caption{\textbf{Predictor improvements over the LCNet baseline.} Validation RMSE reduction in
percent relative to LCNet. WS = weight-statistic, Avg = averaged SANE latent, LW = latent-window,
LT = latent-token.}
\label{tab:app-predictor-improvement-vs-lcnet}
\begingroup
\setlength{\tabcolsep}{3pt}
\footnotesize
\noindent\textbf{Final value.}\\[0.3em]
\resizebox{\textwidth}{!}{%
\begin{tabular}{@{}lrrrrrrr@{}}
\toprule
\textbf{Dataset} & \textbf{WS-XGB} & \textbf{WS-MLP} & \textbf{Avg-Ridge} & \textbf{Avg-XGB} & \textbf{Avg-MLP} & \textbf{LW-LSTM} & \textbf{LT-Trans.} \\
\midrule
cifar10\_cnn\_large\_uniform & \textbf{5.6} & -25.5 & -21.9 & -11.6 & -41.3 & -47.7 & -1.4 \\
cifar10\_cnn\_large\_hyp\_random\_seeds & 56.4 & 50.8 & 0.2 & 40.8 & 7.5 & 28.6 & \textbf{64.7} \\
cifar10\_cnn\_large\_hyp\_fixed\_seeds & 32.1 & 17.1 & 2.7 & 37.8 & -11.7 & 8.0 & \textbf{56.8} \\
stl10\_cnn\_large\_uniform & -0.6 & -58.3 & -33.6 & -18.7 & -52.2 & -58.3 & -42.4 \\
stl10\_cnn\_large\_hyp\_random\_seeds & 21.9 & 11.8 & -20.7 & 16.7 & -7.3 & 18.2 & \textbf{25.8} \\
stl10\_cnn\_large\_hyp\_fixed\_seeds & 37.4 & 26.3 & -26.7 & 30.7 & 25.7 & \textbf{37.9} & 12.9 \\
tinyimagenet\_resnet18\_phase\_transitions & 42.7 & 28.6 & 19.3 & 26.4 & 52.7 & \textbf{66.8} & 39.4 \\
cifar10\_resnet18\_width\_2-256 & -13.6 & -27.2 & -217.1 & -87.3 & -125.2 & \textbf{44.6} & -98.3 \\
cifar10\_vit & 65.1 & \textbf{78.1} & -14.4 & -65.7 & -20.4 & -46.3 & -143.8 \\
\bottomrule
\end{tabular}%
}
\par\vspace{1em}\noindent\textbf{Horizon 5.}\\[0.3em]
\resizebox{\textwidth}{!}{%
\begin{tabular}{@{}lrrrrrrr@{}}
\toprule
\textbf{Dataset} & \textbf{WS-XGB} & \textbf{WS-MLP} & \textbf{Avg-Ridge} & \textbf{Avg-XGB} & \textbf{Avg-MLP} & \textbf{LW-LSTM} & \textbf{LT-Trans.} \\
\midrule
cifar10\_cnn\_large\_uniform & \textbf{10.7} & -27.0 & -34.4 & 0.4 & -35.4 & -43.7 & 4.2 \\
cifar10\_cnn\_large\_hyp\_random\_seeds & 49.5 & 39.6 & -4.0 & 37.2 & 8.7 & 7.0 & \textbf{60.3} \\
cifar10\_cnn\_large\_hyp\_fixed\_seeds & 31.8 & -6.7 & -4.3 & 29.2 & 10.5 & -5.8 & \textbf{45.4} \\
stl10\_cnn\_large\_uniform & \textbf{6.7} & -53.6 & -44.3 & -27.8 & -99.6 & -104.0 & -21.5 \\
stl10\_cnn\_large\_hyp\_random\_seeds & \textbf{21.5} & -2.8 & -140.1 & 7.4 & -75.8 & -76.1 & -26.5 \\
stl10\_cnn\_large\_hyp\_fixed\_seeds & \textbf{30.9} & 5.2 & -160.0 & 22.4 & 8.1 & -21.9 & -12.1 \\
tinyimagenet\_resnet18\_phase\_transitions & 15.3 & 2.9 & -6.0 & 11.4 & 2.6 & \textbf{20.4} & 15.9 \\
cifar10\_resnet18\_width\_2-256 & \textbf{3.0} & -3.8 & -45.2 & -1.3 & -17.8 & -39.7 & -28.8 \\
cifar10\_vit & \textbf{28.3} & 3.4 & -48.6 & 11.1 & -71.9 & -75.0 & -181.8 \\
\bottomrule
\end{tabular}%
}
\endgroup
\end{table}

This table is the evidence behind the aggregate claims in Section~\ref{sec:results}: no single
representation family wins on every zoo, several families (Avg-Ridge, Avg-MLP, LT-Trans.\ under
horizon-5) are frequently \emph{worse} than the LCNet baseline, and the CIFAR-10 ResNet-18 width
and ViT zoos in particular show large negative entries for most SANE-sequence predictors,
underscoring that representation utility is population- and horizon-specific rather than
uniform.

\begin{figure}[H]
    \centering
    \includegraphics[width=0.75\linewidth]{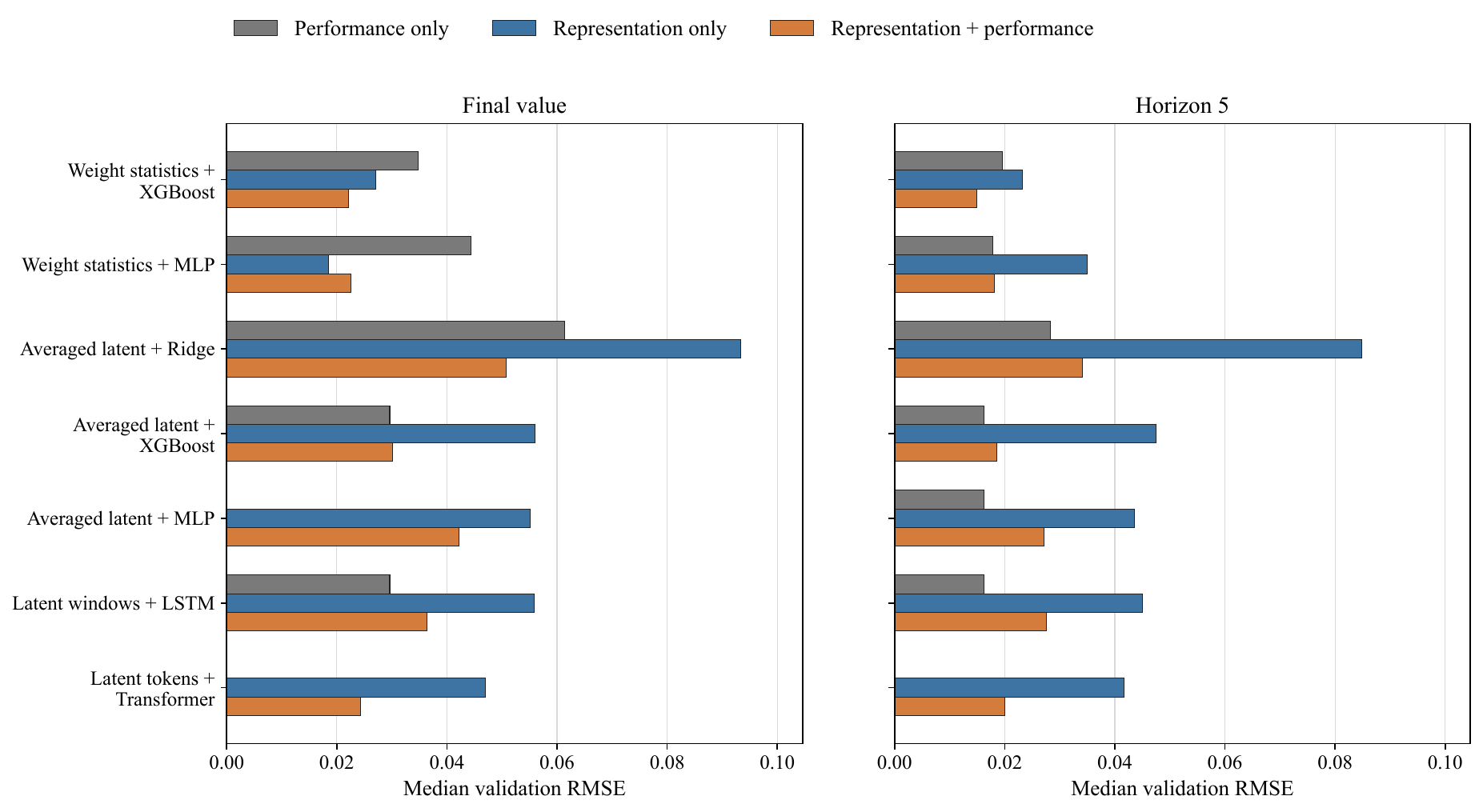}
    \caption{\textbf{Performance-feature ablation --- RMSE view.} Median validation RMSE
    (lower is better) for the performance-only, representation-only, and hybrid ablation
    configurations shown as $R^2$ in Figure~\ref{fig:predictor-feature-ablation}. The
    performance-only ablation for the latent-token Transformer was not run, since that
    architecture requires a latent-token sequence as input.}
    \label{fig:app-feature-ablation-rmse}
\end{figure}

\section{SOTF training configuration and full results}
\label{app:sotf-full-results}

\subsection{Ranking policies and sweep configuration}

Table~\ref{tab:app-sotf-configurations} lists all ranking-policy families evaluated in the
retrospective SOTF training experiments, and Table~\ref{tab:app-sotf-setup-sweeps} gives the exact start
epochs used for each model zoo. All runs maximize test accuracy, use decision interval
$I_{\mathrm{sel}}=5$ epochs, minimum population size $n_{\min}=1$, and retention rates
$\rho\in\{0.10,0.25,0.50,0.75,0.90\}$ ($\rho=1$ is excluded because it performs no pruning).
Start epochs differ by zoo because training horizons differ: phase-transition zoos train for
100--300 epochs, while CNN zoos train for only 50. Consistent with Section~\ref{sec:method},
every checkpoint predictor below is trained once on the training split and then held fixed
throughout SOTF training.

\begin{table}[H]
    \centering
    \caption{\textbf{SOTF training ranking configurations.} Complete list of ranking policies
    compared in SOTF training experiments.}
    \label{tab:app-sotf-configurations}
    \begin{tabularx}{\textwidth}{@{}>{\raggedright\arraybackslash}p{0.36\textwidth} X@{}}
        \toprule
        \textbf{Configuration} & \textbf{Predictor used for ranking} \\
        \midrule
        Naive current-value baseline & Current observed test accuracy. \\
        LC Extrapolation baseline & Parametric learning-curve fit applied to the observed accuracy history. \\
        LCNet baseline & LCNet-style predictor applied to the observed accuracy history. Final value: the predictor from the metric-prediction experiments, trained on the complete training-split learning curves of the zoo and held fixed. Horizon 5: refit at every pruning round on the history observed so far. \\
        Weight-statistic XGBoost variants & Predictors loaded from the metric-prediction experiments, using weight-statistic features with and without observed metrics. \\
        SANE latent variants & Predictors loaded from the metric-prediction experiments, using averaged latent, latent-window LSTM, or latent-token Transformer predictors for selected zoos. \\
        \bottomrule
    \end{tabularx}
\end{table}

\begin{table}[H]
    \centering
    \footnotesize
    \caption{\textbf{SOTF training model-zoo coverage and start epochs.}}
    \label{tab:app-sotf-setup-sweeps}
    \begin{tabularx}{\textwidth}{@{}>{\raggedright\arraybackslash}X c c@{}}
        \toprule
        \textbf{Model zoo} & \textbf{Horizon-5 start epochs} & \textbf{Final-value start epochs} \\
        \midrule
        cifar10\_cnn\_large\_uniform & 5, 10, 25 & 5, 10, 25 \\
        cifar10\_cnn\_large\_hyp\_random\_seeds & 5, 10, 25 & 5, 10, 25 \\
        cifar10\_cnn\_large\_hyp\_fixed\_seeds & 5, 10, 25 & 5, 10, 25 \\
        stl10\_cnn\_large\_uniform & 5, 10, 25 & 5, 10, 25 \\
        stl10\_cnn\_large\_hyp\_random\_seeds & 5, 10, 25 & 5, 10, 25 \\
        stl10\_cnn\_large\_hyp\_fixed\_seeds & 5, 10, 25 & 5, 10, 25 \\
        tinyimagenet\_resnet18\_phase\_transitions & 20, 30, 40 & 20, 30, 40 \\
        cifar10\_resnet18\_width\_2-256 & 20, 40, 60 & 20, 40, 60 \\
        cifar10\_vit & 20, 40, 60 & 20, 40, 60 \\
        \bottomrule
    \end{tabularx}
\end{table}

\subsection{Full results by population regime}

Table~\ref{tab:app-sotf-regime-full} extends Table~\ref{tab:sotf-regime} to all 15 evaluated
ranking policies, and Table~\ref{tab:app-sotf-per-zoo} breaks the final-value results down for
the three ResNet-18 and ViT zoos. All values are averaged over the start epochs in
Table~\ref{tab:app-sotf-setup-sweeps} and the five retention rates, restricted to configurations
in which every policy completed (269 of 270; one final-value configuration of the CIFAR-10
ResNet-18 width zoo is missing for the current-accuracy baseline). Compute savings are
determined by the pruning schedule and are therefore nearly identical across policies at a given
configuration; the sensitivity plots in Section~\ref{app:sotf-sensitivity} report them per cell.

\begin{table}[H]
    \centering
    \caption{\textbf{SOTF training results for all ranking policies by population regime.} Each
    cell reports the mean best-retained-model gap / mean oracle top-$K$ recall, separately for the
    six CNN zoos and the three ResNet-18 and ViT zoos. Final value ranks by predicted test accuracy
    at the end of training; horizon 5 ranks by predicted test accuracy five epochs ahead. Lower gap
    and higher recall are better; the best value per column is in bold. ``perf.'' denotes accuracy
    and loss history features, ``weights'' engineered weight statistics, and ``SANE'' SANE
    representations.}
    \label{tab:app-sotf-regime-full}
    \setlength{\tabcolsep}{4pt}
    \footnotesize
    \begin{tabular}{@{}lcccc@{}}
    \toprule
     & \multicolumn{2}{c}{\textbf{Final value}} & \multicolumn{2}{c}{\textbf{Horizon 5}} \\
    \cmidrule(lr){2-3}\cmidrule(lr){4-5}
    \textbf{Ranking policy} & CNN & ResNet-18 / ViT & CNN & ResNet-18 / ViT \\
    \midrule
    Current accuracy & 0.0051 / \textbf{0.48} & 0.0643 / 0.11 & 0.0051 / 0.48 & 0.0629 / 0.11 \\
    LC extrapolation & 0.0062 / 0.47 & 0.0689 / 0.05 & 0.0050 / \textbf{0.51} & 0.0490 / 0.05 \\
    LCNet & 0.0056 / \textbf{0.48} & 0.0247 / 0.10 & 0.0104 / 0.43 & 0.0656 / 0.11 \\
    \midrule
    XGB (perf.) & 0.0040 / 0.45 & 0.0051 / 0.20 & \textbf{0.0039} / 0.46 & 0.0495 / 0.09 \\
    MLP (perf.) & \textbf{0.0038} / \textbf{0.48} & 0.0082 / 0.22 & 0.0047 / 0.46 & 0.0494 / 0.14 \\
    \midrule
    XGB (weights) & 0.0060 / 0.34 & 0.0018 / 0.67 & 0.0056 / 0.37 & 0.0608 / 0.09 \\
    XGB (perf.\ + weights) & 0.0054 / 0.42 & \textbf{0.0017} / \textbf{0.72} & 0.0043 / 0.45 & \textbf{0.0479} / 0.16 \\
    \midrule
    Avg.\ latent XGB (SANE) & 0.0112 / 0.29 & 0.0410 / 0.48 & 0.0112 / 0.28 & 0.1311 / 0.24 \\
    Avg.\ latent XGB (perf.\ + SANE) & 0.0055 / 0.38 & 0.0185 / 0.44 & 0.0058 / 0.44 & 0.0622 / 0.09 \\
    Avg.\ latent MLP (SANE) & 0.0105 / 0.29 & 0.1130 / 0.14 & 0.0101 / 0.34 & 0.1982 / \textbf{0.27} \\
    Avg.\ latent MLP (perf.\ + SANE) & 0.0081 / 0.33 & 0.0379 / 0.21 & 0.0080 / 0.39 & 0.0579 / 0.18 \\
    Latent-window LSTM (SANE) & 0.0119 / 0.28 & 0.0234 / 0.08 & 0.0134 / 0.27 & 0.1677 / 0.06 \\
    Latent-window LSTM (perf.\ + SANE) & 0.0107 / 0.33 & 0.0241 / 0.27 & 0.0110 / 0.33 & 0.0580 / 0.05 \\
    Latent-token Transf.\ (SANE) & 0.0069 / 0.32 & 0.1140 / 0.42 & 0.0095 / 0.29 & 0.2292 / 0.06 \\
    Latent-token Transf.\ (perf.\ + SANE) & 0.0060 / 0.36 & 0.0991 / 0.19 & 0.0061 / 0.43 & 0.0504 / 0.06 \\
    \bottomrule
    \end{tabular}
\end{table}

\begin{table}[H]
    \centering
    \caption{\textbf{Final-value SOTF training results per ResNet-18 and ViT zoo.} Each cell reports
    the mean best-retained-model gap / mean oracle top-$K$ recall over start epochs and retention
    rates. Lower gap and higher recall are better; the best value per column is in bold.}
    \label{tab:app-sotf-per-zoo}
    \setlength{\tabcolsep}{4pt}
    \footnotesize
    \begin{tabular}{@{}lccc@{}}
    \toprule
    \textbf{Ranking policy} & \textbf{Tiny ImageNet ResNet-18} & \textbf{CIFAR-10 ResNet-18 width} & \textbf{CIFAR-10 ViT} \\
    \midrule
    Current accuracy & 0.0387 / 0.16 & 0.0249 / 0.19 & 0.1267 / 0.00 \\
    LC extrapolation & 0.0461 / 0.13 & 0.0223 / 0.03 & 0.1351 / 0.00 \\
    LCNet & 0.0050 / 0.17 & 0.0031 / 0.04 & 0.0645 / 0.07 \\
    \midrule
    XGB (perf.) & 0.0063 / 0.25 & 0.0037 / 0.21 & 0.0053 / 0.13 \\
    XGB (weights) & 0.0051 / 0.26 & \textbf{0.0000} / \textbf{0.99} & 0.0001 / 0.80 \\
    XGB (perf.\ + weights) & 0.0051 / 0.26 & \textbf{0.0000} / \textbf{0.99} & \textbf{0.0000} / \textbf{0.93} \\
    \midrule
    Avg.\ latent XGB (perf.\ + SANE) & \textbf{0.0000} / \textbf{0.98} & 0.0003 / 0.33 & 0.0540 / 0.00 \\
    Latent-window LSTM (perf.\ + SANE) & 0.0013 / 0.64 & 0.0033 / 0.14 & 0.0664 / 0.00 \\
    Latent-token Transf.\ (perf.\ + SANE) & 0.0065 / 0.19 & 0.0034 / 0.40 & 0.2811 / 0.00 \\
    \bottomrule
    \end{tabular}
\end{table}

Two observations complement the main text. First, appending observed performance features
reduces the gap for almost every SANE predictor and is essential in horizon mode, where
SANE-only predictors are worse than every baseline on the ResNet-18 and ViT zoos. Second, the
per-zoo breakdown shows that the regime averages hide strong zoo-specific differences: SANE
representations retain the best model on Tiny ImageNet but fail on the ViT zoo, whereas weight
statistics are near-perfect on the CIFAR-10 width and ViT zoos. On the width zoo, this may partly
reflect that weight statistics encode network width (Section~\ref{sec:discussion}).

\subsection{Parameter-sensitivity plots}
\label{app:sotf-sensitivity}

\begin{figure}[H]
    \centering
    \includegraphics[width=0.55\linewidth]{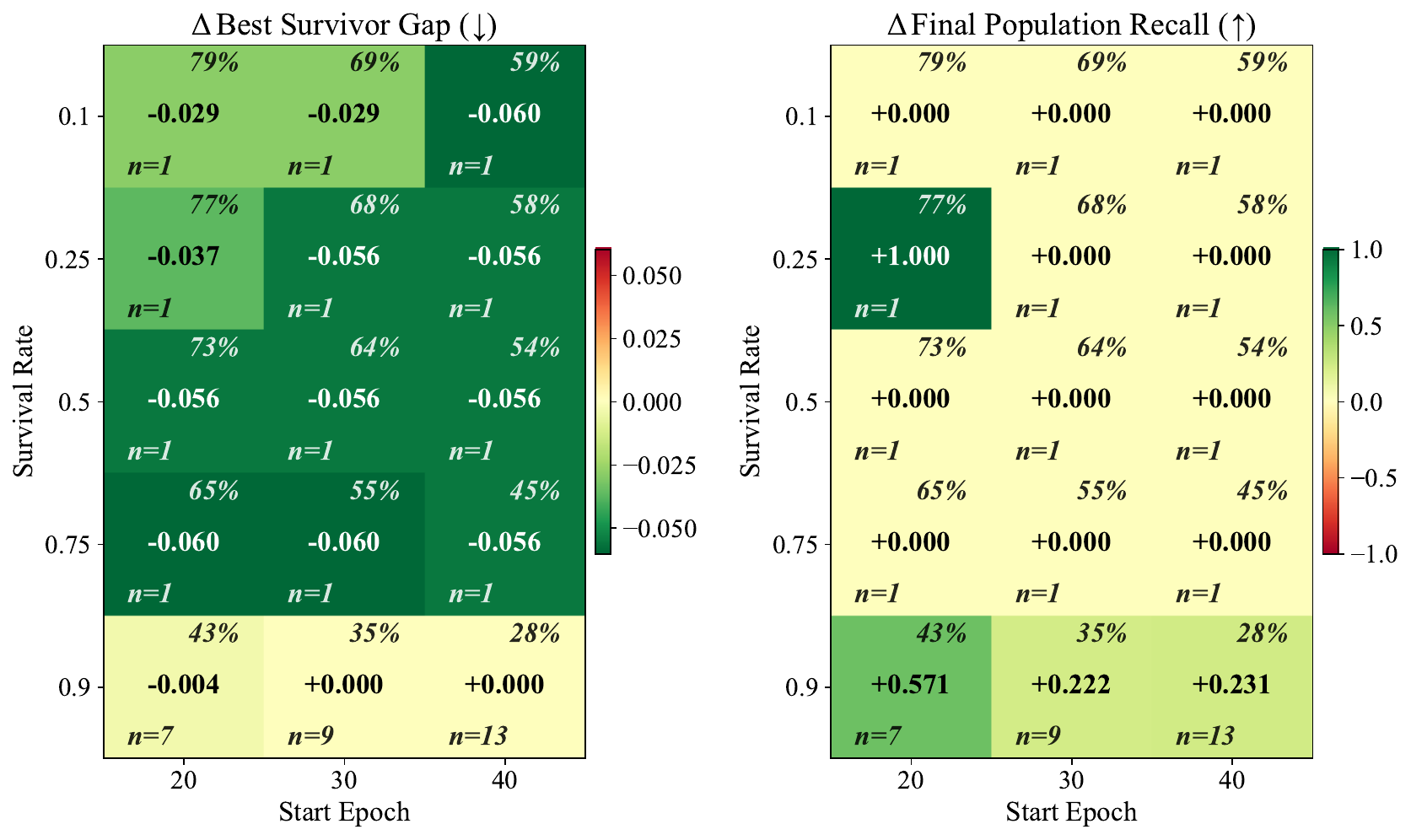}
    \caption{TinyImageNet phase, final value: weight-statistic XGBoost with performance features; mostly robust improvements.}
    \label{fig:app-sens-1}
\end{figure}

\begin{figure}[H]
    \centering
    \includegraphics[width=0.55\linewidth]{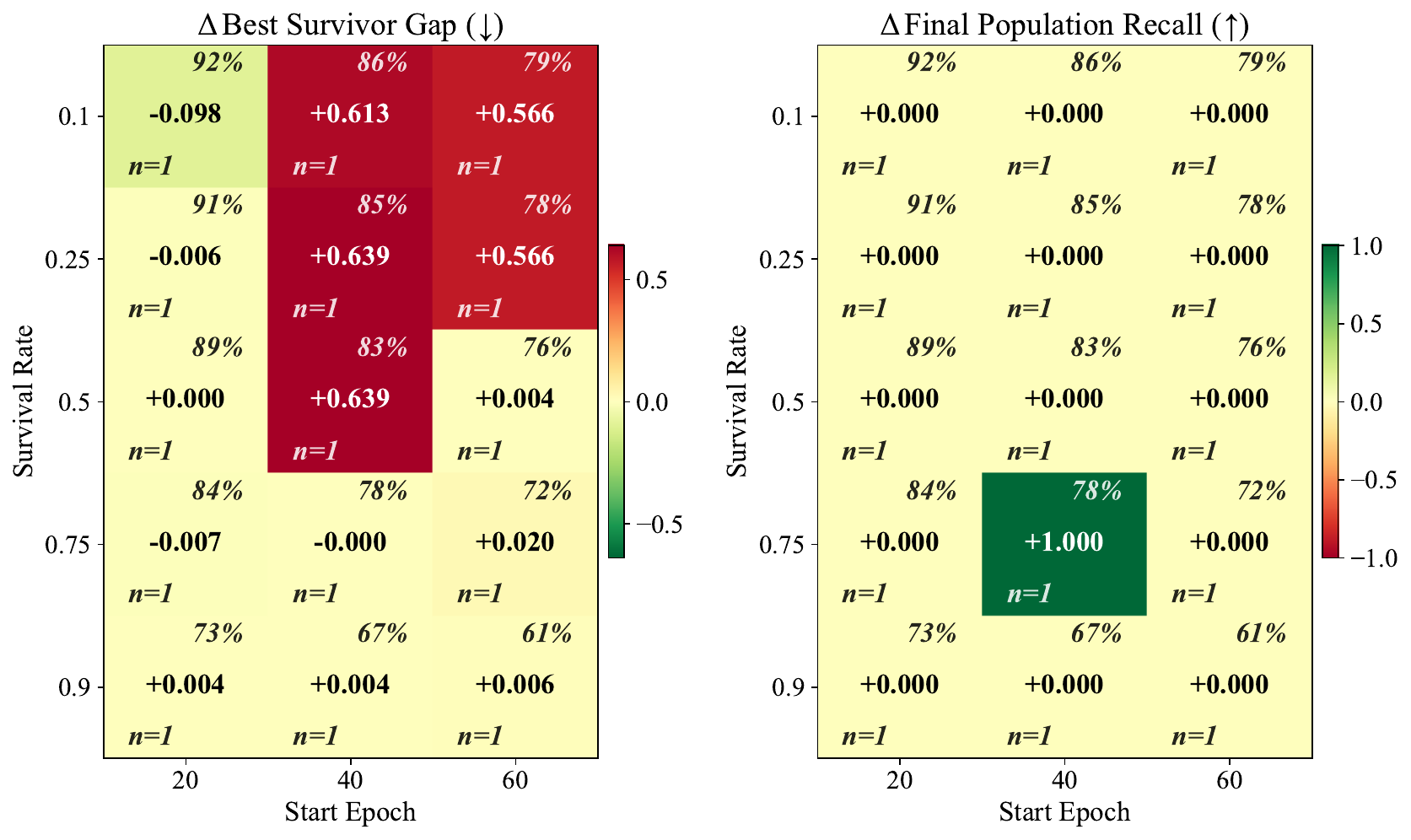}
    \caption{CIFAR-10 ViT, final value: latent-token Transformer; mixed results across pruning schedules.}
    \label{fig:app-sens-2}
\end{figure}

\begin{figure}[H]
    \centering
    \includegraphics[width=0.55\linewidth]{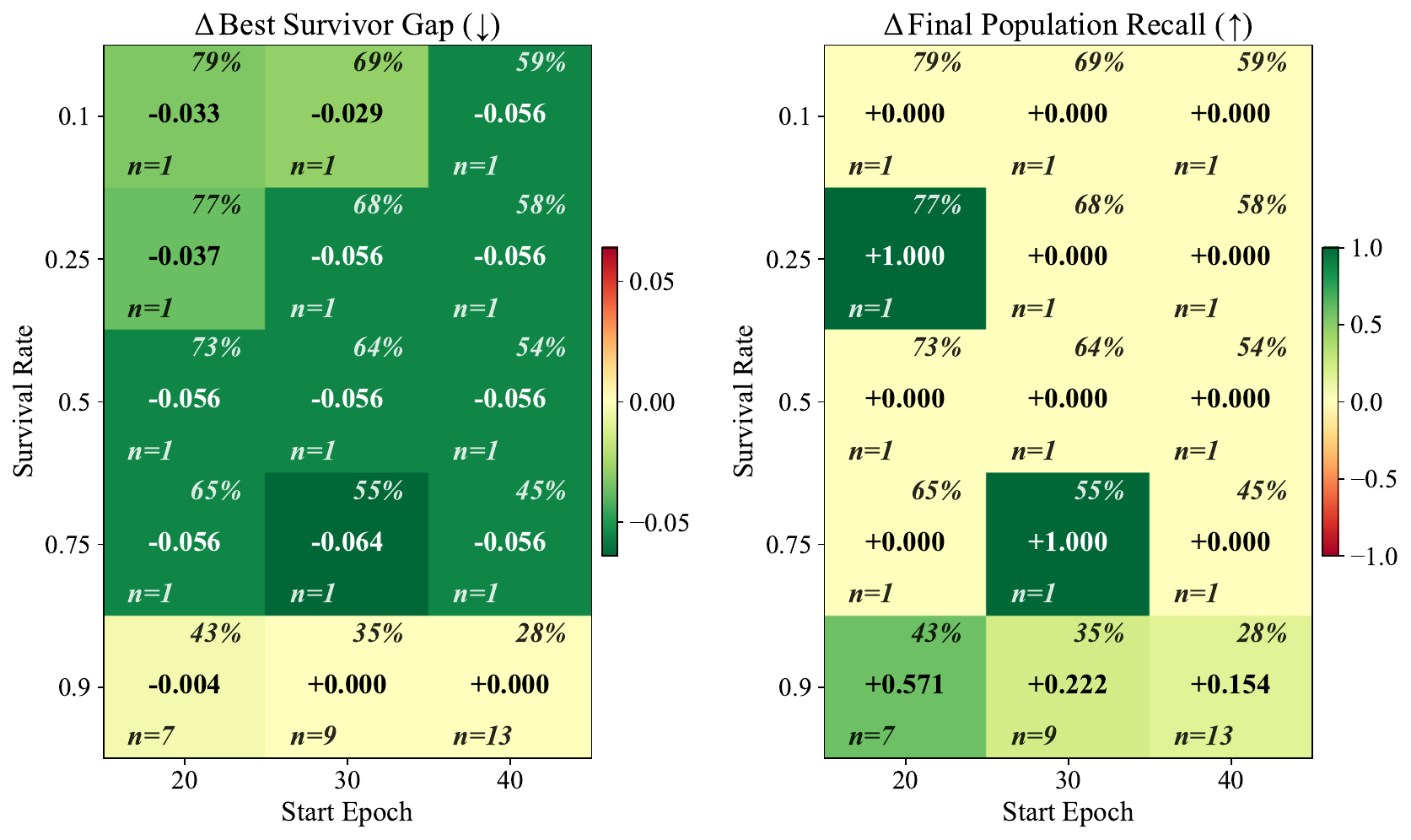}
    \caption{TinyImageNet phase, final value: averaged-latent MLP; consistently better under meaningful pruning.}
    \label{fig:app-sens-3}
\end{figure}

\begin{figure}[H]
    \centering
    \includegraphics[width=0.55\linewidth]{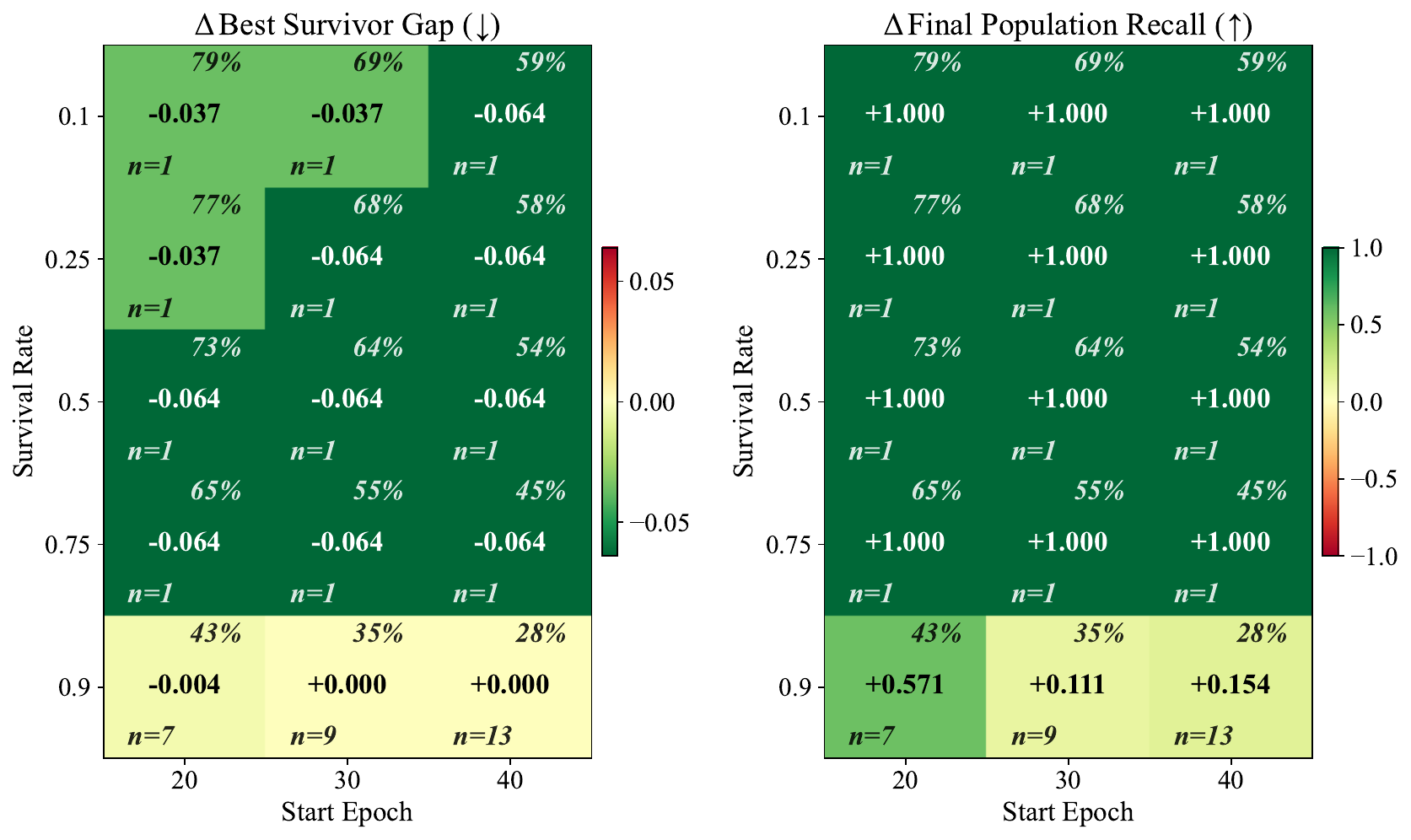}
    \caption{TinyImageNet phase, final value: averaged-latent XGBoost with performance features; robust improvements.}
    \label{fig:app-sens-4}
\end{figure}

\begin{figure}[H]
    \centering
    \includegraphics[width=0.55\linewidth]{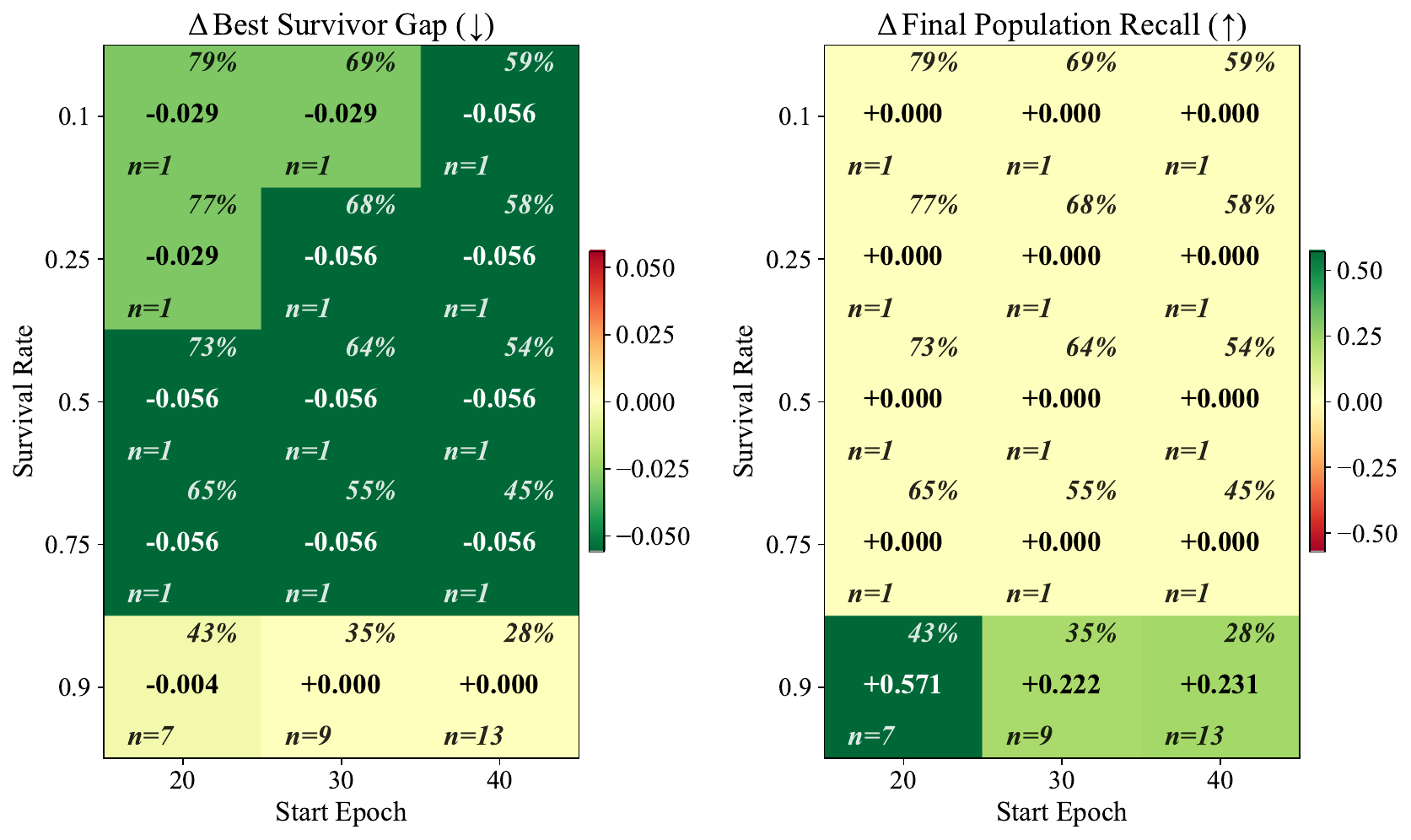}
    \caption{TinyImageNet phase, final value: latent-token Transformer with performance features; broadly effective.}
    \label{fig:app-sens-5}
\end{figure}

\begin{figure}[H]
    \centering
    \includegraphics[width=0.55\linewidth]{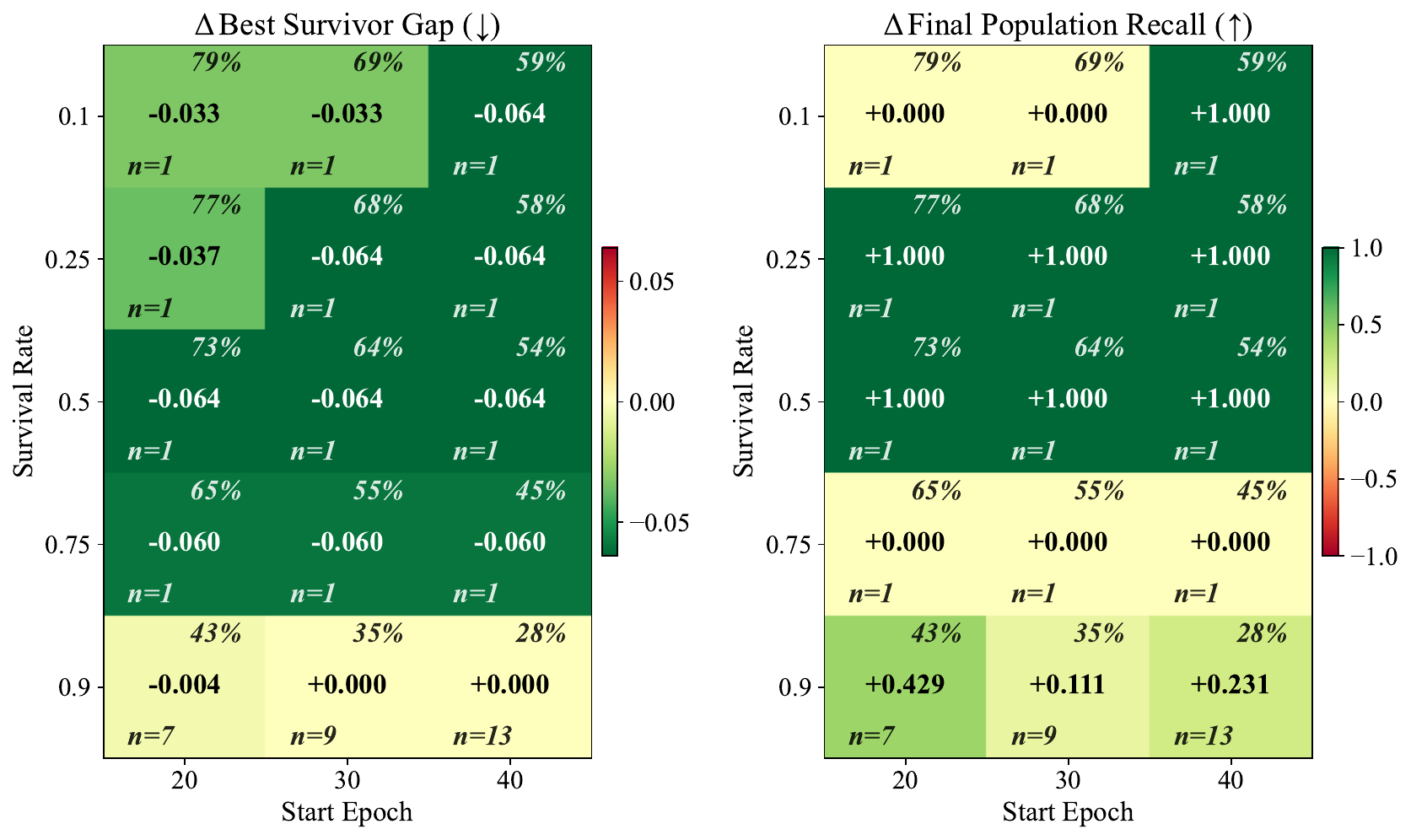}
    \caption{TinyImageNet phase, final value: latent-window LSTM with performance features; mostly robust improvements.}
    \label{fig:app-sens-6}
\end{figure}

\begin{figure}[H]
    \centering
    \includegraphics[width=0.55\linewidth]{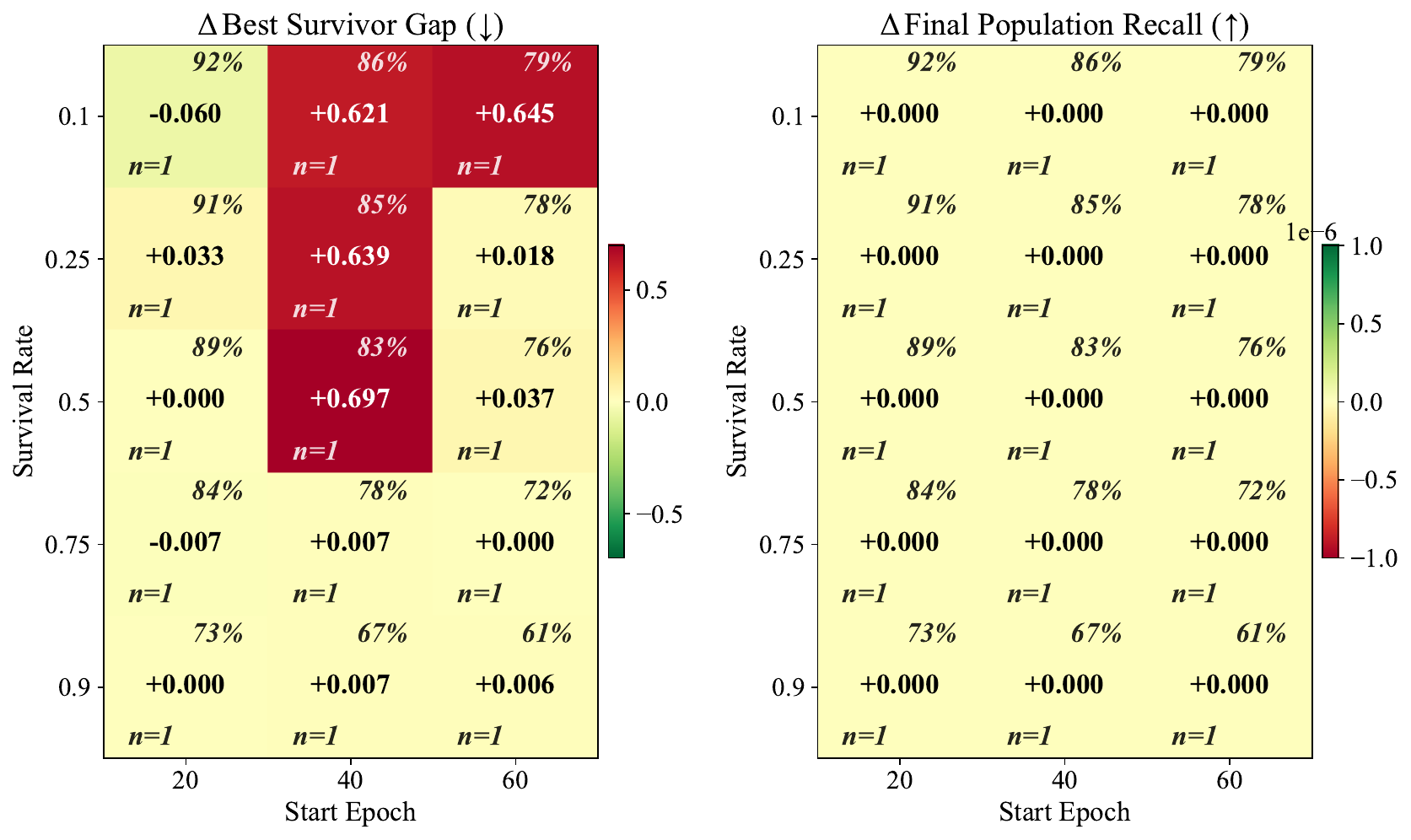}
    \caption{CIFAR-10 ViT, final value: averaged-latent MLP; unstable and frequently worse than LC Extrapolation.}
    \label{fig:app-sens-7}
\end{figure}

\begin{figure}[H]
    \centering
    \includegraphics[width=0.55\linewidth]{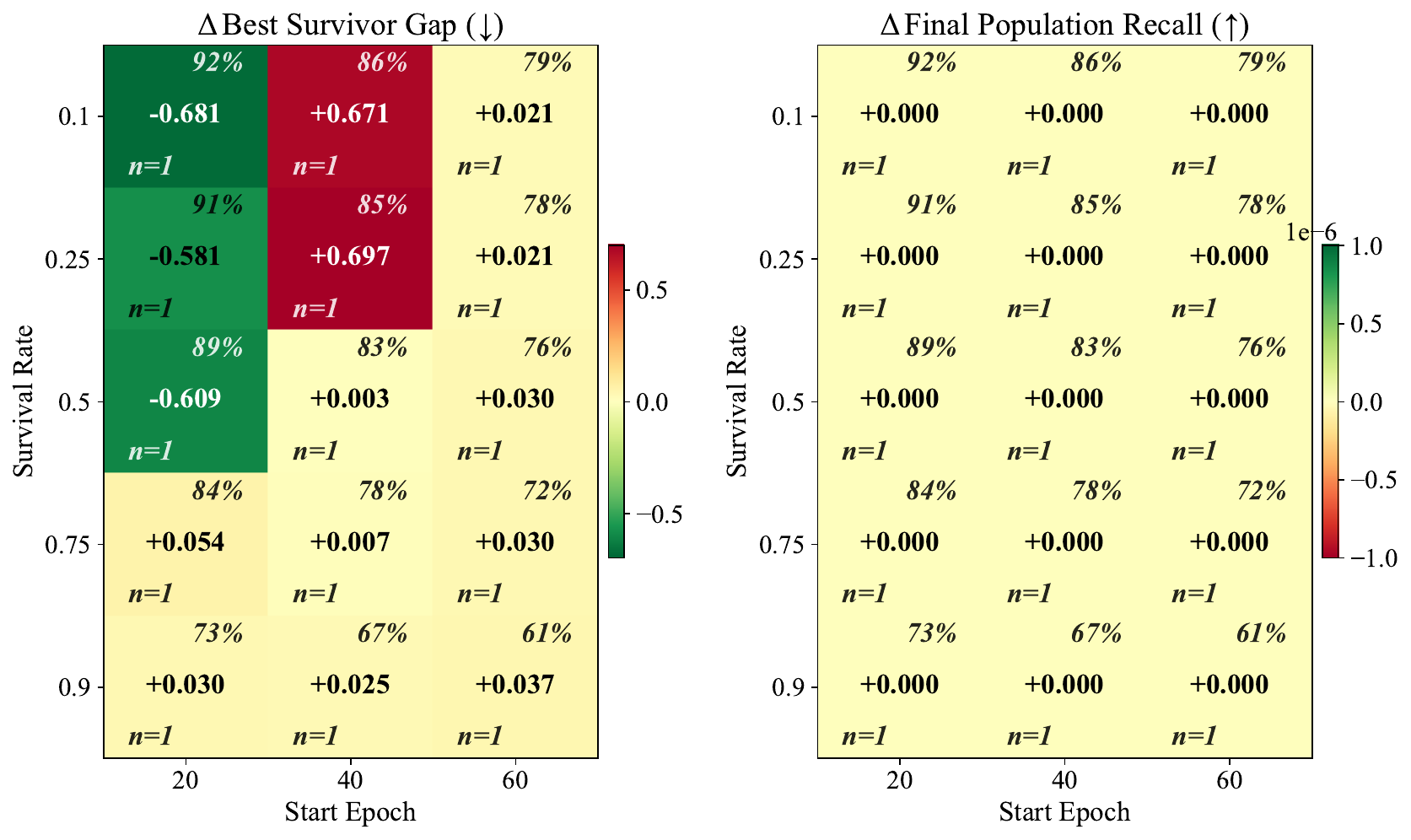}
    \caption{CIFAR-10 ViT, final value: averaged-latent XGBoost; strong early gains but poor transfer to later starts.}
    \label{fig:app-sens-8}
\end{figure}

\begin{figure}[H]
    \centering
    \includegraphics[width=0.55\linewidth]{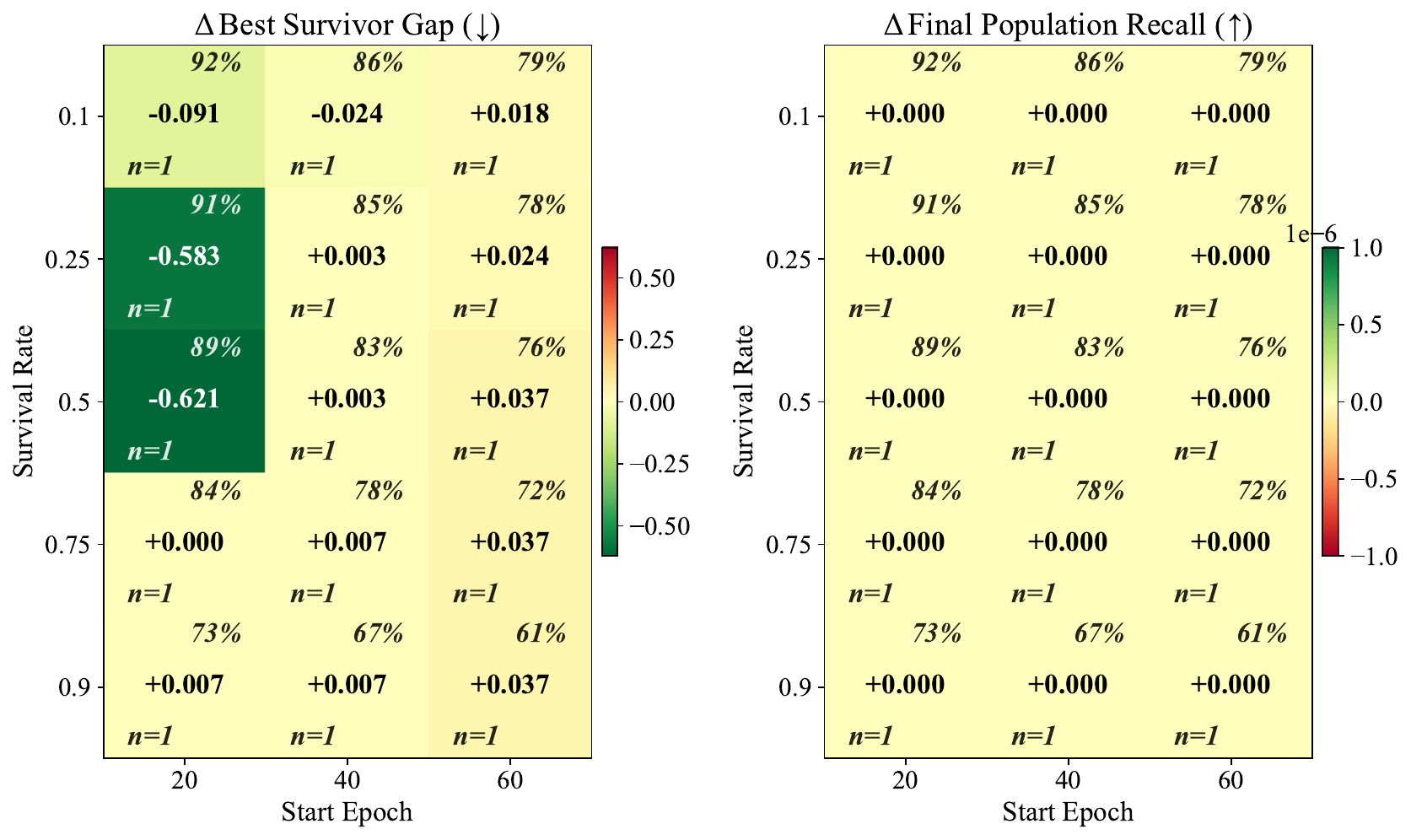}
    \caption{CIFAR-10 ViT, final value: latent-window LSTM; effective early but unreliable elsewhere.}
    \label{fig:app-sens-9}
\end{figure}

\begin{figure}[H]
    \centering
    \includegraphics[width=0.55\linewidth]{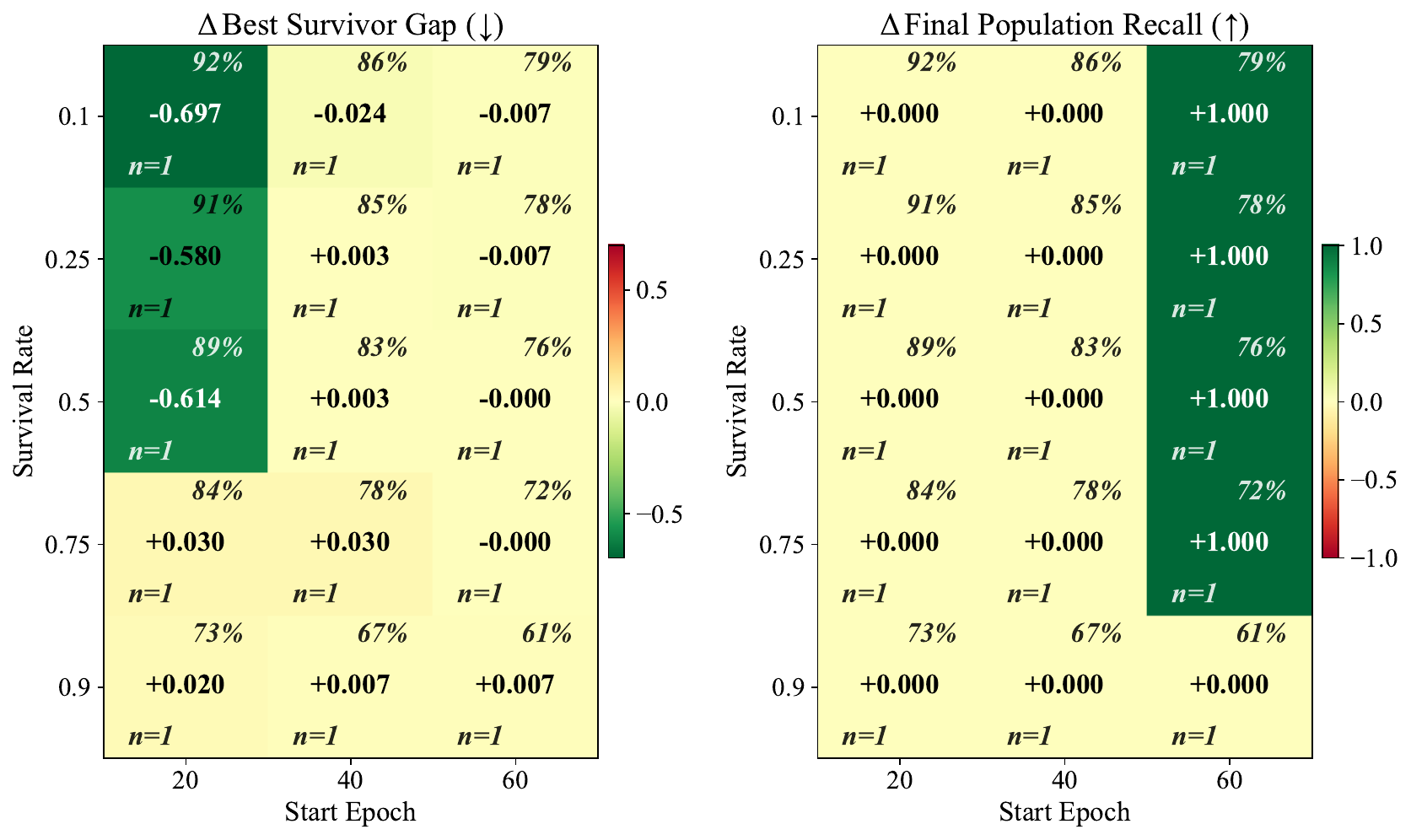}
    \caption{CIFAR-10 ViT, final value: performance-only MLP; large early gains coexist with later degradations.}
    \label{fig:app-sens-10}
\end{figure}

\begin{figure}[H]
    \centering
    \includegraphics[width=0.55\linewidth]{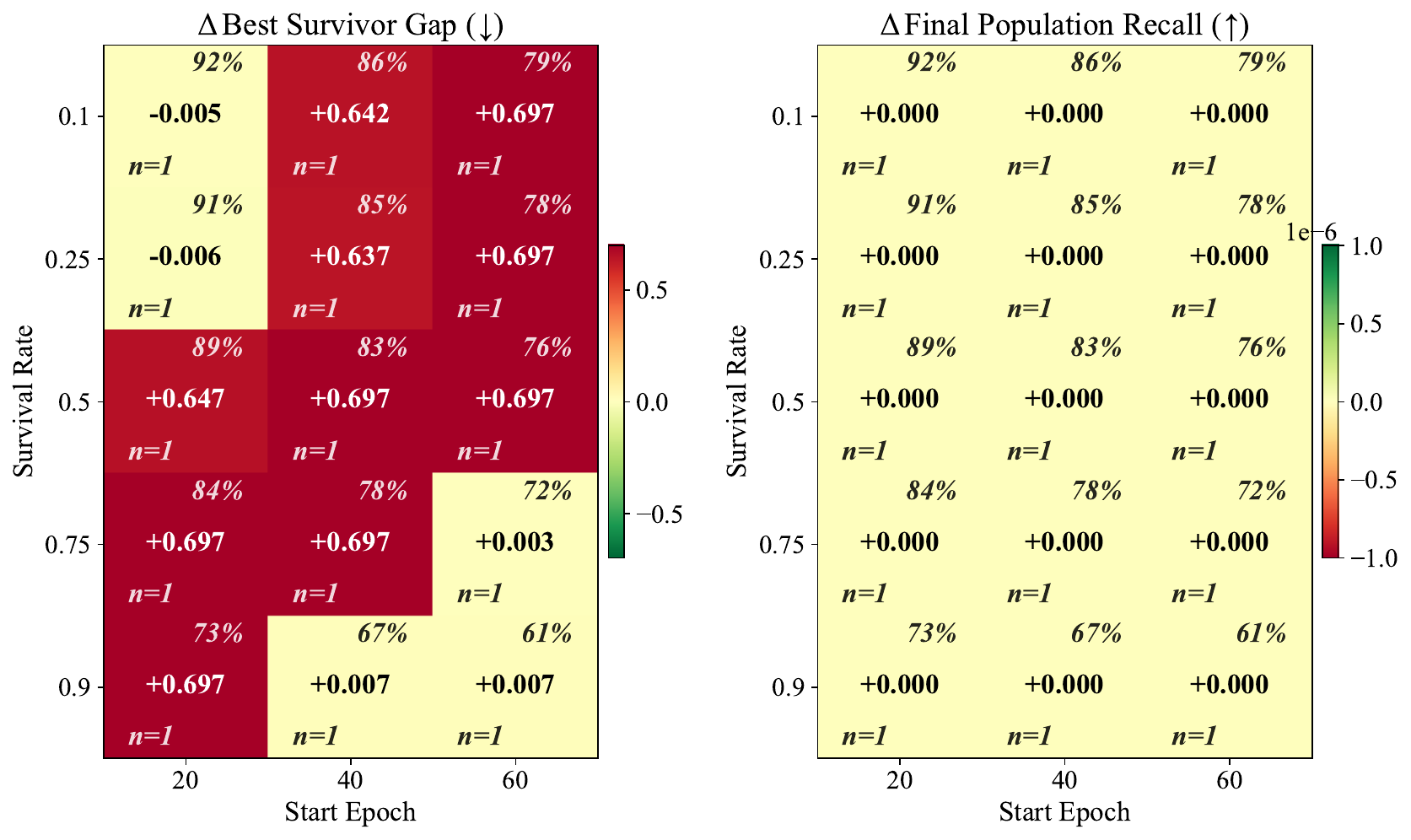}
    \caption{CIFAR-10 ViT, horizon value: averaged-latent MLP; predominantly worse than LC Extrapolation.}
    \label{fig:app-sens-11}
\end{figure}

\begin{figure}[H]
    \centering
    \includegraphics[width=0.55\linewidth]{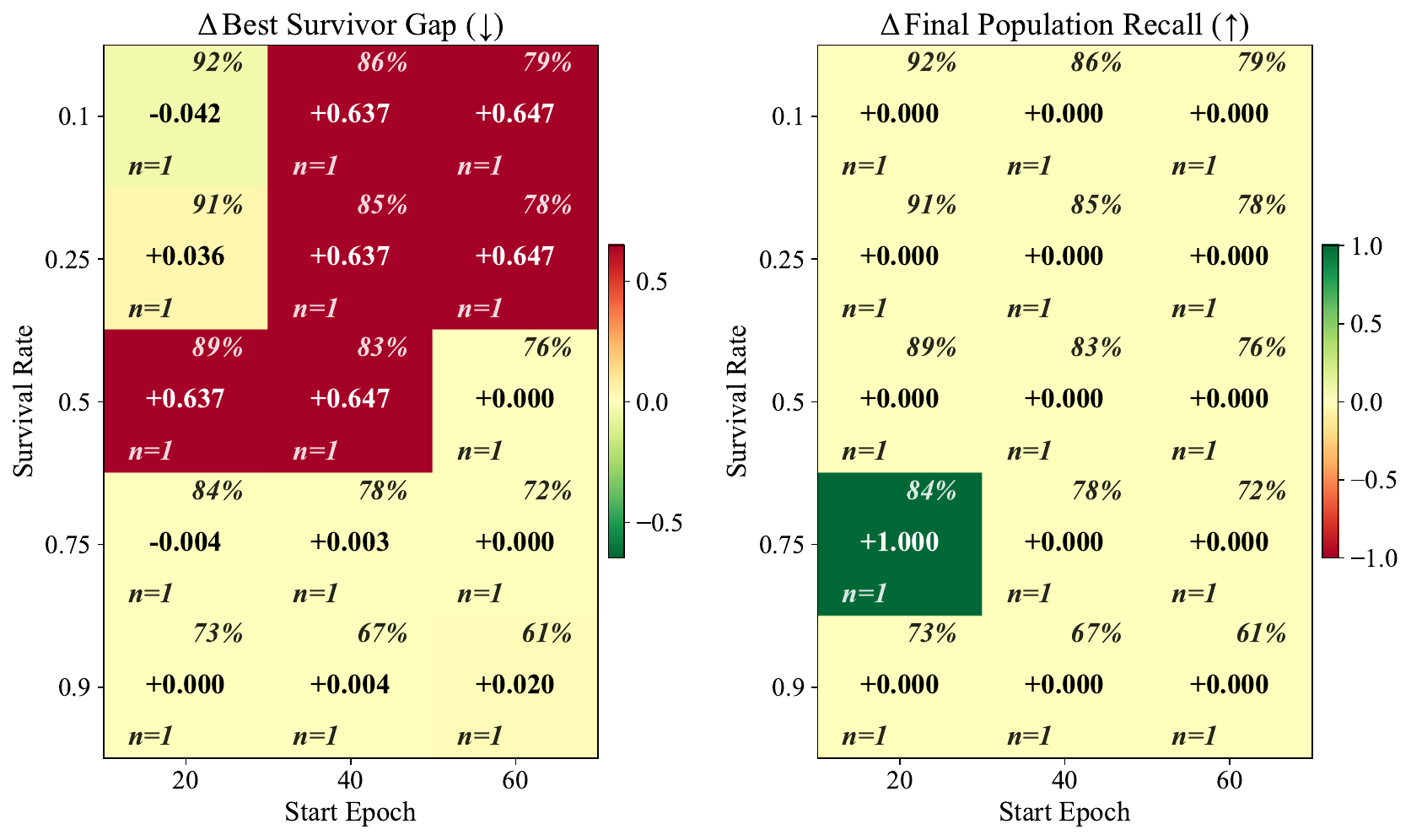}
    \caption{CIFAR-10 ViT, horizon value: averaged-latent XGBoost; mostly unsuccessful except at isolated settings.}
    \label{fig:app-sens-12}
\end{figure}

\begin{figure}[H]
    \centering
    \includegraphics[width=0.55\linewidth]{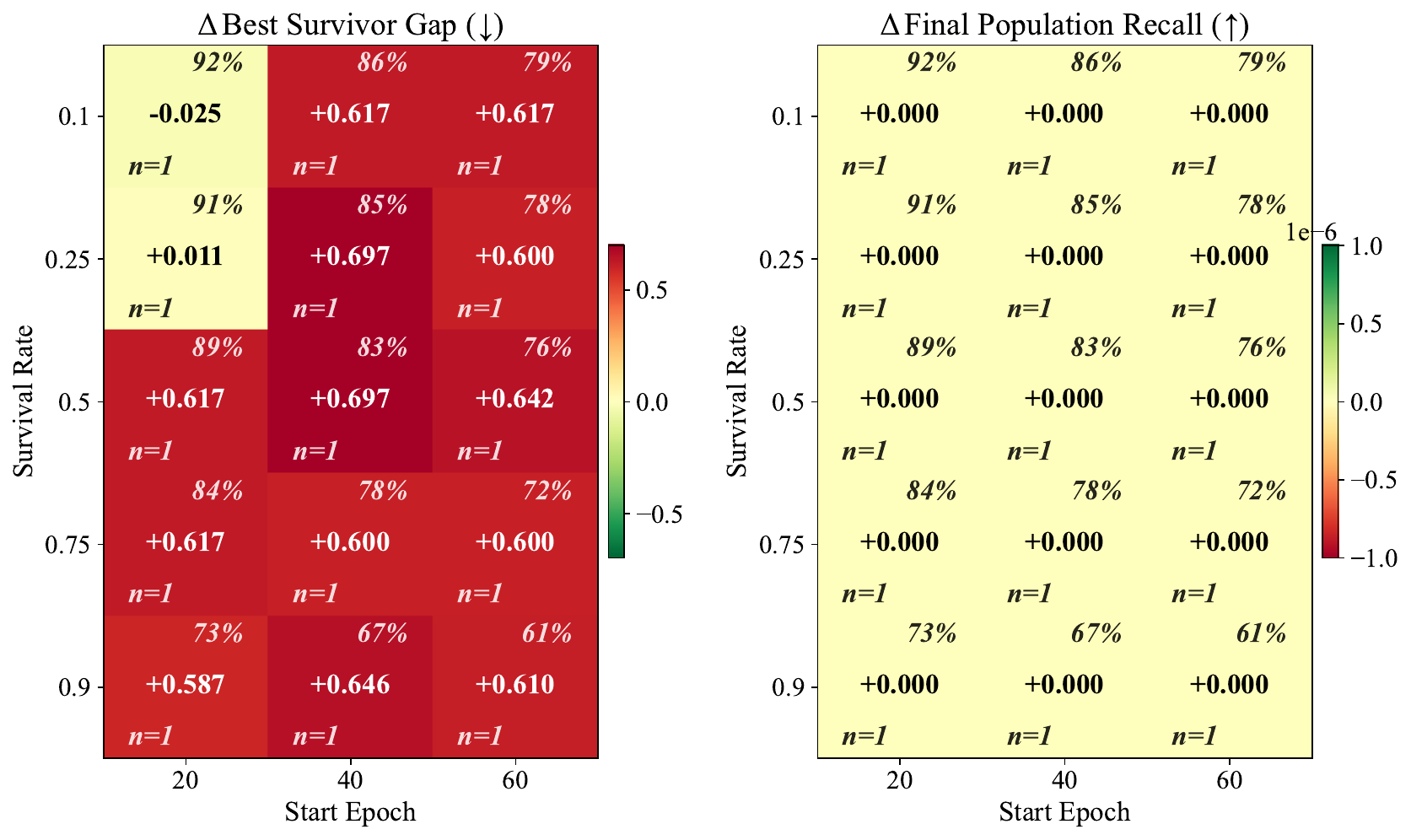}
    \caption{CIFAR-10 ViT, horizon value: latent-token Transformer; consistently poor at moderate pruning rates.}
    \label{fig:app-sens-13}
\end{figure}

\begin{figure}[H]
    \centering
    \includegraphics[width=0.55\linewidth]{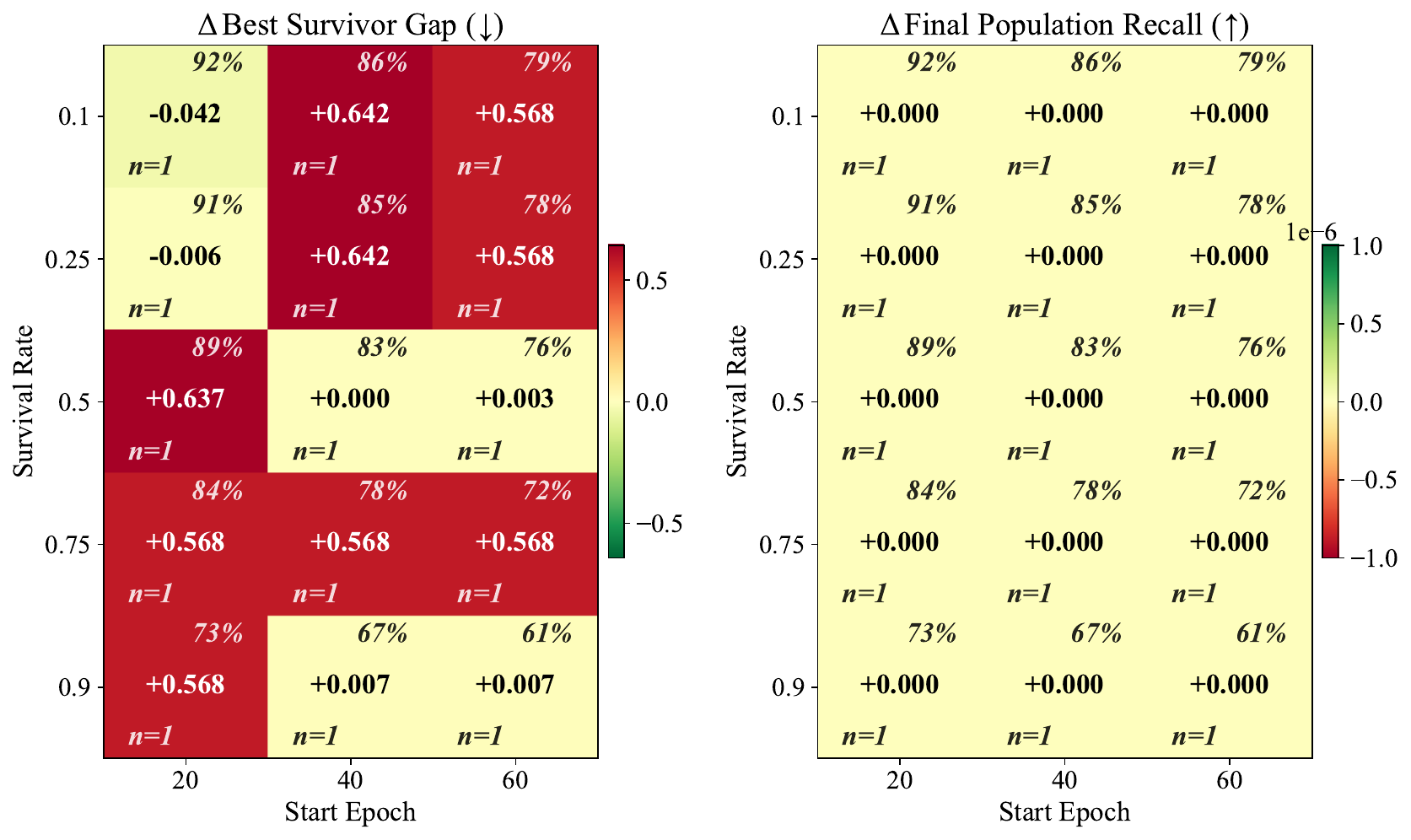}
    \caption{CIFAR-10 ViT, horizon value: latent-window LSTM; generally worse despite high compute savings.}
    \label{fig:app-sens-14}
\end{figure}

\begin{figure}[H]
    \centering
    \includegraphics[width=0.55\linewidth]{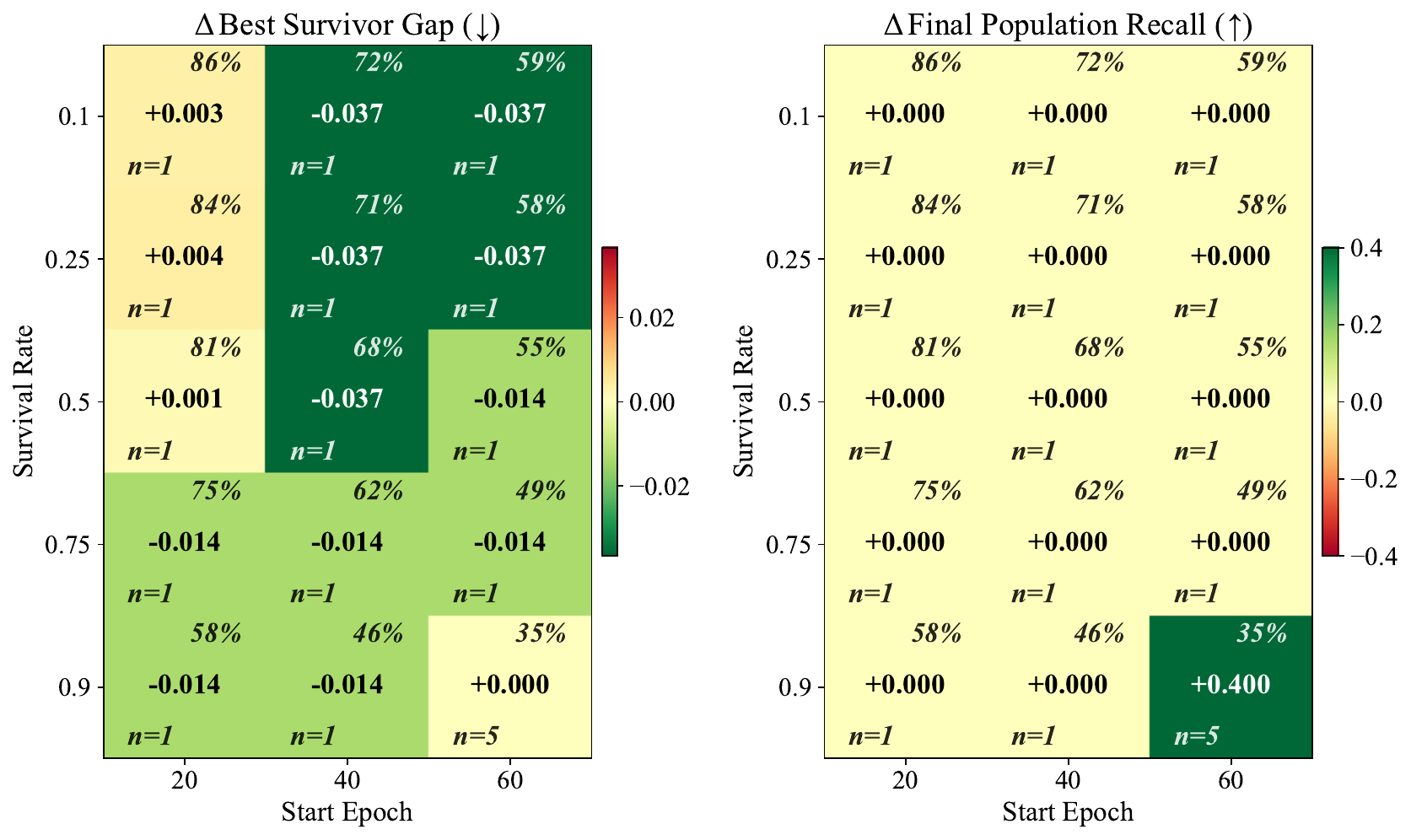}
    \caption{CIFAR-10 width-varying ResNet-18, final value: latent-window LSTM; helps mainly when pruning starts at epoch 40 or 60.}
    \label{fig:app-sens-15}
\end{figure}

Section~\ref{sec:results} reads specific cells of the full start-epoch/retention-rate grid for
the Tiny ImageNet ResNet-18 zoo and the CIFAR-10 ViT zoo, the two zoos with the
largest selection headroom in Table~\ref{tab:app-sotf-per-zoo}.
Figures~\ref{fig:app-sens-1}--\ref{fig:app-sens-14} give the full parameter-sensitivity grids
for a representative selection of configurations on these two zoos, chosen to include
both robust and unsuccessful cases rather than only favorable ones: each cell reports the gap
difference relative to LC Extrapolation, compute savings in the upper-right corner, and final
population size in the lower-left corner. In final-value mode, LCNet is the stronger
learning-curve baseline on the Tiny ImageNet and CIFAR-10 width zoos, with a lower gap than
LC Extrapolation in 13 and 12 of 15 configurations, respectively; gains relative to LCNet are
therefore smaller there than the grids show. All runs use a decision interval of 5 epochs; the
horizon-value runs additionally predict 5 epochs ahead of the decision point. Rows index the
survival rate and columns the start epoch of the first pruning decision.
Figure~\ref{fig:app-sens-15} additionally reports the same grid for the latent-window LSTM
on the CIFAR-10 width-varying ResNet-18 zoo, discussed in Section~\ref{sec:results}
for its schedule-dependent behavior.

The mix of outcomes above is deliberate: several configurations (e.g.\ Figures~\ref{fig:app-sens-7},
\ref{fig:app-sens-11}--\ref{fig:app-sens-14}) are \emph{worse} than the learning-curve
extrapolation baseline across most of the grid. SOTF training's benefit therefore comes from selecting
among these configurations using validation performance, not from every checkpoint-based
predictor improving pruning automatically.

\section{Compute resources and reproducibility}
\label{app:compute-resources}

\paragraph{What is and is not measured.}
As stated in Section~\ref{sec:discussion}, ``training epochs saved'' is an architecture- and
hardware-independent proxy for compute and energy reduction; we do not report GPU-hours,
wall-clock runtime, FLOPs, or energy consumption for zoo training, SANE backbone training,
predictor training, or the SOTF training simulation itself, and the reported compute savings do not
include the cost of computing weight statistics or of training and applying SANE encoders.
This limitation is intentional rather than an oversight of reporting: the underlying model
zoos were trained and checkpointed prior to this work, so their original training cost is not
directly re-measurable, and the SANE backbones and predictors used here were trained once per
zoo, offline, ahead of the SOTF training evaluation, rather than inside a live, wall-clock-
metered training loop.

\paragraph{Code availability.}
The code to reproduce our experiments is available at
\url{https://github.com/RicoStaedeli/sotf-training}.


\end{document}